\documentclass{article} % For LaTeX2e
\usepackage{iclr2025_conference,times}

\usepackage{amsmath,amsfonts,bm}

\def\eqref#1{equation~\ref{#1}}
\def\1{\bm{1}}

\DeclareMathAlphabet{\mathsfit}{\encodingdefault}{\sfdefault}{m}{sl}
\SetMathAlphabet{\mathsfit}{bold}{\encodingdefault}{\sfdefault}{bx}{n}

\usepackage[utf8]{inputenc} % allow utf-8 input
\usepackage[T1]{fontenc}    % use 8-bit T1 fonts
\usepackage{hyperref}       % hyperlinks
\usepackage{url}            % simple URL typesetting
\usepackage{booktabs}       % professional-quality tables
\usepackage{amsfonts}       % blackboard math symbols
\usepackage{nicefrac}       % compact symbols for 1/2, etc.
\usepackage{microtype}      % microtypography
\usepackage{xcolor}         % colors
\usepackage{graphicx}
\usepackage{longtable}
\usepackage{array}
\usepackage{pbox}
\usepackage{amsmath}
\usepackage{amssymb}
\usepackage{tabularx}
\usepackage{multirow}
\usepackage{wrapfig}
\usepackage{pifont}
\usepackage{algorithm}
\usepackage{algorithmic}
\usepackage{lipsum}
\usepackage{subcaption}
\usepackage{tcolorbox}

\usepackage{colortbl} % 为表格增加颜色
\usepackage{newfloat}
\usepackage{listings}

\newcommand{\cmark}{\textcolor{green}{\checkmark}} % Green checkmark
\newcommand{\xmark}{\textcolor{red}{\ding{55}}}      % Red cross

\usepackage{tikz}
\newcommand{\coloredcircle}[2]{%
    \tikz[baseline=(char.base)]{
        \node[shape=circle,draw=#1,fill=#1!20,inner sep=1pt] (char) {\textcolor{black}{#2}};}%
}

\newcommand{\resultone}[1]{\colorbox{green!15}{#1}}
\newcommand{\resulttwo}[1]{\colorbox{cyan!15}{#1}}
\newcommand{\resultthird}[1]{\colorbox{yellow!15}{#1}}

\title{MTU-Bench: A Multi-Granularity Tool-Use Benchmark for Large Language Models}
\author{
Pei Wang$^{1*}$, ~~Yanan Wu$^{1*}$,~~Zekun Wang$^{1*}$,
~\textbf{Jiaheng Liu}$^{1\dagger}$,\\
~\textbf{Xiaoshuai Song}$^{1}$,
~~\textbf{Zhongyuan Peng}$^{1,2}$,~~\textbf{Ken Deng}$^{1}$,~~\textbf{Chenchen Zhang}$^1$, ~~\textbf{Jiakai Wang}$^1$,\\
~\textbf{Junran Peng}$^2$,
~\textbf{Ge Zhang}$^3$,
~\textbf{Hangyu Guo}$^1$,
~~\textbf{Zhaoxiang Zhang}$^{2}$,
~~\textbf{Wenbo Su}$^1$, ~~\textbf{Bo Zheng}$^1$\\
~$^1$Alibaba Group, ~$^2$University of Chinese Academy of Sciences, ~$^3$University of Waterloo\\ 
~~\texttt{\{yupei.wp, lixing.wyn, ljh411989\}@alibaba-inc.com
}\\
}
\iclrfinalcopy % Uncomment for camera-ready version, but NOT for submission.

\begin{document}

\maketitle
\let\oldthefootnote\thefootnote

\let\thefootnote\relax\footnotetext{* First three authors contributed equally. ~~$^\dagger$ Corresponding Author: Jiaheng Liu.}
\let\thefootnote\oldthefootnote
\begin{abstract}
Large Language Models (LLMs) have displayed massive improvements in reasoning and decision-making skills and can hold natural conversations with users.
% Many recent works seek to augment LLM-based assistants with external tools
% so they can access private or up-to-date information and carry out actions on
% behalf of users.
Recently, many tool-use benchmark datasets have been proposed.
However, existing datasets have the following limitations:
(1). Insufficient evaluation scenarios (e.g., only cover limited tool-use scenes).
(2). Extensive evaluation costs (e.g., GPT API costs).
To address these limitations,
in this work,
we propose a multi-granularity tool-use benchmark for large language models called MTU-Bench. For the ``multi-granularity'' property,
our MTU-Bench covers five tool usage scenes (i.e., single-turn and single-tool, single-turn and multiple-tool, multiple-turn and single-tool, multiple-turn and multiple-tool, and out-of-distribution tasks).
Besides,
% For the ``fully automatic'' property,
all evaluation metrics of our MTU-Bench are based on the prediction results and the ground truth without using any GPT or human evaluation metrics.  
Moreover, our MTU-Bench is collected by transforming existing high-quality datasets to simulate real-world tool usage scenarios,
and we also propose an instruction dataset called MTU-Instruct data to enhance the tool-use abilities of existing LLMs.
Comprehensive experimental results demonstrate the effectiveness of our MTU-Bench.
Code and data will be released at ~\url{https://github.com/MTU-Bench-Team/MTU-Bench.git}.
\end{abstract}

\section{Introduction}
\begin{wrapfigure}[21]{r}{0.4\textwidth}
    \centering
    \includegraphics[width=0.4\textwidth]{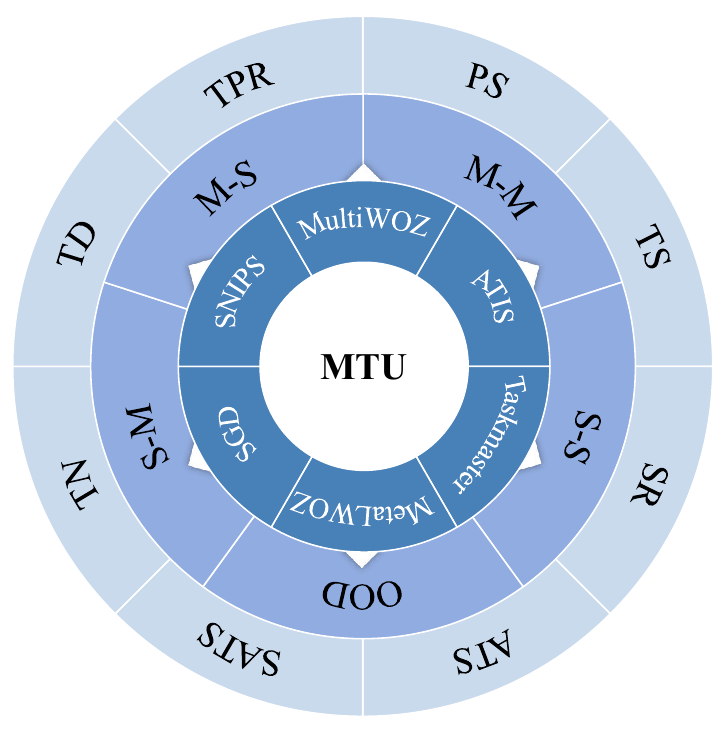}
    \caption{The circles from inside to outside represent the data source, scenes included in \textbf{MTU-Instruct}, and the automatic evaluation metrics in \textbf{MTU-Eval}.}
    \label{fig:intro}
\end{wrapfigure}
Since the release of large language models (LLMs) such as GPT-4~\citep{gpt4}, 
Natural Language Processing (NLP) has entered a new wave of advancements, 
even being considered as the spark of Artificial General Intelligence (AGI)~\citep{bubeck2023sparks-AGI}. 
% Through large-scale pretraining and supervised fine-tuning on diverse tasks, 
% LLMs not only excel at traditional NLP tasks, 
% but also exhibit more human-like capabilities such as 
% mathematical reasoning~\citep{yue2023mammoth, shao2024deepseekmath}, 
% code generation~\citep{rozière2023codellama, Chai2024mceval}, 
% and role-playing~\citep{wang2023rolellm}.
Recently, there has been a surge of research focused on enabling LLMs to interface with external tools, 
such as calculators~\citep{math-verifier}, search engines~\citep{toolformer}, and booking service APIs~\citep{qin2023toolllm}. 
This approach, referred to as Tool Learning~\citep{toolformer, qin2023tool, qin2023toolllm, wang2023interactive}, 
allows LLMs to not only accurately perform precise calculations, 
but also maintain up-to-date information. 
Furthermore, it enables LLMs to function as end-to-end AI assistants that are capable of fulfilling real-world user needs such as booking hotels or ordering food. 
Thus, Tool Learning is a critical step to transform LLMs into general AI agents.

% Recently, there has been a surge of research focused on enabling LLMs 通过使用工具来满足真实世界的用户需求。之前的工作，已经取得了一些进展，但是相较于真实的使用场景，存在一些局限。例如：在真实的场景中，通常是多轮的，此外也需要通过使用多个工具才能满足用户需求。为此我们提出了，MTU 一个更贴近真实世界的工具调用 Benchmark 和 训练集。此外，之前的工作中存在的另一问题是，缺少对模型工具调用错误原因的细粒度分析，为此，我们设置了许多评估模型行为的更细粒度的指标，例如：the accuracy of tool call orders, complex tool calls involving inheritance relationships, per-dialogue turn accuracy of tool name and parameter selection, as well as four-quadrant analysis of single/multi-turn and single/multi-tool scenarios, etc. 通过这些指标我么可以对模型行为有更细致的了解。在 *** 中，我们展示了目前许多模型在我们 Benchmark 上的结果，并且分析了他们各自的优劣势，我们惊奇的发现，在我们的更真实的 Benchmark 上，模型产生了许多不一样的结论。我们期望通过这些发现，可以为未来工具调用的研究提供更新的 Insigt。

Previous works have explored to stimulate the ability to call tools for LLMs~\citep{toolformer, qin2023toolllm, zhuang2023toolqa, tang2023toolalpaca, art-tool, li2023apibank}. 
For example, ~\citep{toolformer} propose to convert tool calls into text spans, such as \textsc{$<$API$>${some parameter key-value pairs}$<$/API$>$} to denote the tool name and parameters with an additional special token (i.e., \textsc{$<$/API$>$}) to show the initiation of a tool execution. 
% Nevertheless, constructing high-quality instruction tuning data to enhance this capability and thoroughly evaluating LLMs' tool-use capabilities remain areas that have not been fully explored.
Moreover, the recent works (i.e., ToolBench~\citep{xu2023toolbench, qin2023toolllm}, APIBench~\citep{patil2023gorilla-apibench}, and API-Bank~\citep{li2023apibank}) have investigated instruction tuning data or evaluation for tool-use.

However, we observe that exhibit several limitations to varying degrees as shown in Table \ref{tab:dataset-comparison}: 
(1) some do not consider multi-turn dialogue scenarios~\citep{patil2023gorilla-apibench, xu2023toolbench, zhuang2023toolqa}; 
(2) some do not address multi-tool usage scenarios~\citep{tang2023toolalpaca, patil2023gorilla-apibench, li2023apibank, xu2023toolbench}; 
(3) several works use external API tools to deduce user instructions, but these synthesized instructions often do not accurately align with actual real-world user needs~\citep{qin2023toolllm}; 
(4) many of them rely on GPT for the evaluation, leading to heavy evaluation costs~\citep{qin2023toolllm, tang2023toolalpaca}; 
and (5) many do not comprehensively assess fine-grained aspects of tool-use~\citep{li2023apibank, qin2023toolllm, patil2023gorilla-apibench}, such as the accuracy of tool call orders, complex tool calls involving inheritance relationships, per-dialogue turn accuracy of tool and parameter selection, four-quadrant analysis of single/multi-turn and single/multi-tool settings, etc. 

\begin{table}[t]
    \scriptsize
    \centering
    \caption{Comparison of various tool-use benchmark datasets. ``Auto. Eval.'' denotes ``automatic evaluation without GPT''. ``\textbf{S-S}'', ``\textbf{S-M}'', ``\textbf{M-S}'', and ``\textbf{M-M}'' denote single-turn single-tool, single-turn multi-tool, multi-turn single-tool, and multi-turn multi-tool, respectively. ``OOD'' refers to whether the benchmark accounts for an out-of-distribution setting, where the test set consists of domains different from those in the training set. ``Real-World'' means whether the dialogues in the benchmark are sampled from real-world scenarios. The numbers in the evaluation range mean: \protect\coloredcircle{red}{1} tool selection, \protect\coloredcircle{orange}{2} parameter selection, \protect\coloredcircle{brown}{3} dialogue-level success rate, \protect\coloredcircle{green}{4} turn-based success rate, \protect\coloredcircle{blue}{5} tool number, and \protect\coloredcircle{purple}{6} tool order.}
    % \begin{tabular}{p{2.2cm}|p{1.2cm}p{0.7cm}p{1.5cm}p{0.6cm}p{0.6cm}p{1.7cm}p{0.4cm}p{0.4cm}p{0.4cm}}
    \begin{tabular}{m{2.2cm}|m{1.2cm}m{0.7cm}m{1.5cm}m{0.6cm}m{0.6cm}m{1.7cm}m{0.4cm}m{0.4cm}m{0.4cm}}
    \toprule
    \textbf{Dataset} & \textbf{\#Dialogues} & \textbf{\#Tools} & \textbf{\#Turn-\#Tool} & \textbf{Real-World} & \textbf{Auto. Eval.} & \textbf{Eval. Range} & \textbf{Train} & \textbf{Test} & \textbf{OOD} \\
    \midrule
    \textbf{MetaTool} \newline ~\citep{metatool} & 21,127 & 199 & S-S, S-M & \xmark & \cmark & \coloredcircle{red}{1}\coloredcircle{orange}{2}\coloredcircle{brown}{3} & \xmark & \cmark & \xmark \\ \midrule
    \textbf{API-Bank} \newline ~\citep{li2023apibank} & 2,202 & 2,211 & S-S, S-M,\newline M-S, M-M & \xmark & \cmark & \coloredcircle{red}{1}\coloredcircle{orange}{2}\coloredcircle{brown}{3} & \cmark & \cmark & \xmark \\ \midrule
    \textbf{ToolLLM} \newline ~\citep{qin2023toolllm} & 12,657 & 16,464 & S-S, S-M & \xmark & \xmark & \coloredcircle{orange}{2}\coloredcircle{brown}{3} & \cmark & \cmark & \cmark \\ \midrule
    \textbf{API-Bench} \newline ~\citep{patil2023gorilla-apibench} & 17,002 & 1,645 & S-S & \xmark & \cmark & \coloredcircle{orange}{2}\coloredcircle{brown}{3} & \cmark & \cmark & \xmark \\ \midrule
    \textbf{ToolAlpaca} \newline ~\citep{tang2023toolalpaca} & 3,938 & 400 & S-S, S-M,\newline M-S, M-M & \xmark & \xmark & \coloredcircle{red}{1}\coloredcircle{orange}{2}\coloredcircle{brown}{3}\coloredcircle{green}{4} & \cmark & \cmark & \xmark \\ \midrule
    \textbf{ToolQA} \newline ~\citep{zhuang2023toolqa} & 1,530 & 13 & S-S, S-M & \xmark & \cmark & \coloredcircle{orange}{2}\coloredcircle{brown}{3} & \xmark & \cmark & \xmark \\ \midrule
    \textbf{MTU-Bench} \newline (Ours) & 159,061 & 136 & S-S, S-M,\newline M-S, M-M & \cmark & \cmark & \coloredcircle{red}{1}\coloredcircle{orange}{2}\coloredcircle{brown}{3}\coloredcircle{green}{4}\coloredcircle{blue}{5}\coloredcircle{purple}{6} & \cmark & \cmark & \cmark \\
    \bottomrule
\end{tabular}
    \vspace{-5mm}
    \label{tab:dataset-comparison}
\end{table}

To remedy these issues, in Figure~\ref{fig:intro}, we introduce \textbf{MTU-Bench} (\underline{M}ulti-Granularity \underline{T}ool-\underline{U}se \underline{Bench}mark), which comprises both \textbf{MTU-Instruct} for training and \textbf{MTU-Eval} for evaluation. 
As illustrated in Figure \ref{fig:mtu-bench}, we sample real-world user instructions from various existing open-source dialogue datasets such as MultiWOZ~\citep{budzianowski2018multiwoz} and SGD~\citep{rastogi2020sgd, lee2022sgd-x}. 
After instruction clustering, the detected user intents and slot filling are leveraged to synthesize API calls using GPT-4~\citep{gpt4}. 
The synthesized data includes the thoughts, the actions (\textit{i.e.}, tool names), the action parameters, and the observations (\textit{i.e.}, the generated API execution results). 
This data forms our MTU-Bench dataset. 
Following meticulous quality verification by GPT-4 and manual check, 
we split the MTU-Bench data into training and testing splits, involving 54798 dialogues in total, as well as 136 tools. 
In our MTU-Eval, we propose a series of fine-grained metrics such as tool selection accuracy, parameter selection accuracy, success rate, turn success rate, task process rate, tool number accuracy, tool order accuracy, etc., to evaluate the tool-use abilities in a comprehensive manner,
where the GPT API costs are not needed for evaluation.
Moreover,
we also pick out a hard subset from the test split to include more complex tool-use scenarios such as easily confusable tools, nonsensical or noisy tools, tool parameter updating, etc. 

Finally, by fine-tuning LLaMA3~\citep{dubey2024llama3} on MTU-Bench, we find that our resulting model, MTU-LLaMA, performs the best in various scenarios and metrics, demonstrating the effectiveness of our MTU-Instruct.

In summary, our contributions are as follows:
(1). \textbf{MTU-Bench}: We introduce a novel automated data synthesis pipeline designed to derive high-quality, fine-grained tool-use datasets from pre-existing task-oriented dialogue datasets. This pipeline facilitates the creation of MTU-Bench, comprising MTU-Instruct for training purposes and MTU-Eval for evaluation. (2). \textbf{MTU-Instruct and MTU-Eval}: We introduce the high-quality and diverse instruction tuning dataset, MTU-Instruct, to improve models' tool-use capabilities in real-world scenarios. Additionally, we propose a novel automatic evaluation framework, MTU-Eval, which assesses various tool-use settings through comprehensive and fine-grained metrics, free of GPT-based evaluators. (3). \textbf{MTU-LLaMA and Experimental Findings}: After instruction tuning on MTU-Instruct, we obtain a strong open-source model for tool-use, MTU-LLaMA. Our comprehensive experiments reveal several findings regarding the tool-use capabilities of LLMs, particularly in terms of multi-turn dialogue scenarios, multi-tool settings, and error cases. These findings offer valuable insights for advancing tool-use in LLMs.
    
% \begin{enumerate}
%     \item \textbf{MTU-Bench}: We introduce a novel automated data synthesis pipeline designed to derive high-quality, fine-grained tool-use datasets from pre-existing task-oriented dialogue datasets. This pipeline facilitates the creation of MTU-Bench, comprising MTU-Instruct for training purposes and MTU-Eval for evaluation.
%     \item \textbf{MTU-Instruct and MTU-Eval}: We introduce the high-quality and diverse instruction tuning dataset, MTU-Instruct, to improve models' tool-use capabilities in real-world scenarios. Additionally, we propose a novel automatic evaluation framework, MTU-Eval, which assesses various tool-use settings through comprehensive and fine-grained metrics, free of GPT-based evaluators.
%     % This benchmark is the first multi-turn, multi-tool, and multi-granular tool-use benchmark.
%     \item \textbf{MTU-LLaMA and Experimental Findings}: After instruction tuning on MTU-Instruct, we obtain a strong open-source model for tool-use, MTU-LLaMA. Our rigorous and comprehensive experiments reveal several findings regarding the tool-use capabilities of LLMs, particularly in terms of multi-turn dialogue scenarios, multi-tool settings, and error cases. These findings offer valuable insights for advancing tool-use in LLMs.
% \end{enumerate}

\section{MTU-Bench}
The MTU-Bench involves both MTU-Instruct for training and MTU-Eval for evaluation. 
We first present the data construction and analysis in \S\ref{sec:mtu-instruct}, and then show the evaluation procedure in \S\ref{sec:mtu-eval}. 

\subsection{MTU-Instruct\label{sec:mtu-instruct}}

\begin{figure*}[h]
    \centering
    \includegraphics[width=\linewidth]{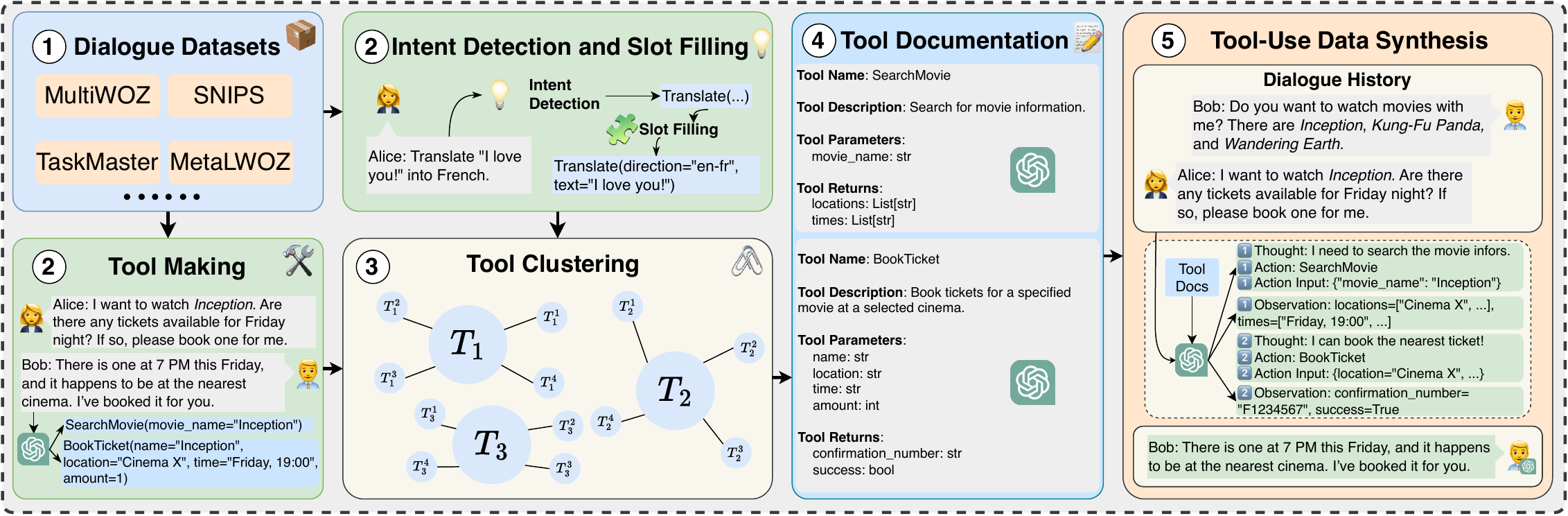}
    \vspace{-5mm}
    \caption{The workflow for MTU-Instruct construction. It involves five steps: (1) Data Collection, (2) Tool Creation, (3) Tool Clustering, (4) Tool Documentation, and (5) Tool-Use Synthesis.}
    \label{fig:mtu-bench}
    \vspace{-5mm}
\end{figure*}

\subsubsection{Data Construction}
% 数据集构建整体解释：1. 先前基于爬取的API文档和GPT4合成方法，受限于数据多样性，泛化能力较弱；2.当前还有广泛的NLP多年沉淀的数据没有被充分挖掘利用。受意图与工具，槽位与参数的映射关系启发。因此，我们设计了一套从传统对话数据到工具使用数据的自动转换流程。 为提高数据多样性，我们从8组数据源中进行数据收集。并将其规范为统一的工具文档。多样性可见表x，图x验证了我们数据质量和有效性。

% Collection：Motivation; 源数据集；
% Tool Creation: Motivation适应多种数据源，考虑多轮中的复杂情况
    % 1)基于规则的方式，对于已有intent和slot的；
    % 2)基于LLM的方式，对于缺少intent或slot的；spec tool/no call/lack info for ask/confirm info/。% 结合history&current, 根据assistant情况划分，若正面解决回答了用户问题，则提取对应的工具；若从历史对话中返回信息information already exists；无意义，则nocall；若反问信息；若确认信息，则confirm；
    % (示例需要换一下)
% clsutring: 缓解数据集间工具名和参数重叠。instag:https://github.com/OFA-Sys/InsTag/tree/main
% Documentation: 统一不同源为同一格式。（有空地可以补充一个结构示例）
% Tool-use syn: 
    % 参考react格式，分为thought,action,action_input三部分
    % 补充obsevation
    % 校验：校验前后数据量，校验规则；
    % 最后，分为四场景，数据总量。

Many previous methods based on scraped API documentation and GPT-4 inspired synthesis faced limitations due to a lack of data diversity, resulting in weak generalization capabilities. 
% We discovered that a vast amount of NLP dataset remains underutilized. 
Inspired by the mapping relationships between intents and APIs, as well as between slots and tool parameters, we designed an automated data synthesis pipeline that transforms traditional dialogue datasets into tool-use datasets. To enhance data diversity, we collected datasets from multiple NLP datasets and standardized them into a unified tool documentation. The diversity is illustrated in Figure \ref{fig:domain-sunburst}.
% while Table \ref{} validates the quality and effectiveness of our dataset. \\
As shown in Figure \ref{fig:mtu-bench}, the construction of MTU-Bench involves five primary steps: 
(1) collecting task-oriented dialogue datasets, particularly those containing intents and slots,
(2) creating tools through grammar-based transformations or GPT-4 synthesis, 
(3) clustering the tools based on their similarities, 
(4) generating tool documentation using GPT-4, and 
(5) synthesizing tool-use samples consisting of thoughts, actions(tool calls), action inputs, observations, and adjusted responses based on the dialogue and tool documentation, followed by a holistic validation process.

% As illustrated in Figure \ref{fig:mtu-bench}, the construction of MTU-Bench involves five primary steps: 
% (1) collecting task-oriented dialogue datasets, particularly those that have undergone intent detection and slot-filling procedures, 
% (2) creating tools through GPT-4 synthesis or by converting detected intents and filled slots into tool names and parameters, 
% (3) clustering the tools based on their similarities and standardizing tool names within each cluster to their corresponding centroids, 
% (4) generating tool documentation using GPT-4, 
% and (5) synthesizing tool-use samples based on the dialogues and tool documentation, which includes dialogue history, thoughts, actions (tool calls), observations, and final responses.

\paragraph{Data Collection.} 
% TODO： 相关数据集在附录补充具体介绍。
To improve the diversity of our dataset, we collect several open-source task-oriented dialogue datasets as our data sources. 
These datasets focus on dialogues for specific tasks such as flight reservations or movie bookings, which are highly suitable for synthesizing tool-use data. 
The multi-turn dialogue datasets include MultiWOZ~\citep{budzianowski2018multiwoz}, SGD~\citep{rastogi2020towards}, TaskMaster~\citep{byrne2019taskmaster1} and MetaLWOZ~\citep{shalyminov2020fast}.
The single-turn dialogue datasets include ATIS~\citep{Hemphill1990TheAS} and SNIPS~\citep{snips}.
They provide diverse task-oriented dialogues across various domains, real-world conversation and fine-grained annotation, encompassing both single-turn and multi-turn dialogues. 

\paragraph{Tool Creation.} 
We employ two approaches to create tools. 
(1) \textbf{Grammar-based creation}. For dialogue datasets that already have detected intents and filled slots, we directly convert the intents into tool names and the slots into tool parameters. For example, in the user query \textit{``find a flight from charlotte to las vegas''}, the intent \textit{``Flight''} will convert to the tool name, and the slot \textit{``from\_location=charlotte, to\_location=las vegas''} will convert to the parameters, resulting in the tool call \textsc{Flight(from\_location=``charlotte'',to\_location=``las vegas'')}.
(2) \textbf{LLM-based creation}. For dialogue datasets without predefined intents or slots, we utilize GPT-4 to make the tools. 
Based on the contextual situation, we categorize it into five scenarios: information missing, information exists, information confirmed,  aimless chatting and specific API call.
Provided with the historical dialogue context and the current round of conversation, especially the assistant's response, LLM needs to determine which situation belongs to the current situation.
% and whether an API call is required to address the user's query. 
\textbf{Information missing.} If the response is asking for important information, it should be the situation of missing information, no tool call should be made, and necessary parameters should be accumulated for related tools.
\textbf{Information exists.} When the LLM can provide a response based solely on the information from the dialogue history, no tool call will be made and the model can directly reply.
\textbf{Information confirmed.} When the assistant is confirming information (e.g.,\textit{Would you like to confirm this flight reservation?}), this is classified as ``information to be confirmed''.
\textbf{Aimless chatting.} If the scenario pertains to aimless chatting or situations that do not necessitate tool invocation, no tool call should be made.
\textbf{Specific API call.} Only if the LLM determines that an API call is necessary to fulfill the user’s request, it is encouraged to generate an appropriate pseudo-tool for invocation, along with a corresponding description and parameters of this tool.

\paragraph{Tool Clustering.} 
Due to the diversity in both the intent detection and slot-filling strategies, as well as the creation of specific tools based on LLM, the synthesized tool set can be highly redundant.
% Due to the diversity in both the intent detection and slot filling strategies, as well as in the GPT-4 tool-making process during the tool creation phase, the synthesized tool set can be highly redundant. 
For example, tools like ``search\_movie'' and ``find\_movie'' may have different names but essentially perform the same function. 
To address this redundancy, we introduce a tool clustering phase. 
Specifically, we cluster the tool names based on InsTag\citep{lu2023instag} with a fixed distance threshold. Then, all tool names and their parameter names are standardized to the centroid of their respective cluster, resulting in a reduction ratio of 20:1.

\paragraph{Tool Documentation.} 
To enable models to use specific tools effectively, we compile all tool usage into a comprehensive tool document. 
This document allows the model to determine the appropriate tool names and their usage. We prompt the GPT-4 model to write a description for each tool generated in the previous step, along with information about its parameters(required and optional) and returns. 
The collection of these tool entries forms the final tool document, which is included as part of the LLMs' context. Please refer to Appendix \ref{appendix: tool_document} for the details.

\paragraph{Tool-Use Data Synthesis.} 
In this step, we convert all samples from task-oriented dialogue datasets into tool-use dataset with GPT-4, following the format of ReAct~\citep{yao2023react}.
We provide the dialogue history and the tool document generated in the previous steps as context for GPT-4, and then prompt it to generate three key components: 
(1) \textbf{thought}: the reasoning process behind the tool selection, 
(2) \textbf{action}: the name of the tool being invoked, 
and (3) \textbf{action input}: the parameters used in the tool call along with their values. 
This chain of thought prompting technique enhances the model's ability to reason over the most appropriate tool and accurately input the parameter values. 
% thereby reducing unnecessary tool calls and incomplete tool usage. 
We also allow the model to generate any additional parameters needed that are beyond those listed in the tool document, to ensure completeness and flexibility of tool-use.

We further ask the GPT-4 to simulate tool execution, generating \textbf{observation} (i.e., the results of the simulated tool execution) and then produce the final \textbf{response} for the current dialogue turn. 
The observations are aligned with the return information in the tool document and are generated in a structured format, such as a JSON dictionary. 
The model then formulates a response based on the observations, either to report the status of the tool execution or to complete the dialogue turn.

To ensure data quality, we apply various quality filters and adjustments, including heuristic rules, GPT-4, and manual annotation. 
For example, we split some multi-turn dialogues into single-turn dialogues to balance the data distribution. 
For the training set, we use GPT-4 to verify the accuracy and necessity of tool selection, check parameter matching, adjust thoughts, rewrite response and ensure consistency in the tool definitions. Through GPT-4, we filter out about 10\% of defective samples.
For the test set, we hired multiple experts to conduct manual quality checks based on similar principles. Each sample was checked by three experts, and the differences in labeling were determined by the fourth expert.

Based on the number of dialogue turns and the number of tools in each dialogue, 
the synthesized data can be categorized into four types: (1) Single-turn Single-tool (\textbf{S-S}), (2) Single-turn Multi-tool (\textbf{S-M}), (3) Multi-turn Single-tool (\textbf{M-S}), and (4) Multi-turn Multi-tool (\textbf{M-M}).

For more detailed information about the construction of MTU-Bench, including prompt templates and the tool documentation, please refer to the Appendix \ref{appx:mtu-bench-construction-details}.

\subsubsection{Data Analysis}

\begin{figure}[h]
    \centering
    \begin{minipage}[t]{0.6\linewidth}
        \centering
        \includegraphics[width=\linewidth]{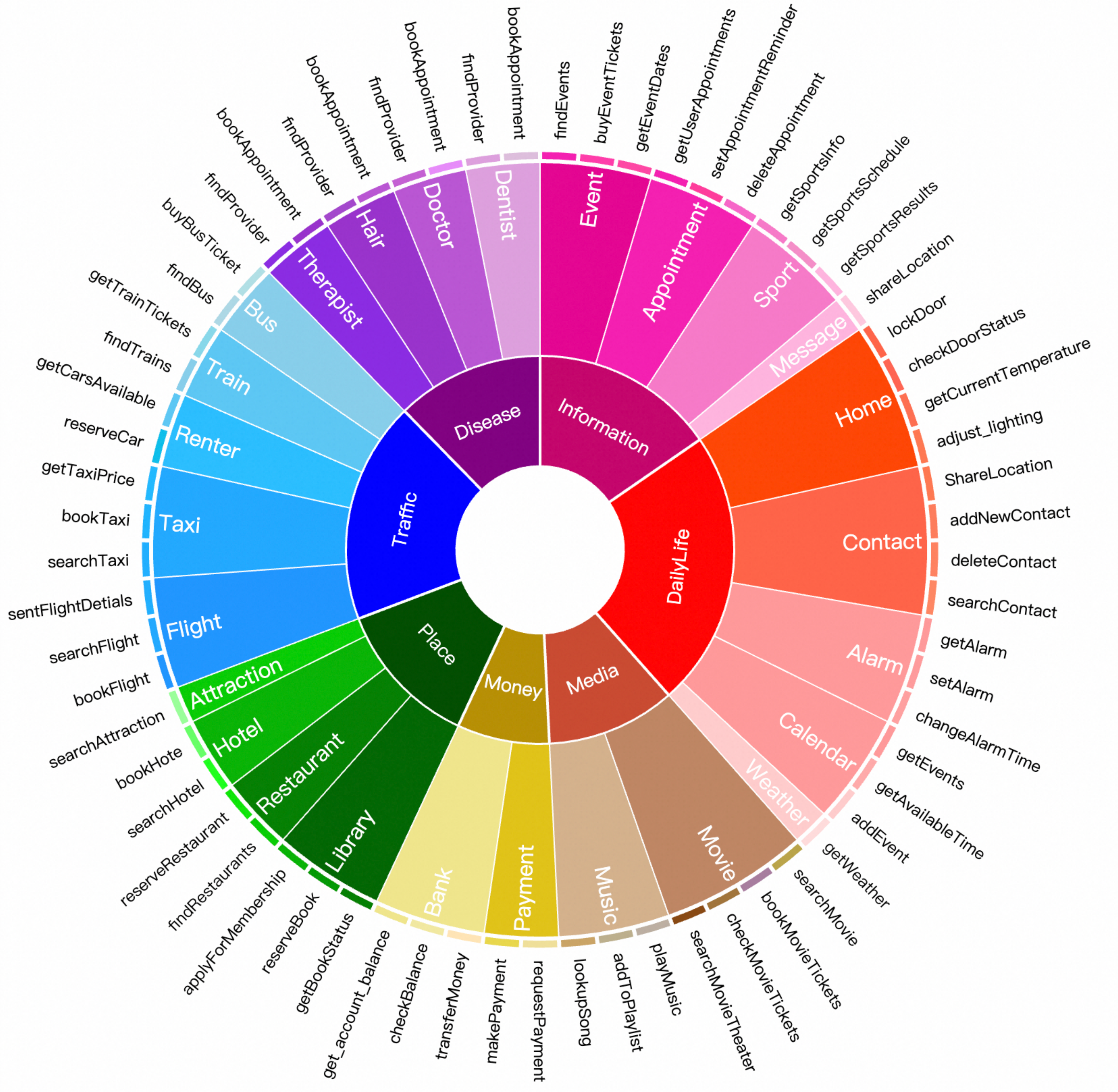}
        % \vspace{-2mm}
        \caption{The domain distribution of tools.}
        \label{fig:domain-sunburst}
    \end{minipage}%
    \hfill
    \begin{minipage}[t]{0.4\linewidth}
        \vspace{-8.6cm}
        \centering
        \scriptsize
        \begin{tabular}{p{1.3cm}|p{0.5cm}p{0.5cm}p{1.3cm}}
        \toprule
        \textbf{Domain} & \textbf{\#Topics} & \textbf{\#Tools} & \textbf{\#Dialogues (Train/Test)} \\
        \midrule
        \textbf{Place} & 4 & 27 & 29,667 / 150 \\ \midrule
        \textbf{Media} & 2 & 18 & 13,730 / 92 \\ \midrule
        \textbf{Traffic} & 5 & 22 & 17,720 / 89 \\ \midrule
        \textbf{Daily Life} & 5 & 31 & 23,706 / 163 \\ \midrule
        \textbf{Money} & 2 & 10 & 1,740 / 32 \\ \midrule
        \textbf{Information} & 4 & 12 & 9,125 / 75 \\ \midrule
        \textbf{Disease} & 4 & 2 & 3,371 / 125 \\ \midrule
        \textbf{Others} & 5 & 14 & - / 68 \\ \midrule
        \toprule
        \multicolumn{4}{l}{\textbf{Additional Statistics}} \\ 
        \midrule
        \#Dialogues (Train/Test) & \multicolumn{3}{c}{54,367 / 431} \\ \midrule
        \#Tools (Train/Test) & \multicolumn{3}{c}{122 / 14} \\ \midrule
        Avg. Turns per Dialogue & \multicolumn{3}{c}{2.6} \\ \midrule
        Avg. Tools per Dialogue & \multicolumn{3}{c}{5.6} \\ \midrule
        Avg. Tools per Turn & \multicolumn{3}{c}{2.2} \\ 
        \bottomrule
        \end{tabular}
        \vspace{-2mm}
        \caption{Statistics of MTU-Bench.}
        \label{table:statistics}
    \end{minipage}
    \vspace{-5mm}
\end{figure}

\paragraph{Statistics.} 
Figure \ref{table:statistics} delineates the statistical metrics of our MTU-Bench, underscoring its substantial scale and diversity. It is particularly designed to encompass multi-tool and multi-turn settings, as well as \emph{real-world} domains and topics.

% \paragraph{Toolset Visualization.} 
% Figure \ref{fig:tool-visualization} illustrates the distribution of tools within our MTU-Bench, demonstrating the diversity and practical utility of the set of tools.
\paragraph{Diversity.} 
Figure \ref{fig:domain-sunburst} illustrates the domain and topic distribution of the tools within our MTU-Bench. Figures \ref{fig:length_distribution_train} and \ref{fig:length_distribution_eval} depict the distributions of dialogue turn counts, word counts, tool numbers, and other length-related metrics across both training and testing splits. These figures underscore the diversity of our MTU-Bench in terms of length and topic distribution.

We refer the readers to Appendix \ref{appx:data-analysis} for a more detailed data analysis and statistics.

\subsection{MTU-Eval\label{sec:mtu-eval}}

% To comprehensively evaluate the tool-use capabilities of LLMs, 
We propose \textbf{MTU-Eval}—the first evaluation framework that encompasses multiple levels of difficulty, diverse domains and cases of tool-use, varying numbers of dialogue turns and tools, as well as multifaceted evaluation by considering various granularities and aspects of LLM tool-use.

We present MTU-Eval in two parts: (1) Test Set Splitting and (2) Evaluation Metrics.

\subsubsection{Test Set Splitting}

We construct two distinct test sets from MTU-Bench through manual sampling: (1) the normal test set and (2) the more challenging hard test set.

\paragraph{Normal/Hard Test Set.} 
MTU-Bench includes data from 31 different topics, such as weather-related and calendar-related tasks. 
Initially, we select data from 5 topics as an Out-of-Distribution (OOD) test split. 
From the remaining 26 topics, we further split the data into a training set and an in-domain test set. 
For more challenging evaluations, we manually curate a hard test set,
which includes more complex tool-use cases, such as those involving long tool parameters, easily confusable tools, parameter value updating, scenarios with a large number of tools, etc., as listed in Appendix \ref{appx:hard-cases}. 

\subsubsection{Metrics}

\paragraph{For Single-Tool Scenarios (S-S \& M-S).} 
For scenarios where there is just a single tool involved, we evaluate two metrics: \textbf{Tool Selection Accuracy (TS)} and \textbf{Parameter Selection Accuracy (PS)}.

\paragraph{For Multi-Turn Scenarios (M-S \& M-M).} 
In multi-turn dialogues, the following metrics are introduced: 
(1) \textbf{Success Rate (SR):} A binary metric where the entire dialogue is considered successful for tool-use: (=1) only if there are no errors throughout all turns; otherwise, it is considered unsuccessful (=0). 
(2) \textbf{Averaged Turn Success Rate (ATS):} We first evaluate each dialogue turn with the tool-use success rate (where each turn is marked as either 0 or 1), then average the binary scores with the total dialogue turns of a dialogue session. This score takes into account the finer-grained success rate of tool-use at the turn level. 
(3) \textbf{Soft Averaged Turn Success Rate (SATS):} This metric adjusts the ATS based on the proximity of errors to the current turn. Specifically: If a turn is incorrect, the score is 0. If a turn is correct, given $j$ as the index of this turn and $i$ as the index of the last incorrect turn, the score is $1$ when $j < i$ and $1 - e^{-(j-i)}$ when $j > i$. This design is based on the intuition that a closer incorrect turn can negatively impact subsequent turns. Moreover, the closer the turn becomes incorrect, the lower the overall accumulated score, even if the remaining turns are correct. 
(4) \textbf{Task Process Rate (TPR):} This is calculated as the ratio of the first incorrect turn to the total number of turns. This metric is included to capture how early in the dialogue the first mistake occurs, as earlier errors tend to disrupt the overall task flow more significantly.

% 写下motivation

\paragraph{For Multi-Tool Scenarios (S-M \& M-M).} 
For scenarios involving multiple tools, the following metrics are introduced:
(1) \textbf{Tool Number Accuracy (TN):} Denote the predicted tool list as ``Pred'' and the ground truth tool list as ``GT'', $\text{TN} = \mid \text{Pred} \cap \text{GT} \mid/\mid\text{Pred} \cup \text{GT}\mid$,
where $\mid \mid$ denotes the number of tools.
(2) \textbf{Tool Order Accuracy (TO):} This metric evaluates the correctness of the tool sequence, adjusted by a decay factor: $\text{TO} = t \times \mid \text{LCR(GT, Pred)} \mid/\mid \text{GT} \mid$, where LCR is the longest common subsequence, and $t$ is a decay coefficient calculated as: $t = \cos\left((\pi/2) \times (i/|\text{Pred}|\right)$, where $i$ is the starting position of the longest common subsequence. The value of $t$ ranges from 0 to 1, with a faster decay for positions later in the sequence.

These metrics offer a comprehensive and fine-grained evaluation of LLMs' tool-use capabilities. 
Unlike conventional approaches that focus only on overall success rates~\citep{qin2023toolllm}, our metrics account for the dynamics along dialogue turns and the dependencies between multiple tools. 
 We refer the reader to Appendix \ref{appx:example-metrics} for more examples of how to compute these metrics.

\section{Experiments}

% \subsection{Experimental Setup}

\paragraph{Experimental Setup.} 
% We assess the tool-use abilities of existing advanced LLMs using our MTU-Eval, including 2 close-sourced LLMs (i.e., GPT-3.5/GPT-4~\citep{gpt4}) and 12 open-sourced LLMs (i.e., LLaMA2-7B/70B~\citep{touvron2023llama}, LLaMA3-8B/70B~\citep{dubey2024llama3}, Qwen1.5-14B/72B~\citep{qwen}, Qwen2-7B/72B~\citep{yang2024qwen2}, Mistral-7B~\citep{jiang2023mistral}, ChatGLM3-6B~\citep{du2022glm}, ChatGLM4-9B~\citep{glm2024chatglm}). 
% We assess the tool-use abilities of existing advanced LLMs using our MTU-Eval. 
We evaluate 5 closed-source LLMs such as GPT-3.5~\citep{gpt4}, GPT-4~\citep{gpt4}, Qwen-Max~\citep{qwen1.5}, GLM-4-Plus\footnote{\url{https://bigmodel.cn/dev/api/normal-model/glm-4}} and DeepSeek2.5~\citep{deepseekv2}. 
We also evaluate numerous open-source LLMs such as LLaMA2~\citep{touvron2023llama} and LLaMA3~\citep{llama3modelcard} series, Qwen1.5~\citep{qwen1.5} and Qwen2~\citep{yang2024qwen2} series, Mistral~\citep{jiang2023mistral}, ChatGLM3 and GLM-4~\citep{glm2024chatglm, du2022glm, zeng2022glm} series. 
Then, we also compare 2 models specifically enhanced for tool-use: ToolLLaMA~\citep{qin2023toolllm} and our \textbf{MTU-LLaMA}, which isfine-tuned on MTU-Instruct based on LLaMA3-8B-Instruct. 
Note that all baselines are \textbf{instruction-tuned models}. 
We refer the readers to Appendix \ref{appx:mtu-eval-details} for more details on the hard cases, the prompt templates, and the metric computation. 

\subsection{Main Results}

\begin{table*}[t]
\caption{Results of different models on the \textbf{normal} set of MTU-Eval \textbf{(S-S \& M-S)}. ``S-S'' and ``M-S'' denote ``Single-Turn Single-Tool'' and ``Multi-Turn Single-Tool'' settings, respectively. We utilize \resultone{green} (1st), \resulttwo{blue} (2nd), and \resultthird{yellow} (3rd) backgrounds to distinguish the top three results within both open-source and tool-use-specific models. We employ \textbf{bold} and \underline{underlined} text to denote the top and second-best results across all model categories (same markers for the other tables). All the baselines are instruction-tuned models.}

\resizebox{1.0\textwidth}{!}{
% \begin{tabular}{c|ccc|ccc|ccc|ccc|c}
% \begin{tabular}{c|cccccccccccc|>{\columncolor[gray]{0.93}}c@{}}
\begin{tabular}{c|ccc|ccccccc}
\toprule
\multirow{2}{*}{\textbf{Models}} & \multicolumn{3}{c|}{\textbf{S-S}} & \multicolumn{7}{c}{\textbf{M-S}}\\
\cmidrule(lr){2-11} 

& \textbf{TS} & \textbf{PS}   &\textbf{Avg.}  & \textbf{TS}   & \textbf{PS}    & \textbf{ATS}     & \textbf{SATS}  & \textbf{SR}     & \textbf{TPR}        & \textbf{Avg.}   \\ 

\midrule
\multicolumn{11}{c}{\textit{Closed-Source Large Language Models}}\\ 
\midrule
GPT-4 &\textbf{95.83} & \underline{52.08} &\textbf{73.96}	&\textbf{88.10} 	&\textbf{74.49} 	&\textbf{73.67} 	&\textbf{67.36} 	&\textbf{29.63} 	&\textbf{45.35} 	&\textbf{63.10}   \\
GPT-3.5 &84.62 	&46.15 	&65.38 &69.05 &50.68 	&50.51 	&40.81 	&1.85 	&12.70 	&37.60 \\
Qwen-Max &91.67 	&50.00 	&70.83 &\underline{86.73} 	&66.67 	&64.96 	&57.88 	&20.37 	&35.53 	&55.36 \\
GLM-4-Plus &\textbf{95.83} 	&50.00 &\underline{72.92}  &85.03 & \underline{71.43} 	&\underline{72.01} 	& \underline{65.19} 	&\underline{24.07} 	&\underline{44.18} 	&\underline{60.32}    \\
DeepSeek V2.5 &\underline{93.75} 	&43.75 	&68.75 	&86.39 &69.39 &68.09 &60.20 &18.52 &38.04 &56.77 \\ 
\midrule
\multicolumn{11}{c}{\textit{Open-Source Large Language Models}} \\ \midrule
LLaMA2-7B &15.38 	&3.85 	&9.62 	&33.67 	&28.91 	&26.78 	&20.67 	&0.00 	&7.12 	&19.53 \\
LLaMA2-70B &70.59 	&33.33 	&47.28 &47.28 	&30.95 	&30.18 	&23.78 &0.00 	&9.82 	&23.67   \\ \midrule
LLaMA3-8B &65.38 	&30.77 	&48.08 	&35.71 	&17.35 	&17.53 	&12.34 	&0.00 	&1.72 	&14.11   \\ 
LLaMA3-70B &\resultthird{86.54} 	& \resultone{\textbf{57.69}} 	& \resultone{72.12} &79.25 	&\resulttwo{61.90}  	& \resulttwo{62.18}	&\resulttwo{54.81} 	& \resulttwo{14.81} 	&\resulttwo{32.33} 	&\resulttwo{50.88}  \\ \midrule
Qwen1.5-14B &75.00 	&34.62 	&54.81 	&62.93 	&36.73 	&35.17 	&27.37 	&1.85 	&6.34 	&28.40   \\
Qwen1.5-72B &78.85 	&38.46 	&58.65 	&\resulttwo{80.95} 	&\resultthird{61.22} 	&\resultthird{59.08} 	&\resultthird{50.88} 	&\resultone{16.67} 	&\resultthird{28.75} 	&\resultthird{48.75}   \\ \midrule
Qwen2-7B &73.08 	&38.46 &55.77 	&71.09 	&49.66 	&49.59 	&40.14 	&5.56 	&13.30 	&38.22  \\
Qwen2-72B &\resultthird{86.54} 	& \resultthird{48.08} 	& 67.31 	&\resultthird{79.93} 	&\resultthird{61.22} 	&58.52 	&50.28 	& \resulttwo{14.81} 	&25.15 	&48.32  \\ \midrule
% Mistral
Mistral-7B &60.42 	&25.00 	&42.71 	&61.22 	&37.07 	&37.33 	&29.14 	&3.70 	&9.98 	&29.74  \\ \midrule
ChatGLM3-6B &10.00 	&0.00 	&5.00 	&22.90 &5.99 	&5.79 	&3.66 	&0.00 	&0.00 	&6.39  \\
GLM-4-9B & \resulttwo{91.67} 	& 45.83 	&\resultthird{68.75} 	&63.95 	&42.18  	&42.72 	&35.98 	&3.70 	&19.01 	& 34.59\\ \midrule
% \midrule
\multicolumn{11}{c}{\textit{Tool-Use-Specific Large Language Models}} \\ \midrule
ToolLLaMA2-7B & 85.42 	&18.75 	&52.08 	&31.97 &7.82 &7.6 &5.20 &0.0 &5.73 &9.72  \\
MTU-LLaMA (Ours) & \resultone{92.31}	& \resulttwo{50.00}	& \resulttwo{71.15} & \resultone{81.63} 	& \resultone{67.69} 	&\resultone{66.94}	&\resultone{58.74} 	&\resultthird{9.26} 	&\resultone{32.47} 	&\resultone{52.79}   \\

\bottomrule
\end{tabular}}
\vspace{-2mm}
% Note that ``ZS'', ``ZS-COT'', ``FS'' represents ``zero-shot'', ``zero-shot w/ chain-of-thought'' and ``few-shot'', repsectively. Models are grouped roughly according to their model sizes.}
\vspace{-5mm}
\label{tab:singletool-normal}
\end{table*}

\begin{table*}[t]
\caption{Results of different models on the \textbf{normal} set of MTU-Eval \textbf{(S-M \& M-M)}. ``S-M'' and ``M-M'' denote ``Single-Turn Multi-Tool'' and ``Multi-Turn Multi-Tool'' settings, respectively.}
\vspace{-3mm}
\resizebox{1.0\textwidth}{!}{
% \begin{tabular}{c|ccc|ccc|ccc|ccc|c}
% \begin{tabular}{c|cccccccccccc|>{\columncolor[gray]{0.93}}c@{}}
\begin{tabular}{c|ccc|ccccccc}
\toprule
\multirow{2}{*}{\textbf{Models}} & \multicolumn{3}{c|}{\textbf{S-M}} & \multicolumn{7}{c}{\textbf{M-M}}\\
\cmidrule(lr){2-11} 

& \textbf{TN} & \textbf{TO}   &\textbf{Avg.}  & \textbf{TN}   & \textbf{TO}    & \textbf{ATS}     & \textbf{SATS}  & \textbf{SR}     & \textbf{TPR}        & \textbf{Avg.}   \\ 

\midrule
\multicolumn{11}{c}{\textit{Closed-Source Large Language Models}} \\
\midrule
GPT-4 & \textbf{66.85} & \textbf{70.52} & \textbf{68.68}  & \textbf{72.10} 	& \textbf{73.38} 	& \textbf{68.77} 	& \textbf{66.07} 	& \textbf{30.95} 	& \textbf{59.52} 	& \textbf{61.80}     \\
GPT-3.5 &32.64 	&38.22 &35.43	&24.72 	&25.46 	&16.11 	&12.22 	&0.00 	&3.97  &13.75 \\
Qwen-Max &39.76 	&48.82 	&39.29 &\underline{62.00} 	&\underline{64.07} 	&\underline{56.27} 	&\underline{55.27} 	&  4.76 	&\underline{52.38} 	&\underline{49.13}  \\
GLM-4-Plus  &45.76 &48.48 	&47.12 &53.95 	&54.58 	&49.17 	&45.72 	&4.76 	&39.48 	&41.28 \\
DeepSeek V2.5 & \underline{56.88} 	& \underline{60.28} 	& \underline{58.58}  & 50.15 	& 51.79 	& 44.84 	& 41.26 	& \underline{7.14} 	& 34.72 	& 38.32 \\ 
\midrule
\multicolumn{11}{c}{\textit{Open-Source Large Language Models}} \\ \midrule
LLaMA2-7B &3.39 	&3.94 	&3.67 	&22.22 	&22.22 	&22.90 	&21.80 	&0.00 	& \resultthird{19.92} &19.92  \\
LLaMA2-70B  &6.82	&8.48 &7.65 &30.12	&30.49	&28.77	&28.77	&0.00	&\resulttwo{28.77} 	&\resultthird{28.77} \\ \midrule 
LLaMA3-8B &14.79 &20.30 &17.55 	&9.43 	&10.04 	&4.44 	&2.81 	&0.00 	&0.00 	&4.46   \\ 
LLaMA3-70B &26.85 	& \resultthird{32.68} 	&\resultthird{29.76} 	&\resultthird{33.60} 	&\resultthird{35.71} 	&26.94 	&23.82 	&0.00 	&17.86 	&22.99    \\ \midrule 
Qwen1.5-14B &22.12 	&28.22 	&25.17 	&27.78 	&28.67 	&21.07 	&19.04 	&0.00 	&14.88 	&18.57   \\
Qwen1.5-72B &\resultthird{28.04} 	& 30.60 	& 29.32 	&23.00 	&23.31 	& \resultone{52.94} 	& \resultone{51.86} 	&0.00 	&7.34 	&26.41   \\ \midrule 
Qwen2-7B   &24.52 	&29.59 	&27.05 	&21.04 	&22.75 	&15.24 	&11.52 	&0.00 	&4.76 	&12.55    \\ 
Qwen2-72B & \resulttwo{52.76} 	& \resultone{59.98} 	& \resulttwo{56.37} 	& \resultone{45.93} 	& \resultone{47.67} 	& \resulttwo{42.02} 	& \resulttwo{38.07} 	& \resultone{\underline{7.14}} 	& \resultone{29.76} 	& \resulttwo{29.76}   \\ \midrule 
Mistral-7B &14.21 	&18.22 	&16.22 	&10.15 	&11.11 	&5.44 	&3.66 	&0.00 	&0.60  &5.16 \\ \midrule 
ChatGLM3-6B &6.53 	&8.56 	&7.55 	&10.64 	&11.11 	&9.21 	&8.01 	&\resulttwo{2.38} 	&5.95 	&7.88  \\
GLM-4-9B  &23.64 	&27.58 	&25.61 	&16.17 	&16.45 	&9.48 	&6.13 	&0.00 	&0.00 	&8.04  \\
\midrule 
\multicolumn{11}{c}{\textit{Tool-Use-Specific Large Language Models}} \\ \midrule
ToolLLaMA2-7B &11.52 &11.52 &11.51 &4.07 	&4.07 	&2.78 	&2.34 	&0.00 	&1.59 &2.48  \\
MTU-LLaMA (Ours) & \resultone{55.39} 	& \resulttwo{58.55} 	& \resultone{56.97}  	& \resulttwo{42.47} 	& \resulttwo{43.42} 	&\resultthird{39.64} 	&\resultthird{32.50} 	& \resultone{\underline{7.14}} 	& 19.05 	& \resultone{30.70}    \\

\bottomrule
\end{tabular}}
\vspace{-2mm}
% Note that ``ZS'', ``ZS-COT'', ``FS'' represents ``zero-shot'', ``zero-shot w/ chain-of-thought'' and ``few-shot'', repsectively. Models are grouped roughly according to their model sizes.}
\vspace{-3mm}
\label{tab:multitool-normal}
\end{table*}

\paragraph{Overall Performance.} 
The experimental results for the normal set are presented in Table \ref{tab:singletool-normal} (S-S \& M-S) and Table \ref{tab:multitool-normal} (S-M \& M-M). 
The results on the hard set are illustrated in Table \ref{tab:hardset_avg}. 
These results reveal several key findings: 
(1) Open-source models typically exhibit inferior performance compared to closed-source models in nearly all metrics, with the exception of GPT-3.5. However, certain models, including LLaMA3-70B and Qwen2-72B, demonstrate results comparable to those achieved by closed-source models. 
(2) GPT-4 consistently exhibits superior performance on the normal set; however, its performance decreases on the hard set compared to GLM-4-Plus. Qwen-Max demonstrates exceptional performance in the M-M setting, with its advantages becoming more pronounced in the hard setting, even surpassing GPT-4. Similarly, GLM-4-Plus exhibits outstanding performance in the S-S and M-S settings, and its superiority is further amplified in the hard setting, also exceeding that of GPT-4. DeepSeek V2.5 performs admirably in the S-M setting. 
(3) Our MTU-LLaMA exhibits substantial advancements over its initialization, \textit{i.e.}, LLaMA3-8B-Instruct, across all settings and metrics. It is also competitive with some closed-source models, underscoring the effectiveness of our MTU-Instruct. 
(4) Generally, all the models perform better on the normal set than on the hard set, indicating LLMs' limitations in handling more challenging tool-use scenarios. 
(5) Notably, despite being fine-tuned specifically for tool-use, ToolLLaMA exhibits poor performance across all settings and metrics, suggesting its limited generalizability.

\paragraph{Effect of Multi-Turn.} 
We compare the single-turn (S-S, S-M) and multi-turn (M-S, M-M) settings across the Tables \ref{tab:singletool-normal}, \ref{tab:multitool-normal}, and \ref{tab:hardset_avg}, and have following findings: 
(1) Both closed-source and open-source models tend to perform worse in multi-turn settings (M-S and M-M) compared to single-turn settings (S-S and S-M). 
% This trend is observed in 
(2) Our MTU-LLaMA shows relatively better adaptation and robustness to multi-turn settings.
% but still exhibits some performance decline. 
(3) Based on our novel TPR metric, we can observe that LLMs typically experience tool-use errors within the initial 30\%-50\% turns for closed-source models, and within the first 0\%-30\% turns for open-source models. 
% (4) The ATS scores generally exceed the corresponding TPR scores by a significant margin, indicating that while models frequently encounter tool-use errors in the initial turns, they can still correctly use tools in subsequent turns in most cases. However, this correctness does not account for the cascading effect of previous errors, but solely considers the success rate of independent tool usage. 
(4) Most models such as Qwen2-72B have significantly higher ATS scores than TPR scores. This implies that while LLMs frequently encounter tool-use errors in the initial turns, they can still correctly use tools in subsequent turns in most cases. However, this correctness does not account for the cascading effect of previous errors, but solely considers the success rate of independent tool usage. 
(5) Fortunately, the SATS scores can be treated as an equilibrium between ATS and TPR metrics, which simultaneously account for the positions at which tool-use errors occur and the subsequent impact on later turns.

\begin{wraptable}[20]{r}{0.5\textwidth}
\vspace{-0.82cm}
\caption{Average scores on the \textbf{hard} set of MTU-Eval. (Detailed results are shown in Appendix \ref{appendex:hard_result}).}
\vspace{-3mm}
\resizebox{0.495\textwidth}{!}{
\begin{tabular}{c|c|c|c|c}
\toprule
\textbf{Models} & \textbf{S-S} & \textbf{M-S} & \textbf{S-M} & \textbf{M-M} \\ 
\midrule
\multicolumn{5}{c}{\textit{Closed-Source Large Language Models}} \\  \midrule
GPT-4 &77.88  	&44.61  & \textbf{58.07} & \underline{41.36} \\
GPT-3.5 &41.96 	&30.86 &18.39 &11.87  \\
Qwen-Max &77.88  &42.11 &24.01  & \textbf{45.08} \\
GLM-4-Plus &\textbf{82.69}	& \textbf{47.61} &30.90 &39.53 \\
DeepSeek V2.5 & \underline{80.77} & \underline{44.94} & \underline{40.01} &30.62 \\
\midrule
\multicolumn{5}{c}{\textit{Open-Source Large Language Models}} \\  \midrule
LLaMA2-7B &28.57 	&17.13   &2.35 &11.76 \\
LLaMA2-70B &28.57 	&23.46  &1.74 &16.79 \\\midrule
LLaMA3-8B &25.89 	&12.85  &9.91 &5.89 \\ 
LLaMA3-70B &\resultone{71.43} 	& \resulttwo{40.40}  & \resultthird{20.67} & \resultthird{20.56} \\\midrule
Qwen1.5-14B &44.64 	&29.39 &12.81 &9.37 \\
Qwen1.5-72B &56.73 &29.92 &18.85 &17.93 \\\midrule
Qwen2-7B &\resultthird{58.93} &28.73 &17.50 &10.17 \\
Qwen2-72B &\resulttwo{68.40} &\resultthird{38.42} & \resulttwo{37.14} & \resultone{25.13} \\\midrule
Mistral-7B &26.92 &26.04 &11.48 &10.84 \\\midrule
ChatGLM3-6B &9.09 &5.57  &18.89 &9.52 \\
GLM-4-9B &47.12 &30.22 & 18.98 &9.52\\
% \midrule
\midrule
\multicolumn{5}{c}{\textit{Tool-Use-Specific Large Language Models}} \\
\midrule
ToolLLaMA2-7B &18.27 	&10.19  &0.51 &2.34\\
MTU-LLaMA (Ours) &37.5 	&\resultone{43.10}   & \resultone{39.31} & \resulttwo{24.70} \\
\bottomrule
\end{tabular}}

\vspace{-10mm}
\label{tab:hardset_avg}
\end{wraptable}
 % \vspace{-10pt}
% % Draft for Multi-Turn before 2024.09.30, written by Yanan Wu
% % 分析多轮场景下的指标
% %不同模型的多轮指标相较于单轮指标要低一些。
% %SR惨淡
% We analyze the multi-turn metric and find that: (1) Observing the metrics of the M-S and M-M reveals that most LLMs perform worse in multi-turn compared to single-turn on the same metrics. For example, in the case of GPT-4, the M-S-normal scores are 7.73\% lower than the S-S-normal scores for the TS. This may be attributes to the challenges introduced by multi-turn scenarios, which involve more complex dialogue understanding and long context understanding. (2) By comparing the differences among various metrics, we observe that the SR scores are consistently low across nearly all models. Even GPT-4 achieves only 29.63\% on M-S-normal and 0 on M-S-hard. Compared to other metrics, SR serves as a stricter measure of the completion rate for models in task performance. To earn a score on a given dialogue, LLMs must successfully execute tool calls in all rounds. The low SR scores indicate the instability of LLMs' tool using capabilities in multi-turn scenarios.

\paragraph{Effect of Multi-Tool.} 
Based on multi-tool settings (S-M and M-M) across Tables \ref{tab:singletool-normal}, \ref{tab:multitool-normal}, and \ref{tab:hardset_avg}, we derive the following findings: 
(1) Multi-tool settings (S-M and M-M) show significant complexity, leading to noticeable performance drops for most models. 
(2) Despite the complexity, models like GPT-4, Qwen2-72B and our MTU-LLaMA show stronger robustness. 
(3) In contrast, the good models in single-tool settings such as GLM-4-Plus (closed-source) and LLaMA3-70B (open-source), are surpassed by Qwen-Max (closed-source) and Qwen2-72B (open-source), respectively, indicating the superior performance of Qwen series in multi-tool settings. 
(4) The model rankings by TN and TO are highly consistent, implying that models with better control over the number of tools also tend to manage tool sequences effectively, suggesting a strong correlation between these capabilities.

% % Draft for Multi-Tool before 2024.09.30, written by Yanan Wu
% % 分析多工具场景下的指标
% %不同模型的表现参差不齐
% %our model
% From the results in Table \ref{tab:multitool-normal} and \ref{tab:multitool-hard}, we find that different models exhibit significant performance differences under multi-tool scenario. For example, even the open-source model LLaMA3-70B scores only 26.85 and 32.68 on TN and TO, respectively. In comparison, GPT-4 outperforms it by 40.00 and 37.84. This highlights the limitations of open-source models in multi-tool scenarios.

% \begin{figure}[t]
%     \centering
%     \includegraphics[width=1.0\linewidth]{source/models_acc_overall.png}
%     \vspace{-5mm}
%     \caption{Overall Acc (Reference).}
%     \label{fig: dataset_domain_dis_chart}
%     \vspace{-3mm}
% \end{figure}
% \begin{figure}[t]
%     \centering
%     \includegraphics[width=1.0\linewidth]{source/models_acc_dis.png}
%     \vspace{-5mm}
%     \caption{Scene Acc (Reference).}
%     \label{fig: dataset_domain_dis_chart}
%     \vspace{-3mm}
% \end{figure}

\subsection{Analysis}
% 分析维度文档：https://alidocs.dingtalk.com/i/nodes/a9E05BDRVQRkezKGCjq0zPYMJ63zgkYA?utm_scene=person_space
\paragraph{OOD Performance.} To evaluate the generality of MTU-LLaMA, we measure its performance on the OOD test split of MTU-Bench and two other OOD tool-use benchmarks, \textit{i.e.}, API-Bank~\citep{li2023apibank} and ToolTalk~\citep{farn2023tooltalk}. in Table \ref{tab:generalization}, we compare the performance of MTU-LLaMA, LLaMA3-8B-Instruct, and GPT-4 on these benchmarks under the M-S setting,
and We observe that MTU-LLaMA outperforms LLaMA3-8B-Instruct on all three OOD benchmarks. 
% Specifically, MTU-LLaMA improves the average metrics by 9.6\% on MTU-Bench (OOD), 33.95\% on API-Bank, and 3.57\% on ToolTalk. 
Notably, MTU-LLaMA achieves performance comparable to that of GPT-4 on API-Bank,
which show strong generalizability of MTU-LLaMA.

\begin{wraptable}[13]{r}{0.5\textwidth}
\vspace{-0.6cm}
\caption{OOD Performance of our MTU-LLaMA.}
\resizebox{0.495\textwidth}{!}{
\begin{tabular}{c|ccccccc}
\toprule
\multirow{2}{*}{\textbf{Models}} & \multicolumn{7}{c}{\textbf{MTU-Bench (OOD)}}\\
\cmidrule(lr){2-8} 
 & \textbf{TS}   & \textbf{PS}    & \textbf{ATS}     & \textbf{SATS}  & \textbf{SR}     & \textbf{TPR}        & \textbf{Avg.}   \\ 
\midrule
GPT-4 &67.28	&67.57	&65.19	&47.23	&32.31	&37.72	&52.88  \\
LLaMA3-8B &36.71	&36.71	&35.53	&23.00	&7.69	&11.31	&25.16  \\ 
MTU-LLaMA &47.34 	&37.88	&62.05 	&54.58 &9.26 	&35.49 	&41.10  \\ \hline
\multirow{2}{*}{\textbf{Models}} & \multicolumn{7}{c}{\textbf{ToolTalk}}\\
\cmidrule(lr){2-8} 
 & \textbf{TS}   & \textbf{PS}    & \textbf{ATS}     & \textbf{SATS}  & \textbf{SR}     & \textbf{TPR}        & \textbf{Avg.}   \\ 
\midrule
GPT-4 &51.08 &51.60	&44.90	&39.74	&6.90	&27.13	&36.89\\
LLaMA3-8B &30.00	&30.86	&26.72	&22.66	&0.00	&15.44	 &20.95\\ 
MTU-LLaMA  & 30.25	 &31.23	 &30.87	 &28.51	 &3.45	 &22.82	 &24.52 \\ \hline
\multirow{2}{*}{\textbf{Models}} & \multicolumn{7}{c}{\textbf{API-Bank}}\\
\cmidrule(lr){2-8} 
 & \textbf{TS}   & \textbf{PS}    & \textbf{ATS}     & \textbf{SATS}  & \textbf{SR}     & \textbf{TPR}        & \textbf{Avg.}   \\ 
\midrule
GPT-4 &48.56	&48.56	&45.56	&44.59	&38.32	&41.10	&44.45  \\
LLaMA3-8B &14.03 &14.03	&13.86	&11.94	&7.74	&8.63	&11.71  \\ 
MTU-LLaMA &51.80	&51.80	&48.58	&45.57	&38.10	&38.10	&45.66  \\
\bottomrule
\end{tabular}}
\vspace{-2mm}
\label{tab:generalization}
\end{wraptable}

% \begin{figure}[]
%     \centering
%     \begin{subfigure}[b]{0.455\linewidth}
%         \centering
%         \includegraphics[width=\textwidth]{figures/relationships/relationship_turn.pdf}
%         \caption{Effect of Dialogue Turns.}
%         \label{fig:relationship_turn}
%     \end{subfigure}
%     \hfill
%     \begin{subfigure}[b]{0.455\linewidth}
%         \centering
%         \includegraphics[width=\textwidth]{figures/relationships/relationship_tool_number.pdf}
%         \caption{Effect of Tool Numbers.}
%         \label{fig:relationship_tool_number}
%     \end{subfigure}
%     \caption{Effect of Dialogue Turns and Tool Numbers.}
%     \label{fig:metric_relation}
% \end{figure}

\begin{figure}[t]
    \centering
    \resizebox{\textwidth}{!}{
    \includegraphics{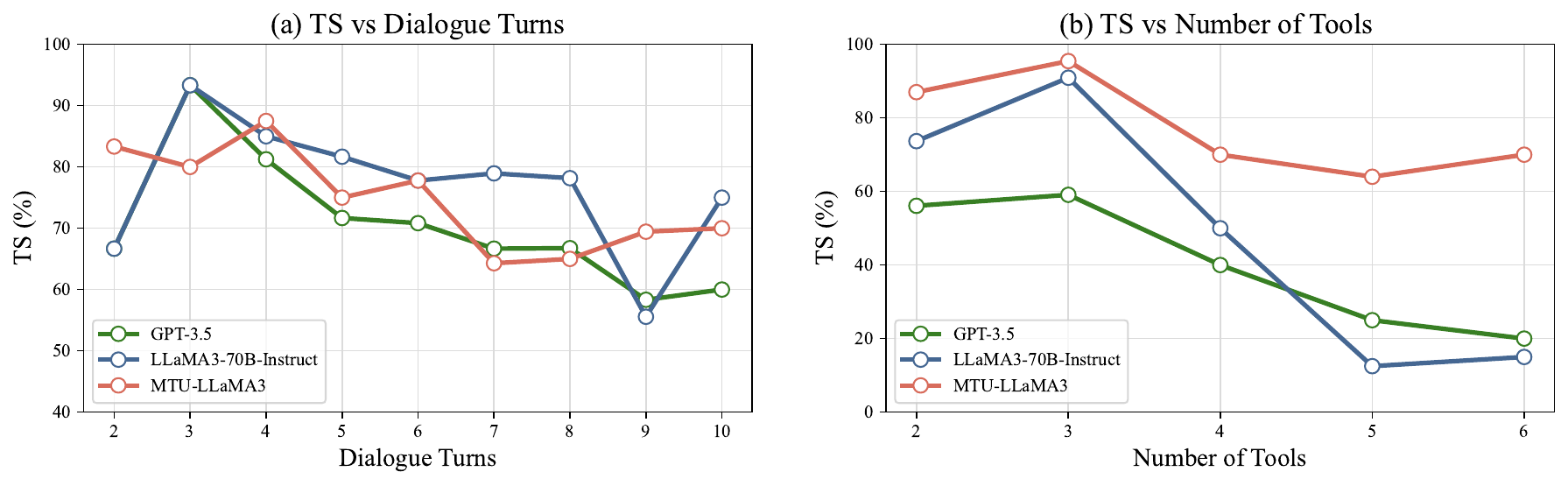}
    }
    \vspace{-0.7cm}
    \caption{(a) Effect of dialogue turns. (b) Effect of different number of tools.}
    \vspace{-4mm}
    \label{fig:metric_relation}
\end{figure}

% \begin{figure}[t]
%     \centering
%     \resizebox{1.0\linewidth}{!}{
% 	\begin{minipage}{0.455\linewidth}
% 		\vspace{3pt}
% 		\centerline{\includegraphics[width=\textwidth]{figures/model_size/model_size-ST.pdf}}
% 		\centerline{(a) For S-S and S-M}
% 	\end{minipage}
 
% 	\begin{minipage}{0.455\linewidth}
% 		\vspace{3pt}
% 		\centerline{\includegraphics[width=\textwidth]{figures/model_size/model_size-MT.pdf}} 
% 		\centerline{(b) For M-S and M-M}
% 	\end{minipage}}
 
% 	\caption{Comparative performance analysis of MTU-LLaMA across various model sizes}
% 	\label{fig:model_size}
% \end{figure}

\paragraph{Effect of Dialogue Turns.} 
We illustrate the impact of dialogue turns on tool selection accuracy in Figure \ref{fig:metric_relation}. We observe that performance slightly declines as the number of dialogue turns increases. Our MTU-LLaMA exhibits the most gradual decrease in performance, demonstrating its robustness to higher dialogue turns. 

% Draft "Effect of Dialogue Turns." before 2024.09.30
% In Figure \ref{fig: metric_relation}(a), we illustrate the relationship between TS Acc and the dialogue round to explore the robustness of models in multi-turn dialogue scenarios. It shows that as the number of the round increases, both GPT-3.5 and LLaMA3-70B exhibit accuracy various decline. Specifically, TS for GPT-3.5 decreases by an average of 5.83\% when the round increases from 3 to 9, while LLaMA3-70B experiences a decrease of 6.29\%. It highlights the significant limitations of current models in multi-turn dialogues. In comparison, MTU-LLaMA shows a more gradual decline, with an average decrease of only 1.76\%. It demonstrates that MTU-LLaMA exhibits greater robustness and stability in multi-turn dialogue.

\paragraph{Effect of Tool Numbers.} 
Figure \ref{fig:metric_relation} shows the impact of tool numbers on tool selection accuracy. As the number of tools increases, both GPT-3.5 and LLaMA3-70B experience notable declines in performance, with LLaMA3-70B showing a sharper drop. In contrast, MTU-LLaMA maintains relatively stable accuracy, demonstrating its superior handling of multiple tool calls.

% Draft "Effect of Tool Numbers." before 2024.09.30
% In Figure 4(b), we show the relationship between the number of tools and TN Accuracy for S-M, shedding light on the robustness of models in multi-tool calls. The results reveal that both GPT-3.5 and LLaMA3-70B experience significant declines in accuracy as the number of tools increases. Specifically, LLaMA3-70B shows a 75.42\% drop in accuracy when comparing the use of 3 tools to 5 tools, while GPT-3.5 exhibits a decline of 34.09\%. These findings highlight substantial limitations in both models when faced with multi-tool invocations. In contrast, MTU-LLaMA consistently maintains a TN Accuracy above 60\%, even when the number of tools exceeds 3. It indicates MTU-LLaMA's superior capability in effectively handling multi-tool calls.
\begin{wrapfigure}{l}{0.5\textwidth}
    \centering
    \includegraphics[width=0.5\textwidth]{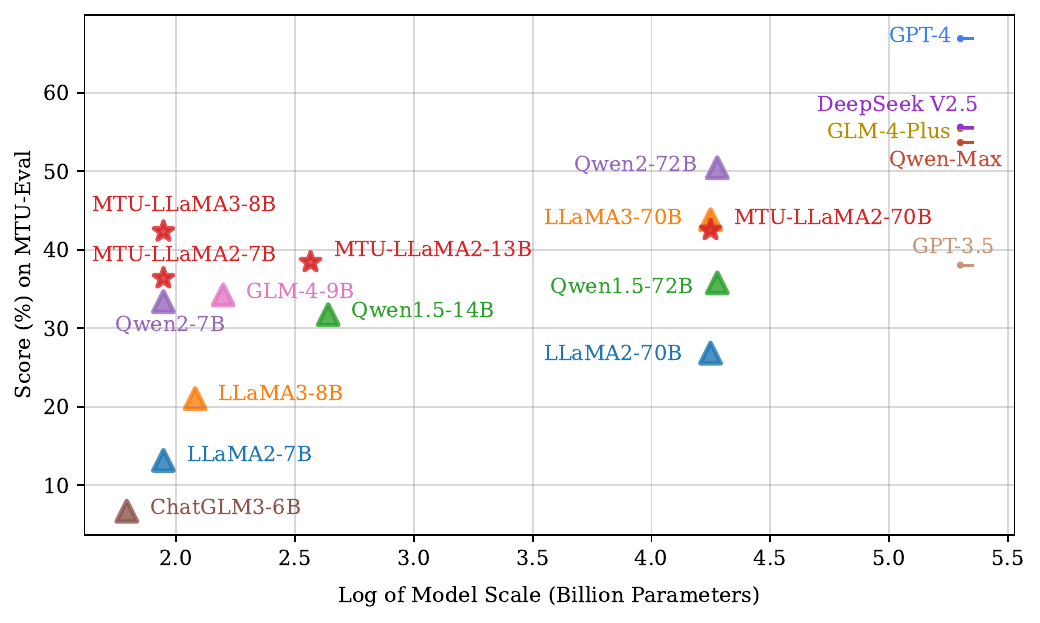}
    \vspace{-5mm}
    \caption{Scaling Law of LLMs on MTU-Eval.}
    \label{fig:scaling}
    \vspace{-2mm}
\end{wrapfigure}

\begin{figure}[t]
    \centering
    \resizebox{\textwidth}{!}{
    \includegraphics{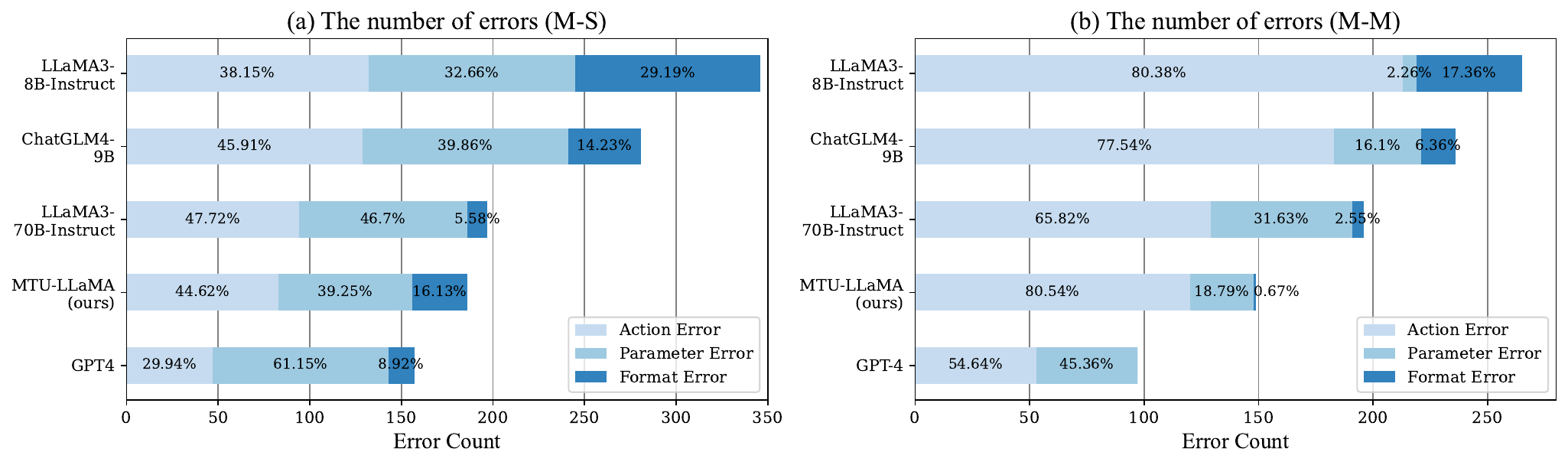}
    }
    \vspace{-0.7cm}
    \caption{Analysis of different error types for different LLMs under M-S and M-M settings.}
    \vspace{-6mm}
    \label{fig:error_type_m-m}
\end{figure}

\paragraph{Error Analysis.} % 错误归因
% 1. 整体上最大的错误类型为工具错误，其中工具错误占比51.64%，参数错误占比37.07%。对于单工具来说，Miss错误更多。对于多工具来说，Excessive错误更多。多轮多工具场景上工具错误严重，占比74.02%。
% 2. 除M-M场景外，最大的错误类型为参数错误。例如，表7中单轮中常见参数遗漏、参数重写。表x中表明多轮中独有的参数幻觉、参数继承问题。
% 3. fomat问题, 模型越弱，format问题越多。LLaMA3-8B问题最严重，LLaMA3-70B则缓解很多，我们的模型的模型几乎无格式问题。
% 4. 我们的模型：工具错误和格式错误上有显著缓解(e.g. M-M场景，213->120, 46->1)，在参数错误上略高(6->28)，这是因为，只有action&format正确才有参数是否正确的考虑。
In Figure~\ref{fig:error_type_m-m},
we use five LLMs to analyze the different types of errors (i.e., ``Action Error'', ``Parameter Error'' and ``Format Error'').
Specifically,
 the ``Action Error'' and ``Parameter Error'' denotes to select wrong tools and wrong parameters,
 respectively,
 and the ``Format Error'' means that the model cannot follow instructions well and outputs wrong formats,
 which cannot be resolved well.
 Figure~\ref{fig:error_type_m-m} illustrates two primary findings: 
 (1) ``Action Error'' occurs more often than other errors, specifically in challenge M-M setting,
 and stronger models show fewer errors.
 (2) ``Format Error'' usually exists in weaker LLMs (e.g., LLaMA3-8B-Instruct and ChatGLM4-9B),
 and our fine-tuned version MTU-LLaMA greatly reduces format issues,
 which shows the effectiveness of MTU-Instruct. 
 See Appendix \ref{appx:error-analysis} for detailed error cases.
 
% \begin{figure}[]
%     \centering
%     \begin{subfigure}[b]{0.455\linewidth}
%         \centering
%         \includegraphics[width=\textwidth]{figures/error_count/wrong_types_M-S.pdf}
%         %\caption{Effect of Dialogue Turns.}
%         \label{fig:error_type_m-s}
%     \end{subfigure}
%     \hfill
%     \begin{subfigure}[b]{0.455\linewidth}
%         \centering
%         \includegraphics[width=\textwidth]{figures/error_count/wrong_types_M-M.pdf}
%         %\caption{Effect of Tool Numbers.}
%         \label{fig:relationship_tool_number}
%     \end{subfigure}
%     \caption{Comparison of Error Type Counts and Rates Across Five Different Models}
%     \label{fig:error_type_m-m}
% \end{figure}

% Table \ref{tab:error_counts} presents the frequency of error types in the models. Our analysis shows: a. Most models struggle more with missed tools than with excessive or incorrect selections. Notably, GLM-4-9B had 24 missed selections, making up 70\% of its errors. b. Parameter filling error is the most common across almost all models, likely due to the complexity of multi-tool parameter interactions. c. While stronger models have few format parsing errors, LLAMA3-8B has 28, highlighting its poor instruction ability. d. MTU-LLaMA outperforms LLAMA3-8B in both tool selection and format parsing errors. Although MTU-LLaMA has 49 parameter filling errors compared to LLAMA3-8B's 22, this stems from LLAMA3-8B's issues with initial tool selection. Consistent with the S-M, a majority of the errors are in missed tool selection and incorrect parameter filling. Parameter filling remains a challenging task, but the proportion of tool selection errors is noticeably increasing. We present the statistical results for S-S and M-M in the Appendix \ref{}. For some error types, we show cases in Appendix \ref{}.

% Error Case
% 多轮单工具：突出与S-S不同的错误类型
    % 重复调用：历史调用obs中有的信息，直接回答即可。
    % 参数幻觉：没有信息的参数易幻觉。多轮场景下，LLM容易脑补当前调用中不存在的信息。比如历史购票为当天，但预定酒店并没有说明离开时间，模型易脑补为1天。
    % 参数继承：有信息的参数忘记继承/更新。
    % TODO：描述需要再调整一下
% 单轮多工具：重点在工具顺序和数量上

\paragraph{Scaling Law} 
We evaluate the performance of MTU-LLaMA across different model sizes, using LLaMA2 models with 7B, 13B, and 70B parameters as initialization, which are fine-tuned with MTU-Instruct. The results in Figure \ref{fig:scaling} show that the performance of MTU-LLaMA improves as the model size increases, suggesting its scalability. 

% Draft "Scaling Law" (Effect of Model Sizes.)
% We investigate the performance of MTU-LLaMA models of different sizes. We select LLaMA2 with three model sizes, 7B, 13B, and 70B, as our base models, and fine-tune them using our synthesized data. Subsequently, we test the models on evaluation sets across S-S, S-M, M-S, and M-M. The results are shown in Figure \ref{fig:model_szie}. We can see that MTU-LLaMA can significantly enhance the performance of the base models across all sizes, indicating that MTU-LLaMA is applicable to base models of varying scales. Specially, as the model size increases, there is an upward trend in the performance of MTU-LLaMA across the four evaluation sets. It suggests that larger model capacity aids in capturing more complex patterns and provides better generalization ability.
% \vspace{-10mm}

\paragraph{Consistency Between Our Proposed Metrics and Human Evaluation.} 
To show the validity of our proposed metrics (SATS, TN, and TO), we evaluate the consistency between these novel metrics and the human evaluation results. 
We randomly sample 50 instances from the M-S subset for the SATS metric and 50 instances from the S-M subset for the TN and TO metrics to compare two models: GPT-3.5 and LLaMA3-8B. 
Human annotators are also asked to compare the two models. 
In Table \ref{tab:metrics}, we report the Pearson correlation coefficient between our metrics and human evaluation results, as well as the Pearson correlation among human annotators, which shows excellent consistency between these metrics and human evaluations. 
More details can be found
% human evaluation procedure is provided 
in Appendix~\ref{app:human}. 

% Draft "Consistency between our proposed metrics and human evaluation."
% To show the validity of our proposed metrics (SATS, TN, and TO), we conduct a consistency experiment between these metrics and human evaluation scores. Specifically, we evaluate two models (\textit{i.e.}, GPT-3.5 and LLaMA3-8B), and randomly sample 50 instances from M-S for SATS metric and 50 instances from S-M for TN and TO metrics, where the human annotators are used to score these predicted results (the detailed human evaluation procedure is shown in Appendix~\ref{app:human}). 
% Based on the human evaluation results,
% In Table \ref{tab:metrics}, we compute the Pearson correlation coefficient between these metrics and human evaluation results, as well as the Pearson correlation coefficient among human annotators, which shows excellent consistency between our metrics and human evaluation.

\begin{wraptable}{r}{0.5\textwidth}
    \centering
    \vspace{-3mm}
    \caption{Consistency for our proposed metrics.}
    \vspace{-2mm}
    \resizebox{0.5\textwidth}{!}{
    \begin{tabular}{c|c|cc}
        \toprule
        \multirow{2}{*}{Metric} & \multirow{2}{*}{Annotation Count} & \multicolumn{2}{c}{Consistency}                \\ \cline{3-4} 
                                &                                   & \multicolumn{1}{c|}{Metric-Human} & Human-Human \\ \midrule
        SATS                      & 50                                & \multicolumn{1}{c|}{0.8280}       & 0.9344      \\ \midrule
        TN                        & 50                                & \multicolumn{1}{c|}{0.8497}       & 0.9245      \\ \midrule
        TO                        & 50                                & \multicolumn{1}{c|}{0.8821}       & 0.9482      \\ \bottomrule
    \end{tabular}}
    
    \vspace{-3mm}
    \label{tab:metrics}
\end{wraptable}

% \end{tabular}

% Draft "Comparing Human Evaluation and MTU-Eval."
% To verify the effectiveness of our three new metrics (SATS, TN, and TO), we measures their consistency with GPT-4 assessments. We test two models, GPT-3.5 and Llama3-8b, using GPT-4 to provide pairwise scores, where 1 indicates GPT-3.5 wins, 0 means a tie, and -1 indicates Llama3-8b wins. We then compare GPT-4’s scores with those derived from new metrics and calculate their correlation coefficient. 

\section{Related Works}
\paragraph{Instruction Tuning for Tool Learning.}
The objective of tool learning is to equip large language models (LLMs) with human-like tool usage capabilities~\citep{yang2023foundation,shen2023hugginggpt,qu2024tool}.  
To achieve this, LLMs are typically fine-tuned with tool instruction data to improve their performance in tool planning, selection, calling, and response generation~\citep{toolformer,liang2023taskmatrixai,kong2023tptu}.
However, existing tool instruction datasets either have limitations in multi-turn dialogue and multi-tool usage scenarios, or are based on synthetic data, resulting in misalignment with real-world user needs~\citep{metatool,li2023apibank,patil2023gorilla-apibench,zhuang2023toolqa}. In this paper, we introduce MTU-Instruct, a large-scale instruction dataset to  improve LLMs' performance in diverse real-world tool-use scenarios.

\paragraph{Evaluation Benchmarks for Tool Learning.}
% Recently,
Many tool-use benchmarks have been proposed, but they still have many limitations.
Firstly, in Table~\ref{tab:dataset-comparison}, these benchmarks have limited capabilities to assess complex scenarios (e.g., multi-turn dialogues, multiple tools, and cross-domain tool generalization)~\citep{metatool,li2023apibank,patil2023gorilla-apibench,tang2023toolalpaca}.  
Secondly, some benchmarks rely excessively on GPT models, potentially leading to subjective and unstable results with heavy costs~\citep{qin2023toolllm,tang2023toolalpaca}.
Finally, existing assessments often overlook critical dimensions, such as the order of multi-tool invocation, the impact of erroneous calls on subsequent interactions, and the accuracy of tool parameter selection, resulting in evaluations lacking comprehensiveness and depth~\citep{zhuang2023toolqa,li2023apibank,patil2023gorilla-apibench}. 
In contrast, MTU-Bench not only includes extensive multi-turn dialogues and multiple tool scenarios but also introduces testing for OOD tool generalization. By employing automated evaluation and incorporating metrics like SATS, TN, and TO, MTU-Eval achieves a more 
% objective and granular 
comprehensive assessment.

\section{Conclusion}
% Large language models (LLMs) have displayed massive improvements in reasoning and decision-making skills and can hold natural conversations with users.
% Many recent works seek to augment LLM-based assistants with external tools
% so they can access private or up-to-date information and carry out actions on
% behalf of users.
% Recently, many tool-use benchmark datasets have been proposed.
% However, existing datasets have the following limitations:
% (1). Insufficient scenarios (e.g., ).
% (2). Extensive evaluation costs (e.g.,).
% To address these limitations,
In this work,
we propose a multi-granularity tool-use benchmark for LLMs called \textbf{MTU-Bench},
which consists of MTU-Instruct and MTU-Eval.
Specifically,
first,
the MTU-Instruct dataset is used to enhance the tool-use abilities of existing LLMs,
and the MTU-Eval with multiple tool-use scenes is applied to benchmark the tool-use abilities comprehensively.
Notably,
% For the ``fully automatic'' property,
all evaluation metrics of our MTU-Eval are based on the prediction results and the ground truth without using any GPT or human evaluation metrics.  
Moreover,
Comprehensive experimental results demonstrate the effectiveness of our MTU-Bench.
Finally, we hope MTU-Bench can guide developers and researchers in understanding the tool-use capabilities of LLMs and facilitate the growth of foundation models.
% \bibliography{iclr2024_conference}

% \newpage

\section*{Ethics}

In developing MTU-Bench and MTU-LLaMA, we recognize several ethical considerations that arise from the broader context of integrating tool-use capabilities into large language models (LLMs). As these models become more capable of interacting with real-world systems—such as those involving financial services, healthcare, and other critical domains—we must consider the potential risks associated with misuse. For instance, there is the possibility that LLMs could be exploited to access sensitive tools or manipulate information in ways that could harm individuals or organizations.

While our work aims to improve the accuracy and efficiency of tool use, we are mindful of the importance of ensuring that these technologies are deployed responsibly. We advocate for the implementation of robust safeguards, including transparency in decision-making processes, fairness in how tools are applied, and accountability in real-world usage. Furthermore, we encourage future research and development efforts to focus on mitigating potential biases and ensuring that these systems are secure and trustworthy when handling sensitive tasks.

\bibliography{references}

\begin{thebibliography}{40}
\providecommand{\natexlab}[1]{#1}
\providecommand{\url}[1]{\texttt{#1}}
\expandafter\ifx\csname urlstyle\endcsname\relax
  \providecommand{\doi}[1]{doi: #1}\else
  \providecommand{\doi}{doi: \begingroup \urlstyle{rm}\Url}\fi

\bibitem[AI@Meta(2024)]{llama3modelcard}
AI@Meta.
\newblock Llama 3 model card.
\newblock 2024.
\newblock URL \url{https://github.com/meta-llama/llama3/blob/main/MODEL_CARD.md}.

\bibitem[Bubeck et~al.(2023)Bubeck, Chandrasekaran, Eldan, Gehrke, Horvitz, Kamar, Lee, Lee, Li, Lundberg, Nori, Palangi, Ribeiro, and Zhang]{bubeck2023sparks-AGI}
Sébastien Bubeck, Varun Chandrasekaran, Ronen Eldan, Johannes Gehrke, Eric Horvitz, Ece Kamar, Peter Lee, Yin~Tat Lee, Yuanzhi Li, Scott Lundberg, Harsha Nori, Hamid Palangi, Marco~Tulio Ribeiro, and Yi~Zhang.
\newblock Sparks of artificial general intelligence: Early experiments with gpt-4.
\newblock March 2023.
\newblock URL \url{https://www.microsoft.com/en-us/research/publication/sparks-of-artificial-general-intelligence-early-experiments-/with-gpt-4/}.

\bibitem[Budzianowski et~al.(2018)Budzianowski, Wen, Tseng, Casanueva, Ultes, Ramadan, and Gašić]{budzianowski2018multiwoz}
Paweł Budzianowski, Tsung-Hsien Wen, Bo-Hsiang Tseng, Iñigo Casanueva, Stefan Ultes, Osman Ramadan, and Milica Gašić.
\newblock Multiwoz - a large-scale multi-domain wizard-of-oz dataset for task-oriented dialogue modelling.
\newblock \emph{arXiv preprint arXiv: 1810.00278}, 2018.

\bibitem[Byrne et~al.(2019)Byrne, Krishnamoorthi, Sankar, Neelakantan, Duckworth, Yavuz, Goodrich, Dubey, Cedilnik, and Kim]{byrne2019taskmaster1}
Bill Byrne, Karthik Krishnamoorthi, Chinnadhurai Sankar, Arvind Neelakantan, Daniel Duckworth, Semih Yavuz, Ben Goodrich, Amit Dubey, Andy Cedilnik, and Kyu-Young Kim.
\newblock Taskmaster-1: Toward a realistic and diverse dialog dataset.
\newblock \emph{arXiv preprint arXiv: 1909.05358}, 2019.

\bibitem[Cobbe et~al.(2021)Cobbe, Kosaraju, Bavarian, Chen, Jun, Kaiser, Plappert, Tworek, Hilton, Nakano, Hesse, and Schulman]{math-verifier}
Karl Cobbe, Vineet Kosaraju, Mohammad Bavarian, Mark Chen, Heewoo Jun, Lukasz Kaiser, Matthias Plappert, Jerry Tworek, Jacob Hilton, Reiichiro Nakano, Christopher Hesse, and John Schulman.
\newblock Training verifiers to solve math word problems.
\newblock \emph{arXiv preprint arXiv: Arxiv-2110.14168}, 2021.

\bibitem[DeepSeek-AI(2024)]{deepseekv2}
DeepSeek-AI.
\newblock Deepseek-v2: A strong, economical, and efficient mixture-of-experts language model, 2024.

\bibitem[Du et~al.(2022)Du, Qian, Liu, Ding, Qiu, Yang, and Tang]{du2022glm}
Zhengxiao Du, Yujie Qian, Xiao Liu, Ming Ding, Jiezhong Qiu, Zhilin Yang, and Jie Tang.
\newblock Glm: General language model pretraining with autoregressive blank infilling.
\newblock In \emph{Proceedings of the 60th Annual Meeting of the Association for Computational Linguistics (Volume 1: Long Papers)}, pp.\  320--335, 2022.

\bibitem[Dubey et~al.(2024)Dubey, Jauhri, Pandey, Kadian, Al-Dahle, Letman, Mathur, Schelten, Yang, Fan, Goyal, Hartshorn, Yang, Mitra, Sravankumar, Korenev, Hinsvark, Rao, Zhang, Rodriguez, Gregerson, Spataru, Roziere, Biron, Tang, Chern, Caucheteux, Nayak, Bi, Marra, McConnell, Keller, Touret, Wu, Wong, Ferrer, Nikolaidis, Allonsius, Song, Pintz, Livshits, Esiobu, Choudhary, Mahajan, Garcia-Olano, Perino, Hupkes, Lakomkin, AlBadawy, Lobanova, Dinan, Smith, Radenovic, Zhang, Synnaeve, Lee, Anderson, Nail, Mialon, Pang, Cucurell, Nguyen, Korevaar, Xu, Touvron, Zarov, Ibarra, Kloumann, Misra, Evtimov, Copet, Lee, Geffert, Vranes, Park, Mahadeokar, Shah, van~der Linde, Billock, Hong, Lee, Fu, Chi, Huang, Liu, Wang, Yu, Bitton, Spisak, Park, Rocca, Johnstun, Saxe, Jia, Alwala, Upasani, Plawiak, Li, Heafield, Stone, El-Arini, Iyer, Malik, Chiu, Bhalla, Rantala-Yeary, van~der Maaten, Chen, Tan, Jenkins, Martin, Madaan, Malo, Blecher, Landzaat, de~Oliveira, Muzzi, Pasupuleti, Singh, Paluri, Kardas, Oldham, Rita,
  Pavlova, Kambadur, Lewis, Si, Singh, Hassan, Goyal, Torabi, Bashlykov, Bogoychev, Chatterji, Duchenne, Çelebi, Alrassy, Zhang, Li, Vasic, Weng, Bhargava, Dubal, Krishnan, Koura, Xu, He, Dong, Srinivasan, Ganapathy, Calderer, Cabral, Stojnic, Raileanu, Girdhar, Patel, Sauvestre, Polidoro, Sumbaly, Taylor, Silva, Hou, Wang, Hosseini, Chennabasappa, Singh, Bell, Kim, Edunov, Nie, Narang, Raparthy, Shen, Wan, Bhosale, Zhang, Vandenhende, Batra, Whitman, Sootla, Collot, Gururangan, Borodinsky, Herman, Fowler, Sheasha, Georgiou, Scialom, Speckbacher, Mihaylov, Xiao, Karn, Goswami, Gupta, Ramanathan, Kerkez, Gonguet, Do, Vogeti, Petrovic, Chu, Xiong, Fu, Meers, Martinet, Wang, Tan, Xie, Jia, Wang, Goldschlag, Gaur, Babaei, Wen, Song, Zhang, Li, Mao, Coudert, Yan, Chen, Papakipos, Singh, Grattafiori, Jain, Kelsey, Shajnfeld, Gangidi, Victoria, Goldstand, Menon, Sharma, Boesenberg, Vaughan, Baevski, Feinstein, Kallet, Sangani, Yunus, Lupu, Alvarado, Caples, Gu, Ho, Poulton, Ryan, Ramchandani, Franco, Saraf,
  Chowdhury, Gabriel, Bharambe, Eisenman, Yazdan, James, Maurer, Leonhardi, Huang, Loyd, Paola, Paranjape, Liu, Wu, Ni, Hancock, Wasti, Spence, Stojkovic, Gamido, Montalvo, Parker, Burton, Mejia, Wang, Kim, Zhou, Hu, Chu, Cai, Tindal, Feichtenhofer, Civin, Beaty, Kreymer, Li, Wyatt, Adkins, Xu, Testuggine, David, Parikh, Liskovich, Foss, Wang, Le, Holland, Dowling, Jamil, Montgomery, Presani, Hahn, Wood, Brinkman, Arcaute, Dunbar, Smothers, Sun, Kreuk, Tian, Ozgenel, Caggioni, Guzmán, Kanayet, Seide, Florez, Schwarz, Badeer, Swee, Halpern, Thattai, Herman, Sizov, Guangyi, Zhang, Lakshminarayanan, Shojanazeri, Zou, Wang, Zha, Habeeb, Rudolph, Suk, Aspegren, Goldman, Molybog, Tufanov, Veliche, Gat, Weissman, Geboski, Kohli, Asher, Gaya, Marcus, Tang, Chan, Zhen, Reizenstein, Teboul, Zhong, Jin, Yang, Cummings, Carvill, Shepard, McPhie, Torres, Ginsburg, Wang, Wu, U, Saxena, Prasad, Khandelwal, Zand, Matosich, Veeraraghavan, Michelena, Li, Huang, Chawla, Lakhotia, Huang, Chen, Garg, A, Silva, Bell, Zhang, Guo,
  Yu, Moshkovich, Wehrstedt, Khabsa, Avalani, Bhatt, Tsimpoukelli, Mankus, Hasson, Lennie, Reso, Groshev, Naumov, Lathi, Keneally, Seltzer, Valko, Restrepo, Patel, Vyatskov, Samvelyan, Clark, Macey, Wang, Hermoso, Metanat, Rastegari, Bansal, Santhanam, Parks, White, Bawa, Singhal, Egebo, Usunier, Laptev, Dong, Zhang, Cheng, Chernoguz, Hart, Salpekar, Kalinli, Kent, Parekh, Saab, Balaji, Rittner, Bontrager, Roux, Dollar, Zvyagina, Ratanchandani, Yuvraj, Liang, Alao, Rodriguez, Ayub, Murthy, Nayani, Mitra, Li, Hogan, Battey, Wang, Maheswari, Howes, Rinott, Bondu, Datta, Chugh, Hunt, Dhillon, Sidorov, Pan, Verma, Yamamoto, Ramaswamy, Lindsay, Lindsay, Feng, Lin, Zha, Shankar, Zhang, Zhang, Wang, Agarwal, Sajuyigbe, Chintala, Max, Chen, Kehoe, Satterfield, Govindaprasad, Gupta, Cho, Virk, Subramanian, Choudhury, Goldman, Remez, Glaser, Best, Kohler, Robinson, Li, Zhang, Matthews, Chou, Shaked, Vontimitta, Ajayi, Montanez, Mohan, Kumar, Mangla, Ionescu, Poenaru, Mihailescu, Ivanov, Li, Wang, Jiang, Bouaziz,
  Constable, Tang, Wang, Wu, Wang, Xia, Wu, Gao, Chen, Hu, Jia, Qi, Li, Zhang, Zhang, Adi, Nam, Yu, Wang, Hao, Qian, He, Rait, DeVito, Rosnbrick, Wen, Yang, and Zhao]{dubey2024llama3}
Abhimanyu Dubey, Abhinav Jauhri, Abhinav Pandey, Abhishek Kadian, Ahmad Al-Dahle, Aiesha Letman, Akhil Mathur, Alan Schelten, Amy Yang, Angela Fan, Anirudh Goyal, Anthony Hartshorn, Aobo Yang, Archi Mitra, Archie Sravankumar, Artem Korenev, Arthur Hinsvark, Arun Rao, Aston Zhang, Aurelien Rodriguez, Austen Gregerson, Ava Spataru, Baptiste Roziere, Bethany Biron, Binh Tang, Bobbie Chern, Charlotte Caucheteux, Chaya Nayak, Chloe Bi, Chris Marra, Chris McConnell, Christian Keller, Christophe Touret, Chunyang Wu, Corinne Wong, Cristian~Canton Ferrer, Cyrus Nikolaidis, Damien Allonsius, Daniel Song, Danielle Pintz, Danny Livshits, David Esiobu, Dhruv Choudhary, Dhruv Mahajan, Diego Garcia-Olano, Diego Perino, Dieuwke Hupkes, Egor Lakomkin, Ehab AlBadawy, Elina Lobanova, Emily Dinan, Eric~Michael Smith, Filip Radenovic, Frank Zhang, Gabriel Synnaeve, Gabrielle Lee, Georgia~Lewis Anderson, Graeme Nail, Gregoire Mialon, Guan Pang, Guillem Cucurell, Hailey Nguyen, Hannah Korevaar, Hu~Xu, Hugo Touvron, Iliyan Zarov,
  Imanol~Arrieta Ibarra, Isabel Kloumann, Ishan Misra, Ivan Evtimov, Jade Copet, Jaewon Lee, Jan Geffert, Jana Vranes, Jason Park, Jay Mahadeokar, Jeet Shah, Jelmer van~der Linde, Jennifer Billock, Jenny Hong, Jenya Lee, Jeremy Fu, Jianfeng Chi, Jianyu Huang, Jiawen Liu, Jie Wang, Jiecao Yu, Joanna Bitton, Joe Spisak, Jongsoo Park, Joseph Rocca, Joshua Johnstun, Joshua Saxe, Junteng Jia, Kalyan~Vasuden Alwala, Kartikeya Upasani, Kate Plawiak, Ke~Li, Kenneth Heafield, Kevin Stone, Khalid El-Arini, Krithika Iyer, Kshitiz Malik, Kuenley Chiu, Kunal Bhalla, Lauren Rantala-Yeary, Laurens van~der Maaten, Lawrence Chen, Liang Tan, Liz Jenkins, Louis Martin, Lovish Madaan, Lubo Malo, Lukas Blecher, Lukas Landzaat, Luke de~Oliveira, Madeline Muzzi, Mahesh Pasupuleti, Mannat Singh, Manohar Paluri, Marcin Kardas, Mathew Oldham, Mathieu Rita, Maya Pavlova, Melanie Kambadur, Mike Lewis, Min Si, Mitesh~Kumar Singh, Mona Hassan, Naman Goyal, Narjes Torabi, Nikolay Bashlykov, Nikolay Bogoychev, Niladri Chatterji, Olivier
  Duchenne, Onur Çelebi, Patrick Alrassy, Pengchuan Zhang, Pengwei Li, Petar Vasic, Peter Weng, Prajjwal Bhargava, Pratik Dubal, Praveen Krishnan, Punit~Singh Koura, Puxin Xu, Qing He, Qingxiao Dong, Ragavan Srinivasan, Raj Ganapathy, Ramon Calderer, Ricardo~Silveira Cabral, Robert Stojnic, Roberta Raileanu, Rohit Girdhar, Rohit Patel, Romain Sauvestre, Ronnie Polidoro, Roshan Sumbaly, Ross Taylor, Ruan Silva, Rui Hou, Rui Wang, Saghar Hosseini, Sahana Chennabasappa, Sanjay Singh, Sean Bell, Seohyun~Sonia Kim, Sergey Edunov, Shaoliang Nie, Sharan Narang, Sharath Raparthy, Sheng Shen, Shengye Wan, Shruti Bhosale, Shun Zhang, Simon Vandenhende, Soumya Batra, Spencer Whitman, Sten Sootla, Stephane Collot, Suchin Gururangan, Sydney Borodinsky, Tamar Herman, Tara Fowler, Tarek Sheasha, Thomas Georgiou, Thomas Scialom, Tobias Speckbacher, Todor Mihaylov, Tong Xiao, Ujjwal Karn, Vedanuj Goswami, Vibhor Gupta, Vignesh Ramanathan, Viktor Kerkez, Vincent Gonguet, Virginie Do, Vish Vogeti, Vladan Petrovic, Weiwei Chu,
  Wenhan Xiong, Wenyin Fu, Whitney Meers, Xavier Martinet, Xiaodong Wang, Xiaoqing~Ellen Tan, Xinfeng Xie, Xuchao Jia, Xuewei Wang, Yaelle Goldschlag, Yashesh Gaur, Yasmine Babaei, Yi~Wen, Yiwen Song, Yuchen Zhang, Yue Li, Yuning Mao, Zacharie~Delpierre Coudert, Zheng Yan, Zhengxing Chen, Zoe Papakipos, Aaditya Singh, Aaron Grattafiori, Abha Jain, Adam Kelsey, Adam Shajnfeld, Adithya Gangidi, Adolfo Victoria, Ahuva Goldstand, Ajay Menon, Ajay Sharma, Alex Boesenberg, Alex Vaughan, Alexei Baevski, Allie Feinstein, Amanda Kallet, Amit Sangani, Anam Yunus, Andrei Lupu, Andres Alvarado, Andrew Caples, Andrew Gu, Andrew Ho, Andrew Poulton, Andrew Ryan, Ankit Ramchandani, Annie Franco, Aparajita Saraf, Arkabandhu Chowdhury, Ashley Gabriel, Ashwin Bharambe, Assaf Eisenman, Azadeh Yazdan, Beau James, Ben Maurer, Benjamin Leonhardi, Bernie Huang, Beth Loyd, Beto~De Paola, Bhargavi Paranjape, Bing Liu, Bo~Wu, Boyu Ni, Braden Hancock, Bram Wasti, Brandon Spence, Brani Stojkovic, Brian Gamido, Britt Montalvo, Carl
  Parker, Carly Burton, Catalina Mejia, Changhan Wang, Changkyu Kim, Chao Zhou, Chester Hu, Ching-Hsiang Chu, Chris Cai, Chris Tindal, Christoph Feichtenhofer, Damon Civin, Dana Beaty, Daniel Kreymer, Daniel Li, Danny Wyatt, David Adkins, David Xu, Davide Testuggine, Delia David, Devi Parikh, Diana Liskovich, Didem Foss, Dingkang Wang, Duc Le, Dustin Holland, Edward Dowling, Eissa Jamil, Elaine Montgomery, Eleonora Presani, Emily Hahn, Emily Wood, Erik Brinkman, Esteban Arcaute, Evan Dunbar, Evan Smothers, Fei Sun, Felix Kreuk, Feng Tian, Firat Ozgenel, Francesco Caggioni, Francisco Guzmán, Frank Kanayet, Frank Seide, Gabriela~Medina Florez, Gabriella Schwarz, Gada Badeer, Georgia Swee, Gil Halpern, Govind Thattai, Grant Herman, Grigory Sizov, Guangyi, Zhang, Guna Lakshminarayanan, Hamid Shojanazeri, Han Zou, Hannah Wang, Hanwen Zha, Haroun Habeeb, Harrison Rudolph, Helen Suk, Henry Aspegren, Hunter Goldman, Igor Molybog, Igor Tufanov, Irina-Elena Veliche, Itai Gat, Jake Weissman, James Geboski, James Kohli,
  Japhet Asher, Jean-Baptiste Gaya, Jeff Marcus, Jeff Tang, Jennifer Chan, Jenny Zhen, Jeremy Reizenstein, Jeremy Teboul, Jessica Zhong, Jian Jin, Jingyi Yang, Joe Cummings, Jon Carvill, Jon Shepard, Jonathan McPhie, Jonathan Torres, Josh Ginsburg, Junjie Wang, Kai Wu, Kam~Hou U, Karan Saxena, Karthik Prasad, Kartikay Khandelwal, Katayoun Zand, Kathy Matosich, Kaushik Veeraraghavan, Kelly Michelena, Keqian Li, Kun Huang, Kunal Chawla, Kushal Lakhotia, Kyle Huang, Lailin Chen, Lakshya Garg, Lavender A, Leandro Silva, Lee Bell, Lei Zhang, Liangpeng Guo, Licheng Yu, Liron Moshkovich, Luca Wehrstedt, Madian Khabsa, Manav Avalani, Manish Bhatt, Maria Tsimpoukelli, Martynas Mankus, Matan Hasson, Matthew Lennie, Matthias Reso, Maxim Groshev, Maxim Naumov, Maya Lathi, Meghan Keneally, Michael~L. Seltzer, Michal Valko, Michelle Restrepo, Mihir Patel, Mik Vyatskov, Mikayel Samvelyan, Mike Clark, Mike Macey, Mike Wang, Miquel~Jubert Hermoso, Mo~Metanat, Mohammad Rastegari, Munish Bansal, Nandhini Santhanam, Natascha
  Parks, Natasha White, Navyata Bawa, Nayan Singhal, Nick Egebo, Nicolas Usunier, Nikolay~Pavlovich Laptev, Ning Dong, Ning Zhang, Norman Cheng, Oleg Chernoguz, Olivia Hart, Omkar Salpekar, Ozlem Kalinli, Parkin Kent, Parth Parekh, Paul Saab, Pavan Balaji, Pedro Rittner, Philip Bontrager, Pierre Roux, Piotr Dollar, Polina Zvyagina, Prashant Ratanchandani, Pritish Yuvraj, Qian Liang, Rachad Alao, Rachel Rodriguez, Rafi Ayub, Raghotham Murthy, Raghu Nayani, Rahul Mitra, Raymond Li, Rebekkah Hogan, Robin Battey, Rocky Wang, Rohan Maheswari, Russ Howes, Ruty Rinott, Sai~Jayesh Bondu, Samyak Datta, Sara Chugh, Sara Hunt, Sargun Dhillon, Sasha Sidorov, Satadru Pan, Saurabh Verma, Seiji Yamamoto, Sharadh Ramaswamy, Shaun Lindsay, Shaun Lindsay, Sheng Feng, Shenghao Lin, Shengxin~Cindy Zha, Shiva Shankar, Shuqiang Zhang, Shuqiang Zhang, Sinong Wang, Sneha Agarwal, Soji Sajuyigbe, Soumith Chintala, Stephanie Max, Stephen Chen, Steve Kehoe, Steve Satterfield, Sudarshan Govindaprasad, Sumit Gupta, Sungmin Cho, Sunny
  Virk, Suraj Subramanian, Sy~Choudhury, Sydney Goldman, Tal Remez, Tamar Glaser, Tamara Best, Thilo Kohler, Thomas Robinson, Tianhe Li, Tianjun Zhang, Tim Matthews, Timothy Chou, Tzook Shaked, Varun Vontimitta, Victoria Ajayi, Victoria Montanez, Vijai Mohan, Vinay~Satish Kumar, Vishal Mangla, Vlad Ionescu, Vlad Poenaru, Vlad~Tiberiu Mihailescu, Vladimir Ivanov, Wei Li, Wenchen Wang, Wenwen Jiang, Wes Bouaziz, Will Constable, Xiaocheng Tang, Xiaofang Wang, Xiaojian Wu, Xiaolan Wang, Xide Xia, Xilun Wu, Xinbo Gao, Yanjun Chen, Ye~Hu, Ye~Jia, Ye~Qi, Yenda Li, Yilin Zhang, Ying Zhang, Yossi Adi, Youngjin Nam, Yu, Wang, Yuchen Hao, Yundi Qian, Yuzi He, Zach Rait, Zachary DeVito, Zef Rosnbrick, Zhaoduo Wen, Zhenyu Yang, and Zhiwei Zhao.
\newblock The llama 3 herd of models.
\newblock \emph{arXiv preprint arXiv: 2407.21783}, 2024.

\bibitem[Farn \& Shin(2023)Farn and Shin]{farn2023tooltalk}
Nicholas Farn and Richard Shin.
\newblock Tooltalk: Evaluating tool-usage in a conversation setting.
\newblock \emph{arXiv preprint arXiv:2311.10775}, 2023.

\bibitem[GLM et~al.(2024)GLM, Zeng, Xu, Wang, Zhang, Yin, Rojas, Feng, Zhao, Lai, Yu, Wang, Sun, Zhang, Cheng, Gui, Tang, Zhang, Li, Zhao, Wu, Zhong, Liu, Huang, Zhang, Zheng, Lu, Duan, Zhang, Cao, Yang, Tam, Zhao, Liu, Xia, Zhang, Gu, Lv, Liu, Liu, Yang, Song, Zhang, An, Xu, Niu, Yang, Li, Bai, Dong, Qi, Wang, Yang, Du, Hou, and Wang]{glm2024chatglm}
Team GLM, Aohan Zeng, Bin Xu, Bowen Wang, Chenhui Zhang, Da~Yin, Diego Rojas, Guanyu Feng, Hanlin Zhao, Hanyu Lai, Hao Yu, Hongning Wang, Jiadai Sun, Jiajie Zhang, Jiale Cheng, Jiayi Gui, Jie Tang, Jing Zhang, Juanzi Li, Lei Zhao, Lindong Wu, Lucen Zhong, Mingdao Liu, Minlie Huang, Peng Zhang, Qinkai Zheng, Rui Lu, Shuaiqi Duan, Shudan Zhang, Shulin Cao, Shuxun Yang, Weng~Lam Tam, Wenyi Zhao, Xiao Liu, Xiao Xia, Xiaohan Zhang, Xiaotao Gu, Xin Lv, Xinghan Liu, Xinyi Liu, Xinyue Yang, Xixuan Song, Xunkai Zhang, Yifan An, Yifan Xu, Yilin Niu, Yuantao Yang, Yueyan Li, Yushi Bai, Yuxiao Dong, Zehan Qi, Zhaoyu Wang, Zhen Yang, Zhengxiao Du, Zhenyu Hou, and Zihan Wang.
\newblock Chatglm: A family of large language models from glm-130b to glm-4 all tools, 2024.

\bibitem[Hemphill et~al.(1990)Hemphill, Godfrey, and Doddington]{Hemphill1990TheAS}
C.~T. Hemphill, J.~J. Godfrey, and G.~Doddington.
\newblock The atis spoken language systems pilot corpus.
\newblock In \emph{HLT}, 1990.

\bibitem[Huang et~al.(2024)Huang, Shi, Li, Fan, Wu, Zhang, Liu, Zhou, Wan, Gong, and Sun]{metatool}
Yue Huang, Jiawen Shi, Yuan Li, Chenrui Fan, Siyuan Wu, Qihui Zhang, Yixin Liu, Pan Zhou, Yao Wan, Neil~Zhenqiang Gong, and Lichao Sun.
\newblock Metatool benchmark for large language models: Deciding whether to use tools and which to use.
\newblock In \emph{The Twelfth International Conference on Learning Representations, {ICLR} 2024, Vienna, Austria, May 7-11, 2024}. OpenReview.net, 2024.
\newblock URL \url{https://openreview.net/forum?id=R0c2qtalgG}.

\bibitem[Jiang et~al.(2023)Jiang, Sablayrolles, Mensch, Bamford, Chaplot, Casas, Bressand, Lengyel, Lample, Saulnier, et~al.]{jiang2023mistral}
Albert~Q Jiang, Alexandre Sablayrolles, Arthur Mensch, Chris Bamford, Devendra~Singh Chaplot, Diego de~las Casas, Florian Bressand, Gianna Lengyel, Guillaume Lample, Lucile Saulnier, et~al.
\newblock Mistral 7b.
\newblock \emph{arXiv preprint arXiv:2310.06825}, 2023.

\bibitem[Kong et~al.(2023)Kong, Ruan, Chen, Zhang, Bao, Shi, Du, Hu, Mao, Li, et~al.]{kong2023tptu}
Yilun Kong, Jingqing Ruan, Yihong Chen, Bin Zhang, Tianpeng Bao, Shiwei Shi, Guoqing Du, Xiaoru Hu, Hangyu Mao, Ziyue Li, et~al.
\newblock Tptu-v2: Boosting task planning and tool usage of large language model-based agents in real-world systems.
\newblock \emph{arXiv preprint arXiv:2311.11315}, 2023.

\bibitem[Lee et~al.(2022)Lee, Gupta, Rastogi, Cao, Zhang, and Wu]{lee2022sgd-x}
Harrison Lee, Raghav Gupta, Abhinav Rastogi, Yuan Cao, Bin Zhang, and Yonghui Wu.
\newblock Sgd-x: A benchmark for robust generalization in schema-guided dialogue systems.
\newblock In \emph{Proceedings of the AAAI Conference on Artificial Intelligence}, volume~36, pp.\  10938--10946, 2022.

\bibitem[Li et~al.(2023)Li, Song, Yu, Yu, Li, Huang, and Li]{li2023apibank}
Minghao Li, Feifan Song, Bowen Yu, Haiyang Yu, Zhoujun Li, Fei Huang, and Yongbin Li.
\newblock Api-bank: A benchmark for tool-augmented llms, 2023.

\bibitem[Liang et~al.(2023)Liang, Wu, Song, Wu, Xia, Liu, Ou, Lu, Ji, Mao, Wang, Shou, Gong, and Duan]{liang2023taskmatrixai}
Yaobo Liang, Chenfei Wu, Ting Song, Wenshan Wu, Yan Xia, Yu~Liu, Yang Ou, Shuai Lu, Lei Ji, Shaoguang Mao, Yun Wang, Linjun Shou, Ming Gong, and Nan Duan.
\newblock Taskmatrix.ai: Completing tasks by connecting foundation models with millions of apis.
\newblock \emph{arXiv preprint arXiv: Arxiv-2303.16434}, 2023.

\bibitem[Lu et~al.(2023)Lu, Yuan, Yuan, Lin, Lin, Tan, Zhou, and Zhou]{lu2023instag}
Keming Lu, Hongyi Yuan, Zheng Yuan, Runji Lin, Junyang Lin, Chuanqi Tan, Chang Zhou, and Jingren Zhou.
\newblock \# instag: Instruction tagging for analyzing supervised fine-tuning of large language models.
\newblock In \emph{The Twelfth International Conference on Learning Representations}, 2023.

\bibitem[OpenAI(2023)]{gpt4}
OpenAI.
\newblock Gpt-4 technical report.
\newblock \emph{PREPRINT}, 2023.

\bibitem[Paranjape et~al.(2023)Paranjape, Lundberg, Singh, Hajishirzi, Zettlemoyer, and Ribeiro]{art-tool}
Bhargavi Paranjape, Scott Lundberg, Sameer Singh, Hannaneh Hajishirzi, Luke Zettlemoyer, and Marco~Tulio Ribeiro.
\newblock Art: Automatic multi-step reasoning and tool-use for large language models.
\newblock \emph{arXiv preprint arXiv: Arxiv-2303.09014}, 2023.

\bibitem[Patil et~al.(2023)Patil, Zhang, Wang, and Gonzalez]{patil2023gorilla-apibench}
Shishir~G. Patil, Tianjun Zhang, Xin Wang, and Joseph~E. Gonzalez.
\newblock Gorilla: Large language model connected with massive apis.
\newblock \emph{arXiv preprint arXiv: 2305.15334}, 2023.

\bibitem[Qin et~al.(2023{\natexlab{a}})Qin, Hu, Lin, Chen, Ding, Cui, Zeng, Huang, Xiao, Han, Fung, Su, Wang, Qian, Tian, Zhu, Liang, Shen, Xu, Zhang, Ye, Li, Tang, Yi, Zhu, Dai, Yan, Cong, Lu, Zhao, Huang, Yan, Han, Sun, Li, Phang, Yang, Wu, Ji, Liu, and Sun]{qin2023tool}
Yujia Qin, Shengding Hu, Yankai Lin, Weize Chen, Ning Ding, Ganqu Cui, Zheni Zeng, Yufei Huang, Chaojun Xiao, Chi Han, Y.~Fung, Yusheng Su, Huadong Wang, Cheng Qian, Runchu Tian, Kunlun Zhu, Shi Liang, Xingyu Shen, Bokai Xu, Zhen Zhang, Yining Ye, Bo~Li, Ziwei Tang, Jing Yi, Yu~Zhu, Zhenning Dai, Lan Yan, Xin Cong, Ya-Ting Lu, Weilin Zhao, Yuxiang Huang, Jun-Han Yan, Xu~Han, Xian Sun, Dahai Li, Jason Phang, Cheng Yang, Tongshuang Wu, Heng Ji, Zhiyuan Liu, and Maosong Sun.
\newblock Tool learning with foundation models.
\newblock \emph{ARXIV.ORG}, 2023{\natexlab{a}}.
\newblock \doi{10.48550/arXiv.2304.08354}.

\bibitem[Qin et~al.(2023{\natexlab{b}})Qin, Liang, Ye, Zhu, Yan, Lu, Lin, Cong, Tang, Qian, Zhao, Tian, Xie, Zhou, Gerstein, Li, Liu, and Sun]{qin2023toolllm}
Yujia Qin, Shihao Liang, Yining Ye, Kunlun Zhu, Lan Yan, Yaxi Lu, Yankai Lin, Xin Cong, Xiangru Tang, Bill Qian, Sihan Zhao, Runchu Tian, Ruobing Xie, Jie Zhou, Mark Gerstein, Dahai Li, Zhiyuan Liu, and Maosong Sun.
\newblock Toolllm: Facilitating large language models to master 16000+ real-world apis, 2023{\natexlab{b}}.

\bibitem[Qu et~al.(2024)Qu, Dai, Wei, Cai, Wang, Yin, Xu, and Wen]{qu2024tool}
Changle Qu, Sunhao Dai, Xiaochi Wei, Hengyi Cai, Shuaiqiang Wang, Dawei Yin, Jun Xu, and Ji-Rong Wen.
\newblock Tool learning with large language models: A survey.
\newblock \emph{arXiv preprint arXiv:2405.17935}, 2024.

\bibitem[Rastogi et~al.(2020{\natexlab{a}})Rastogi, Zang, Sunkara, Gupta, and Khaitan]{rastogi2020sgd}
Abhinav Rastogi, Xiaoxue Zang, Srinivas Sunkara, Raghav Gupta, and Pranav Khaitan.
\newblock Towards scalable multi-domain conversational agents: The schema-guided dialogue dataset.
\newblock In \emph{Proceedings of the AAAI Conference on Artificial Intelligence}, volume~34, pp.\  8689--8696, 2020{\natexlab{a}}.

\bibitem[Rastogi et~al.(2020{\natexlab{b}})Rastogi, Zang, Sunkara, Gupta, and Khaitan]{rastogi2020towards}
Abhinav Rastogi, Xiaoxue Zang, Srinivas Sunkara, Raghav Gupta, and Pranav Khaitan.
\newblock Towards scalable multi-domain conversational agents: The schema-guided dialogue dataset.
\newblock In \emph{Proceedings of the AAAI conference on artificial intelligence}, volume~34, pp.\  8689--8696, 2020{\natexlab{b}}.

\bibitem[Schick et~al.(2023)Schick, Dwivedi{-}Yu, Dess{\`{\i}}, Raileanu, Lomeli, Zettlemoyer, Cancedda, and Scialom]{toolformer}
Timo Schick, Jane Dwivedi{-}Yu, Roberto Dess{\`{\i}}, Roberta Raileanu, Maria Lomeli, Luke Zettlemoyer, Nicola Cancedda, and Thomas Scialom.
\newblock Toolformer: Language models can teach themselves to use tools.
\newblock \emph{CoRR}, abs/2302.04761, 2023.
\newblock \doi{10.48550/arXiv.2302.04761}.
\newblock URL \url{https://doi.org/10.48550/arXiv.2302.04761}.

\bibitem[Shalyminov et~al.(2020)Shalyminov, Sordoni, Atkinson, and Schulz]{shalyminov2020fast}
Igor Shalyminov, Alessandro Sordoni, Adam Atkinson, and Hannes Schulz.
\newblock Fast domain adaptation for goal-oriented dialogue using a hybrid generative-retrieval transformer.
\newblock In \emph{ICASSP 2020-2020 IEEE International Conference on Acoustics, Speech and Signal Processing (ICASSP)}, pp.\  8039--8043. IEEE, 2020.

\bibitem[Shen et~al.(2023)Shen, Song, Tan, Li, Lu, and Zhuang]{shen2023hugginggpt}
Yongliang Shen, Kaitao Song, Xu~Tan, Dongsheng Li, Weiming Lu, and Yueting Zhuang.
\newblock Hugginggpt: Solving ai tasks with chatgpt and its friends in huggingface.
\newblock \emph{arXiv preprint arXiv: Arxiv-2303.17580}, 2023.

\bibitem[Siddhant et~al.(2018)Siddhant, Goyal, and Metallinou]{snips}
Aditya Siddhant, Anuj Goyal, and A.~Metallinou.
\newblock Unsupervised transfer learning for spoken language understanding in intelligent agents.
\newblock \emph{AAAI Conference on Artificial Intelligence}, 2018.
\newblock \doi{10.1609/AAAI.V33I01.33014959}.
\newblock URL \url{https://arxiv.org/abs/1811.05370v1}.

\bibitem[Tang et~al.(2023)Tang, Deng, Lin, Han, Liang, and Sun]{tang2023toolalpaca}
Qiaoyu Tang, Ziliang Deng, Hongyu Lin, Xianpei Han, Qiao Liang, and Le~Sun.
\newblock Toolalpaca: Generalized tool learning for language models with 3000 simulated cases.
\newblock \emph{arXiv preprint arXiv:2306.05301}, 2023.

\bibitem[Team(2024)]{qwen1.5}
Qwen Team.
\newblock Introducing qwen1.5, February 2024.
\newblock URL \url{https://qwenlm.github.io/blog/qwen1.5/}.

\bibitem[Touvron et~al.(2023)Touvron, Martin, Stone, Albert, Almahairi, Babaei, Bashlykov, Batra, Bhargava, Bhosale, et~al.]{touvron2023llama}
Hugo Touvron, Louis Martin, Kevin Stone, Peter Albert, Amjad Almahairi, Yasmine Babaei, Nikolay Bashlykov, Soumya Batra, Prajjwal Bhargava, Shruti Bhosale, et~al.
\newblock Llama 2: Open foundation and fine-tuned chat models.
\newblock \emph{arXiv preprint arXiv:2307.09288}, 2023.

\bibitem[Wang et~al.(2023)Wang, Zhang, Yang, Shi, Zhou, Hao, Xiong, Li, Sim, Chen, Zhu, Yang, Nik, Liu, Lin, Wang, Liu, Chen, Xu, Liu, Guo, and Fu]{wang2023interactive}
Zekun Wang, Ge~Zhang, Kexin Yang, Ning Shi, Wangchunshu Zhou, Shaochun Hao, Guangzheng Xiong, Yizhi Li, Mong~Yuan Sim, Xiuying Chen, Qingqing Zhu, Zhenzhu Yang, Adam Nik, Qi~Liu, Chenghua Lin, Shi Wang, Ruibo Liu, Wenhu Chen, Ke~Xu, Dayiheng Liu, Yike Guo, and Jie Fu.
\newblock Interactive natural language processing.
\newblock \emph{arXiv preprint arXiv: 2305.13246}, 2023.
\newblock URL \url{https://arxiv.org/abs/2305.13246v1}.

\bibitem[Xu et~al.(2023)Xu, Hong, Li, Hu, Chen, and Zhang]{xu2023toolbench}
Qiantong Xu, Fenglu Hong, Bo~Li, Changran Hu, Zhengyu Chen, and Jian Zhang.
\newblock On the tool manipulation capability of open-source large language models.
\newblock \emph{arXiv preprint arXiv: 2305.16504}, 2023.

\bibitem[Yang et~al.(2024)Yang, Yang, Hui, Zheng, Yu, Zhou, Li, Li, Liu, Huang, et~al.]{yang2024qwen2}
An~Yang, Baosong Yang, Binyuan Hui, Bo~Zheng, Bowen Yu, Chang Zhou, Chengpeng Li, Chengyuan Li, Dayiheng Liu, Fei Huang, et~al.
\newblock Qwen2 technical report.
\newblock \emph{arXiv preprint arXiv:2407.10671}, 2024.

\bibitem[Yang et~al.(2023)Yang, Nachum, Du, Wei, Abbeel, and Schuurmans]{yang2023foundation}
Sherry Yang, Ofir Nachum, Yilun Du, Jason Wei, Pieter Abbeel, and Dale Schuurmans.
\newblock Foundation models for decision making: Problems, methods, and opportunities.
\newblock \emph{arXiv preprint arXiv:2303.04129}, 2023.

\bibitem[Yao et~al.(2023)Yao, Zhao, Yu, Du, Shafran, Narasimhan, and Cao]{yao2023react}
Shunyu Yao, Jeffrey Zhao, Dian Yu, Nan Du, Izhak Shafran, Karthik Narasimhan, and Yuan Cao.
\newblock React: synergizing reasoning and acting in language models (2022).
\newblock \emph{arXiv preprint arXiv:2210.03629}, 2023.

\bibitem[Zeng et~al.(2022)Zeng, Liu, Du, Wang, Lai, Ding, Yang, Xu, Zheng, Xia, et~al.]{zeng2022glm}
Aohan Zeng, Xiao Liu, Zhengxiao Du, Zihan Wang, Hanyu Lai, Ming Ding, Zhuoyi Yang, Yifan Xu, Wendi Zheng, Xiao Xia, et~al.
\newblock Glm-130b: An open bilingual pre-trained model.
\newblock \emph{arXiv preprint arXiv:2210.02414}, 2022.

\bibitem[Zhuang et~al.(2023)Zhuang, Yu, Wang, Sun, and Zhang]{zhuang2023toolqa}
Yuchen Zhuang, Yue Yu, Kuan Wang, Haotian Sun, and Chao Zhang.
\newblock Toolqa: A dataset for llm question answering with external tools.
\newblock \emph{arXiv preprint arXiv:2306.13304}, 2023.

\end{thebibliography}
\bibliographystyle{iclr2025_conference}

\newpage
\appendix

\section{Limitations} 
Despite the strengths of our proposed MTU-Bench, there are still several limitations to consider. 
First, although MTU-Bench incorporates a diverse range of real-world user instructions, it may not fully capture all potential edge cases or highly complex interactions that can occur in dynamic, real-world environments. 
Second, while MTU-Eval provides comprehensive fine-grained metrics for evaluating tool-use abilities, these metrics are based on predefined benchmarks, which may not account for every possible tool-use challenge in evolving real-world applications. 
Although MTU-LLaMA demonstrates strong generalization across various metrics and scenarios, further research is needed to explore its adaptability to increasingly complex and emerging tool functionalities in real-time settings. 
Future work should focus on broadening the benchmark's coverage and exploring more dynamic and complex real-world use cases.

\section{Details of MTU-Bench Construction}
\label{appx:mtu-bench-construction-details}

\paragraph{Prompt Templates for MTU-Bench Construction.} 
The construction procedures of MTU-Bench involves numerous prompt templates, as listed below, for tool making (tool name synthesis and tool parameter synthesis), thought synthesis, observation simulation, and data quality check.

\begin{tcolorbox}[colback=white!95!gray,colframe=gray!50!black,rounded corners, label={tool-name-synthesis}, title={Prompt Template (Tool Name Synthesis)}] 
\begin{lstlisting}[breaklines=true, xleftmargin=0pt, breakindent=0pt, columns=fullflexible] 
I will give you a dialogue, including historical dialogue information and current round dialogue information. Please help me determine whether Assistant needs to call an \"API\" to obtain specific information or perform certain operations in order to solve the user issue in the current round of conversation.
1. If yes, and Assistant's response shows that the user's query has been solved, then return me with \"api_name\" in \"tags\". The name of the API should be concise and easy to understand, such as search_restaurant, book_restaurant, etc. The API name should start with a verb and include the specific domain name. Related domains includes [domain_info], etc. Note that if multiple APIs need to be called, return me with \"api_name, api_name, ...\" in \"tags\".
2. If Assistant can provide the current response without calling the API, return me \"no need to call\"in \"tags\".
3. If Assistant needs to call an API but is unable to do so due to a lack of necessary information, inform me of \"lack of necessary information\" in \"tags\". For example, due to the lack of "restaurant name", the API for "book_restaurant" cannot be called.
4. If Assistant is only confirming existing information with the User, inform me of the \"confirmation information\" in \"tags\".
6. If some information in the historical conversation can help Assistant respond to the current issue without calling APIs, inform me that \"information already exists\" in \"tags\".
Please reply to me in JSON format: {\"Analysis\": str, \"tags\": str}.
\end{lstlisting} 
\end{tcolorbox}

\begin{tcolorbox}[colback=white!95!gray,colframe=gray!50!black,rounded corners, label={tool-parameter-synthesis}, title={Prompt Template (Tool Parameter Synthesis)}] 
\begin{lstlisting}[breaklines=true, xleftmargin=0pt, breakindent=0pt, columns=fullflexible] 
I will provide you with a conversation segment that includes both historical dialogue information and current round dialogue information. Based on this information, please help me determine whether, to solve the problem presented by the User in the current round of dialogue, the Assistant calls a specific API to obtain the necessary information or to perform related actions. Please respond according to the following guidelines:
1. If an API call is required and the Assistant's response solves the User's problem, please specify which API was called and reply to me in the following JSON format: \"Action\": \"api_name\". Also, provide the parameters required for calling that API in the format: \"Action Input\": {\"parameter_name\": \"value\", ...}".
2. If it is impossible to call an API due to missing necessary parameter information, please explain in the \"Thought\" section due to the absence of which parameters, which API cannot be called.
3. If answering the User's question does not require calling an API, please explain in the \"Thought\" section why there is no need to use a API.
4. Please strictly use the API names and parameter names I provide, and refrain from fabricating any. If the required parameter is not defined within our list, you are allowed to introduce new parameter names. Beyond this allowance, do not utilize any API names and parameter names that are beyond the specified range, to ensure consistency and accuracy.
5. Please include a section called \"Thought\" in your answer where you clearly and unambiguously demonstrate your thought process when solving or answering the question.
Below are the APIs and their parameters information you can use:
[apis_information]
Please reply in the following JSON format: {\"Thought\": \"str\", \"Action\": \"api_name\", \"Action Input\": {\"parameter_name\": \"value\", ...}}. If there is no information for \"Action\", \"Action Input\", or \"Thought\", please return an empty string.
\end{lstlisting} 
\end{tcolorbox}

\begin{tcolorbox}[colback=white!95!gray,colframe=gray!50!black,rounded corners, label={thought-synthesis}, title={Prompt Template (Thought Synthesis)}] 
\begin{lstlisting}[breaklines=true, xleftmargin=0pt, breakindent=0pt, columns=fullflexible] 
I will provide you with a multi-round conversation and API call information for each round. Your task is to add or verify the Assistants' thinking process of API call to the dialog.
Your need to return the modified conversation with API call results added to me, strictly following the following format: 
\nModified Dialogue:\n[Round 0]\nUser: xxx.\nThought: xxx.\nAction: xxx.\nAction Input: xxx.\nObservation: xxx.\nAssistant: xxxx. \n[Round 1]\nUser: xxx.\nThought: xxx.\nAssistant: xxx.\n[Round 2]...
That is to say, after each round of the user's question, the Assistant must think about whether to call the API and how to call it, and present this in \"Thought\".
For example:
Modified Dialogue:
[Round 0]
User: Find me some fantasy movies I can watch online, something directed by Neil Marshall.
Thought: The user is looking for fantasy movies directed by Neil Marshall to watch online. I need to query a database or service for movies that fit these criteria.
Action: FindMovies
Action Input: {'directed_by': 'Neil Marshall', 'genre': 'Fantasy'}
Observation: {'directed_by': 'Neil Marshall', 'genre': 'Fantasy', 'subtitles': 'False', 'title': 'Hellboy'}
Assistant: Are you interested in Hellboy?
[Round 1]
User: I would enjoy watching Hellboy.
Thought: The user has not expressed any request, so the tool is not invoked here.
Assistant: Shall I play the movie?
[Round 2]
User: Yes, begin playing the movie. Yes, play it without subtitles.
Thought: The user has confirmed they want to start watching the movie "Hellboy" and specifically requested to play it without subtitles. I'll proceed to execute the play command with subtitles disabled.
Action: PlayMovie
Action Input: {'subtitles': 'False', 'title': 'Hellboy'}
Observation: {'directed_by': 'Neil Marshall', 'genre': 'Fantasy', 'subtitles': 'False', 'title': 'Hellboy'}
Assistant: The movie is now playing.

Requirements:
1. After \"Action\", it shows which API to call. If not, it indicates that API calls are not required for this round. The parameters passed in by calling the API are displayed after \"Action Input\". You just need to supplement the API call return results for the rounds that have called the API, and place them after \"Observation\". \"Observation\" shows the return result of the API.
2. Please return the thinking process in \"Thought\".
5. I will provide you with reference APIs and parameter names.
6. Except for adding \"Thought\" to the round of API calls, all other information remains in its original content.
7. Please return the complete conversation.
The following is a list of APIs and their parameters for your reference:
[apis information]
\end{lstlisting} 
\end{tcolorbox}

\begin{tcolorbox}[colback=white!95!gray,colframe=gray!50!black,rounded corners, label={observation-simulation}, title={Prompt Template (Observation Simulation)}] 
\begin{lstlisting}[breaklines=true, xleftmargin=0pt, breakindent=0pt, columns=fullflexible] 
I will provide you with a multi-round conversation and API call information for each round. Your task is to add the return result of an API call to the dialog based on Assistant's response.
You need to return the modified conversation with API call results added to me, strictly following the following format: 
\nModified Dialogue:\n[Round 0]\nUser: xxx.\nThought: xxx.\nAction: xxx.\nAction Input: xxx.\nObservation: xxx.\nAssistant: xxxx. \n[Round 1]\nUser: xxx.\nThought: xxx.\nAssistant: xxx.\n[Round 2]...
For example:
Modified Dialogue:
[Round 0]
User: Find me some fantasy movies I can watch online, something directed by Neil Marshall.
Thought: The user is looking for fantasy movies directed by Neil Marshall to watch online. I need to query a database or service for movies that fit these criteria.
Action: FindMovies
Action Input: {'directed_by': 'Neil Marshall', 'genre': 'Fantasy'}
Observation: {'directed_by': 'Neil Marshall', 'genre': 'Fantasy', 'subtitles': 'False', 'title': 'Hellboy'}
Assistant: Are you interested in Hellboy?
[Round 1]
User: I would enjoy watching Hellboy.
Assistant: Shall I play the movie?
[Round 2]
User: Yes, begin playing the movie. Yes, play it without subtitles.
Thought: The user has confirmed they want to start watching the movie "Hellboy" and specifically requested to play it without subtitles. I'll proceed to execute the play command with subtitles disabled.
Action: PlayMovie
Action Input: {'subtitles': 'False', 'title': 'Hellboy'}
Observation: {'directed_by': 'Neil Marshall', 'genre': 'Fantasy', 'subtitles': 'False', 'title': 'Hellboy'}
Assistant: The movie is now playing.
Requirements:
1. After \"Action\", it shows which API to call. If not, it indicates that API calls are not required for this round. The parameters passed in by calling the API are displayed after \"Action Input\". You just need to supplement the API call return results for the rounds that have called the API, and place them after \"Observation\".
2. Please return the information in \"Observation\" in JSON format, for example: {\"parameter_name\": \"value\", \"parametername\": \"value\"...}. Specifically, in every round with \"Action\" and \"Action Input\", you should add an \"Observation\" after the \"Action Input\", which should fill in the information returned by the API. For example: \nAction: FindMovies \nAction Input: {'directed_by': 'Neil Marshall', 'genre': 'Fantasy'}\nObservation: {'directed_by': 'Neil Marshall', 'genre': 'Fantasy', 'subtitles': 'False', 'title': 'Hellboy'}.
3. If Assistant's response shows that it has not yet received the specific information returned by the API tool call, that is "\nObservation\": {\"error\": \"Time out.\"}
4. If the result returned by Assistant shows that no relevant information is found, the API call returns an empty dict, such as: "\nObservation\": {}.
5. I will provide you with reference APIs and parameter names. Ensure that all parameter names used are defined and should not be fabricated.
6. Except for adding API return results to the round of API calls, all other information remains in its original content.
7. Please return the complete conversation.
The following is a list of APIs and their parameters for your reference:
[apis information]
\end{lstlisting} 
\end{tcolorbox}

\begin{tcolorbox}[colback=white!95!gray,colframe=gray!50!black,rounded corners, label={data-quality-check}, title={Prompt Template (Data Quality Check)}] 
\begin{lstlisting}[breaklines=true, xleftmargin=0pt, breakindent=0pt, columns=fullflexible] 
Please review the provided conversation snippet, which includes historical dialogue, current round dialogue, and the API call made in this round. Your task is to verify the accuracy of the API and parameters used in this round of dialogue. Assume the Assistant does not have knowledge of real-world information such as cinemas or restaurants; it relies on API calls to access information or carry out actions such as making a reservation. Use the guidelines below to correct any inaccuracies.
1. Check and correct the API selection in the \"Action\" field. Common errors include: a. Assistant's response indicates that an API call was made, but the \"Action\" field is empty. b. Assistant don't need to call any API to reply to the user's current round of conversation. In this case, calling is not necessary, but there is an API name in the \"Action\" field. c. The assistant's response shows that the necessary information required for API calls is missing. Therefore, the assistant is asking the user for additional information, but there is an API name in the \"Action\". d. The API listed in the \"Action\" is incorrect.
2. Verify and correct the parameters listed in \"Action Input\" and ensure they correctly match the API call. The \"Action Input\" should be formatted as {\"parameter_name\": \"value\", \"parameter_name\": \"value\", ...}.
3. Revise the content in \"Thought\" to include the correct rationale for selecting tools and parameters. The thought should only consider historical conversations and current user issues, assuming that the assistant's response is unknown.
4. Ensure all API and parameter names used are as defined and should not be invented.
Here is a list of APIs and their parameters for your reference:
[apis_information]
Please respond in the following JSON format: {\"Thought\": str, \"Action\": \"api_name\", \"Action Input\": {\"parameter_name\": \"value\", ...}}. If there is no \"Action\" or \"Action Input\" information, please return an empty string.
\end{lstlisting} 
\end{tcolorbox}

\begin{figure}[]
    \centering
    \resizebox{1.0\linewidth}{!}{
    \includegraphics[width=\textwidth]{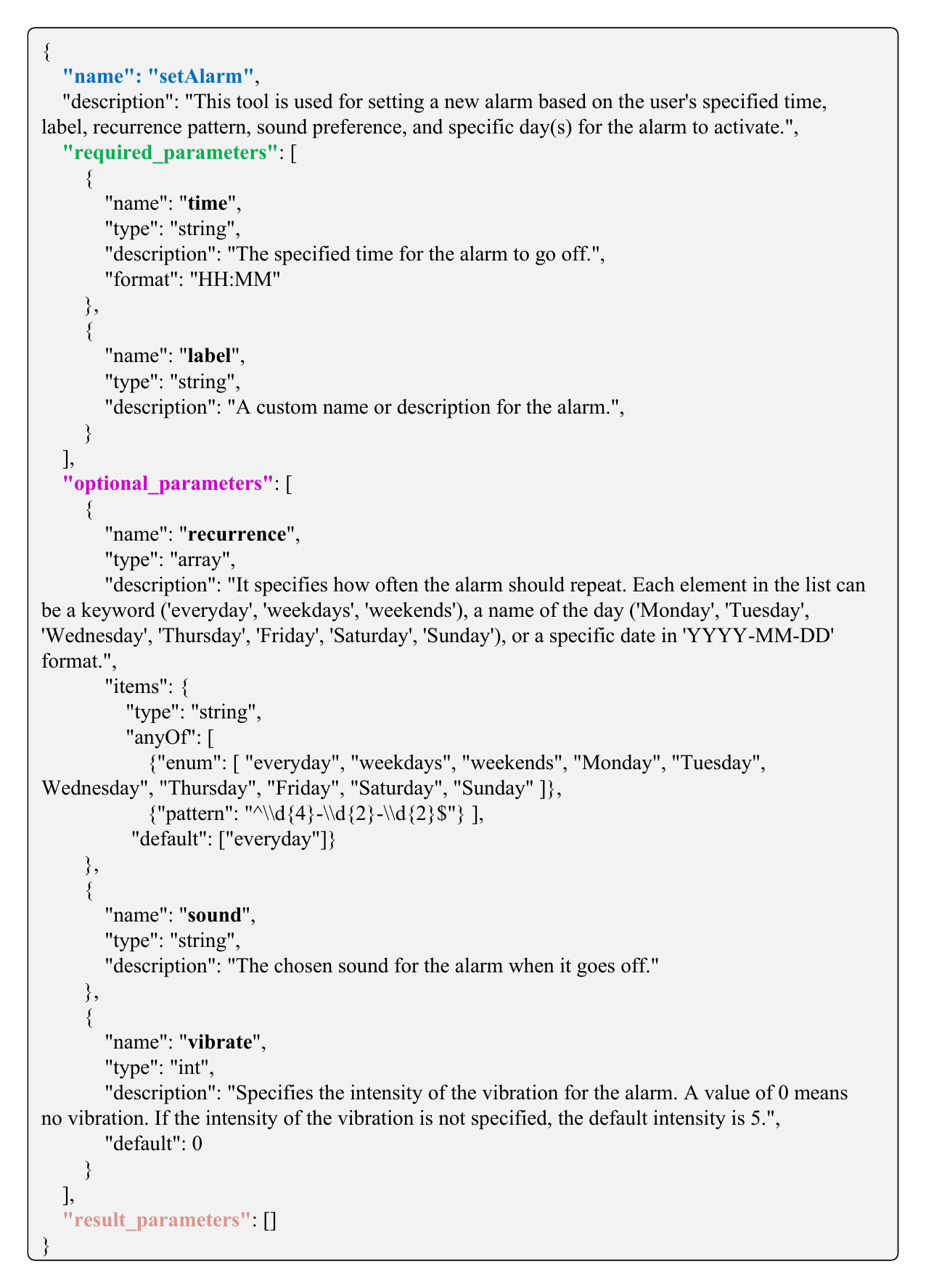}}
    %\vspace{-0.35cm}
    \caption{The JSON structure of the tool ``setAlarm'', which is used to create a new alarm based on user-specified parameters. The structure includes required, optional and result (\textit{i.e.}, return) parameters, along with their corresponding data types, descriptions, formats and default values.}
    \label{fig:tool_document}
    \vspace{-0.45cm}
\end{figure}

\paragraph{Example of Tool Document.\label{appendix: tool_document}} 
As shown in Figure \ref{fig:tool_document}, the tool document allows the model to determine the appropriate tool names and their usages. It contains all the tools we synthesized, each tool including its corresponding tool description, necessary parameters, optional parameters, parameter description and data type, as well as the returns. 

\section{More MTU-Bench Data Analysis.} 
\label{appx:data-analysis}

\begin{figure}[t]
    \centering
    \resizebox{\textwidth}{!}{
    \includegraphics{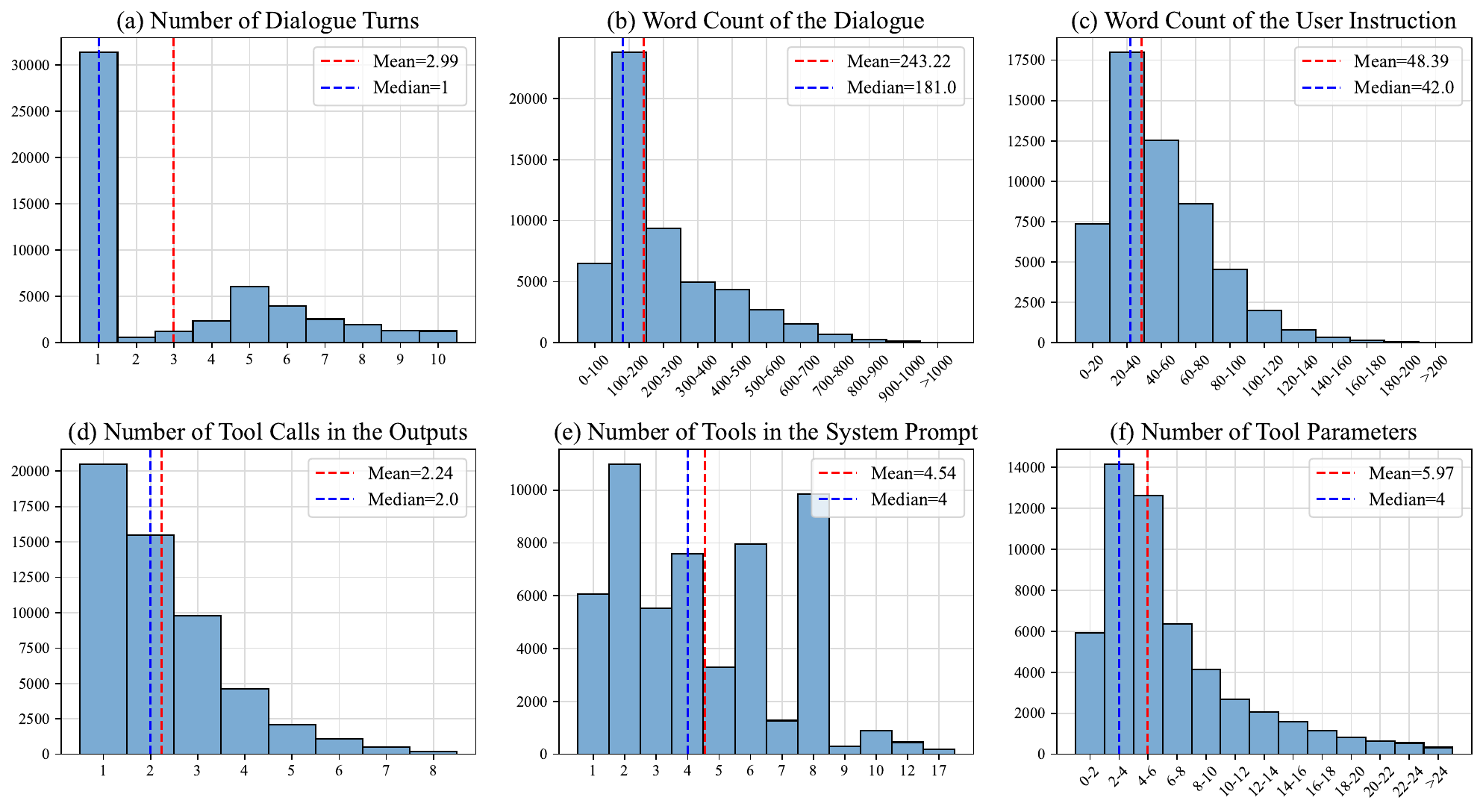}
    }
    \vspace{-0.8cm}
    \caption{Length distributions of the training data.}
    \label{fig:length_distribution_train}
\end{figure}

\begin{figure}[t]
    \centering
    \resizebox{\textwidth}{!}{
    \includegraphics{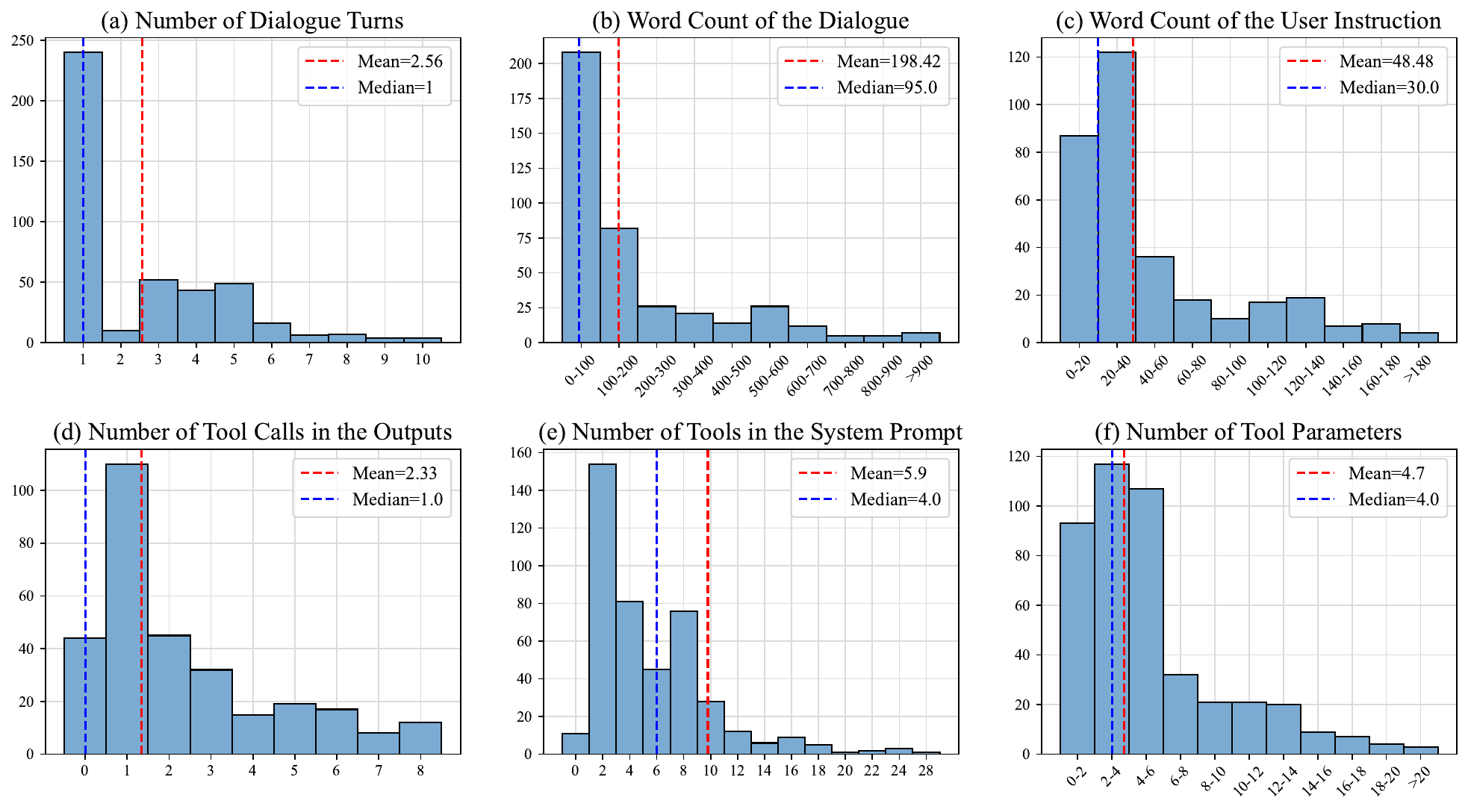}
    }
    \vspace{-0.8cm}
    \caption{Length distributions of the evaluation data.}
    \label{fig:length_distribution_eval}
\end{figure}

\begin{wraptable}[11]{r}{0.4\textwidth}
\caption{The number of dialogues under different settings.}
\begin{tabular}{c|l|cc}
\toprule                                   & \multicolumn{1}{c|}{}                        & \multicolumn{2}{c}{\textbf{Test}}          \\ \cline{3-4} 
\multirow{-2}{*}{\textbf{Setting}} & \multicolumn{1}{c|}{\multirow{-2}{*}{\textbf{Train}}} & \multicolumn{1}{c|}{\textbf{normal}} & \textbf{hard} \\ 
% \hline
\midrule
\textbf{S-S}                                                        & 14277                                        & \multicolumn{1}{c|}{52}     & 56   \\ \midrule
\textbf{S-M}                                                        & 13641                                        & \multicolumn{1}{c|}{55}     & 39   \\ \midrule
\textbf{M-S}                                                        & 19007                                        & \multicolumn{1}{c|}{54}     & 31   \\ \midrule
\textbf{M-M}                                                        & 7442                                         & \multicolumn{1}{c|}{42}     & 37   \\ \midrule
\textbf{OOD}                                                        & \multicolumn{1}{c|}{-}                       & \multicolumn{2}{c}{65}            \\ 
% \hline
\bottomrule
\end{tabular}

\label{tab:more-statistics}
\end{wraptable}

\paragraph{Length Distribution.} 
We illustrate the length distributions in Figure \ref{fig:length_distribution_train}, and Figure \ref{fig:length_distribution_eval} for the training and evaluation data, respectively.

\paragraph{More Statistics.} 
Table \ref{tab:more-statistics} provides a comprehensive summary of the dialogue statistics across diverse settings, distinct splits and subsets.

% \clearpage
% \newpage
% \paragraph{Data Samples.} We present the data samples for each setting below.
% \input{data_samples}
% \clearpage
% \newpage

\section{Details of MTU-Eval} 
\label{appx:mtu-eval-details}

\paragraph{Hard cases in the hard test set\label{appx:hard-cases}} encompass extensive parameters, nonsensical tool names, determination of specific parameter values, inability to call tools, interaction among multiple tools, and multi-turn parameter inheritance, as delineated in Table \ref{tab:hard-cases}.

\begin{table}[ht]
\small
\centering
\caption{Hard cases in the hard test set.} 
\begin{tabular}{p{1.7cm}|p{1.8cm}|p{10cm}}
\toprule
Type & Description & Case \\ \midrule 
Extensive Parameters & The quantity of parameters required to be specified exceeds six. & 
\textcolor{blue}{User}: Can you book me a non-refundable, one-stop flight on Delta Airlines from New York to Chicago, leaving on the 5th and returning on the 11th of this month in Economy class? \newline 
\textcolor{green}{Target}: Action: ReserveRoundtripFlights Action Input: \{``airlines'': `` Delta Airlines'', ``departure\_date'': ``2024-01-05'', ``destination\_city'': ``Chicago'', ``origin\_city'': ``New York'', ``return\_date'': ``2024-01-11'', ``seating\_class'': ``Economy'', ``number\_stops'': ``1'', ``refundable'': False\} 
\\ \midrule 
Nonsensical Tool Names & The designation of the tool lacks significance, for instance, ``abc''. & 
\textcolor{orange}{System Prompt}: ...The following is a list of APIs and their parameters that you can use: \{ \colorbox{orange!15}{``name'': ``eee''}, ``description'': ``Book an appointment at a dentist for a given time and date'', ``required\_parameters'': [``dentist\_name'', ``appointment\_time'', ``appointment\_date'']...\newline 
\textcolor{blue}{User}: I would like to book an appointment with Yvonne Yang at Greenview Dental Care on the 11th at 11:45. Do they offer cosmetic services? \newline 
\textcolor{green}{Target}: \colorbox{orange!15}{Action: eee} Action Input: \newline
\{``appointment\_date'': ``2024-01-11'', ``appointment\_time'': ``11:45'', ``dentist\_name'': ``Yvonne Yang``\} \\ \midrule
Determination of Specific Parameter Values &  Parameter values must conform to specific criteria. & 
\textcolor{orange}{Tool Definition}: ``name'': ``searchHotel'', ``description'': ``To search for hotels based on a set of criteria including rating, type, amenities, location, and price range.'', ``optional\_parameters'': [...{``name'': ``priceRange'', \colorbox{orange!15}{``choices'' = [``cheap'', ``moderate'', ``expensive'']}} \newline
\textcolor{blue}{User}: Can you help me find a moderately priced hotel in the centre with parking and internet availability? \newline
\textcolor{green}{Target}: Action: SearchHotel Action Input: {``area'': ``centre'', ``parking'': True, ``internet'': True, \colorbox{orange!15}{``priceRange'': ``moderate''}, ``return\_info'': [``all'']} \\ \midrule Inability to Call Tools & The list of tools does not contain any suitable options. & 
\textcolor{orange}{System Prompt}: You are a ...The following is a list of APIs and their parameters that you can use... \newline
\textcolor{blue}{Prompt}: Can you find me a drama movie with subtitles directed by Claire Denis? \newline
\textcolor{green}{Target}: Thought: The user wants to find a drama movie with subtitles directed by Claire Denis. However, there is no available API in the provided list that supports the searching for movies. Therefore, \colorbox{orange!15}{it's not possible to meet the user's request with the current tools.} \\ \midrule 
Interaction among Multiple Tools & In a single turn of dialogue, more than four tools are utilized. & 
\textcolor{blue}{User}: I am looking for the date of a concert in New York, followed by the weather forecast for that day. And I want to find a dentist, book an appointment at 10:00 on Febrary 2nd. And please add the dentist's phone to my contact list. Thank you! \newline
\textcolor{green}{Target}
Action: FindEvents Action Input: \{``event\_type'': ``concert'', ``city'': ``New York''\} 
Action : GetWeather Action Input: \{``city'': ``New York'', ``date'': ``FindEvents.date''\}
Action: BookAppointment Action Input: \{``dentist\_name'': ``FindProvider.dentist\_name'', ``appointment\_time'': ``10:00'', ``appointment\_date'': ``2024-02-02''\}
Action: AddNewContact Action Input: \{``contact\_name'': ``FindProvider.dentist\_name'', ``phone\_number'': ``FindProvider.phone\_number''\} \\ \midrule 
Multi-Turn Parameter Inheritance & The parameters for the current turn must be extracted from the historical dialogue. & 
User: Find some music please. \newline
Assistant: Action: LookupMusic \newline
Observation: \{``artist'': ``Ariana Grande'', ``song\_name'': ``Be My Baby''\} \newline
Assistant: Ok, how about \colorbox{orange!15}{Be My Baby} by Ariana Grande? \newline
User: That sounds good. \newline
Assistant: Should I play this now? \newline
\textcolor{blue}{User}: Yes. Play on kitchen speaker please. \newline
\textcolor{green}{Target}: Action: PlayMedia Action Input: \{ ``playback\_device'': ``kitchen speaker'', \colorbox{orange!15}{``song\_name'': ``Be My Baby''} \} \\ \bottomrule
\end{tabular}

\label{tab:hard-cases} % Add a label for referencing if needed
\end{table}

\paragraph{Prompt Templates for MTU-Eval.\label{appx:prompt-templates-evaluation}} 
During the evaluation, the models are provided with distinct system prompts for various settings, which encompass both the comprehensive task specifications and the tool documentation. 
We list the system prompts used for evaluation in Boxes \ref{appx:system-prompt-s-s}. 

\clearpage
\newpage 
\phantomsection
\label{appx:system-prompt-s-s}
\begin{tcolorbox}[colback=white!95!gray,colframe=gray!50!black,rounded corners, label={system-prompt-s-s}, title={System Prompt (S-S)}] 
\begin{lstlisting}[breaklines=true, xleftmargin=0pt, breakindent=0pt, columns=fullflexible] 
Please reply to the user based on their input and historical conversation information. You can choose to call external tools to implement it. Here are the call requirements and information about available APIs.
1. Please provide your thought process in \"Thought\", including user intent analysis, whether to call APIs, and how to call APIs.
2. When a user's request can be satisfied by calling an API, please provide the required calling information in the following format: \nAction: The name of the API to be called.\nAction Input: The parameter information required to call the API, in Json format.
3. If the user's needs can be met without calling the API, then no API call action will be made.
4. If there is a lack of mandatory information that makes it impossible to call a specific API, then no API call action will be made.
5. The name and parameter name of the API must be consistent with the provided API information. The value of the parameter should be extracted from the context, and the information should not be fictional.
6. If none of the apis provided are available to meet the user's requirements, no Action is taken.
7. If you need to call the API, your output format should be:
Thought: xxxx\nAction: xxxx\nAction Input: xxxx
For example: 
Thought: The user is looking for a one-way flight for three people from Las Vegas to Atlanta on a specific date, March 5th. Let's call the SearchOnewayFlight API to find the available flights.\nAction: SearchOnewayFlight\nAction Input: {\"origin_city\": \"Las Vegas\", \"destination_city\": \"Atlanta\", \"departure_date\": \"2024-03-05\", \"seating_class\": \"Economy\"}
If you don't need to call the API, your output format should be:
Thought: xxxx
The following is a list of APIs and their parameters that you can use:
[apis information]
\end{lstlisting} 
\end{tcolorbox}

\label{appx:system-prompt-s-m}
\begin{tcolorbox}[colback=white!95!gray,colframe=gray!50!black,rounded corners, label={system-prompt-s-m}, title={System Prompt (S-M)}] 
\begin{lstlisting}[breaklines=true, xleftmargin=0pt, breakindent=0pt, columns=fullflexible] 
Please reply to the user based on their input. You can choose to call external APIs to implement it. Here are the call requirements and information about available APIs.
1. Please provide your thought process in \"Thought\", including user intent analysis, whether to call APIs, and how to call APIs.
2. When a user's request can be satisfied by calling APIs, please provide the required calling information in the following format: \nAction: The name of the APIs.\nAction Input: the parameter information required to call APIs, in Json format. For example, \nAction: \"api_name_A\"\nAction Input: {\"parameter_name_A.1\": \"parameter_value_A.1\", ...}\nAction: \"api_name_B\"\nAction Input: {\"parameter_name_B.1\": \"parameter_value_B.2\", ...}, ...
3. There may be an interaction relationship between APIs, where the parameter value returned by the previous API call needs to be used as the parameter value for the next API call. Please use \"previous_API_name.return_parameter_name\" as the parameter value for the new API call.
4. Multiple APIs may need to be called to meet the user's needs. Please pay attention to the order of APIs' call.
5. There may be an interaction relationship between APIs, where the parameter value returned by the previous API call needs to be used as the parameter value for the next API call. Please use \"previous_API_name.return_parameter_name\" as the parameter value for the new API call.
6. The name and parameter name of the API must be consistent with the provided API information. The value of the parameter should be extracted from the context, and the information should not be fictional.
7. If you need to call the API, your output format should be:
\nThought: xxxx\nAction: xxxx\nAction Input: xxxx\nAction: xxxx\nAction Input:xxxx\nAction: xxxx\nAction Input: xxxx...
If you don't need to call the API, your output format should be:
\nThought: xxxx
The following is a list of APIs and their parameters that you can use:
[apis information]
\end{lstlisting} 
\end{tcolorbox}

\label{appx:system-prompt-m-s}
\begin{tcolorbox}[colback=white!95!gray,colframe=gray!50!black,rounded corners, label={system-prompt-m-s}, title={System Prompt (M-S)}] 
\begin{lstlisting}[breaklines=true, xleftmargin=0pt, breakindent=0pt, columns=fullflexible] 
Please reply to the user based on their input and historical conversation information. You can choose to call external tools to implement it. Here are the call requirements and information about available APIs.
1. Please provide your thought process in \"Thought\", including user intent analysis, whether to call APIs, and how to call APIs.
2. When a user's request can be satisfied by calling an API, please provide the required calling information in the following format: \nAction: The name of the API to be called.\nAction Input: The parameter information required to call the API, in Json format.
3. \"Observation\" is the information returned by API calls.
4. If the user's needs can be met without calling the API, then no API call action will be made.
5.  If there is a lack of mandatory information that makes it impossible to call a specific API, then no API call action will be made.
6. Note that if the dialogue history already contains the required information, there is no need to call the tool again.
7. The name and parameter name of the API must be consistent with the provided API information. The value of the parameter should be extracted from the context, and the information should not be fictional.
8. If you need to call the API, your output format should be:
\nThought: xxxx\nAction: xxxx\nAction Input: xxxx
Therefore, if you don't need to call the API, your output format should be:
\nThought: xxxx
The following is a list of APIs and their parameters that you can use:
[apis information]
\end{lstlisting} 
\end{tcolorbox}

\label{appx:system-prompt-m-m}
\begin{tcolorbox}[colback=white!95!gray,colframe=gray!50!black,rounded corners, label={system-prompt-m-m}, title={System Prompt (M-M)}] 
\begin{lstlisting}[breaklines=true, xleftmargin=0pt, breakindent=0pt, columns=fullflexible] 
Please reply to the user based on their input and history conversation. You can choose to call external APIs to implement it. Here are the call requirements and information about available APIs.
1. Please provide your thought process in \"Thought\", including user intent analysis, whether to call APIs, and how to call APIs.
2. When a user's request can be satisfied by calling APIs, please provide the required calling information in the following format: \nAction: The name of the APIs.\nAction Input: the parameter information required to call APIs, in Json format. For example, \nAction: \"api_name_A\"\nAction Input: {\"parameter_name_A.1\": \"parameter_value_A.1\", ...}\nAction: \"api_name_B\"\nAction Input: {\"parameter_name_B.1\": \"parameter_value_B.2\", ...}, ...
3. There may be an interaction relationship between APIs, where the parameter value returned by the previous API call needs to be used as the parameter value for the next API call. Please use \"previous_API_name.return_parameter_name\" as the parameter value for the new API call.
4. Multiple APIs may need to be called to meet the user's needs. Please pay attention to the order of APIs' call.
5. The name and parameter name of the API must be consistent with the provided API information. The value of the parameter should be extracted from the context, and the information should not be fictional.
6. If you need to call the API, your output format should be:
Thought: xxxx\nAction: xxxx\nAction Input: xxxx\nAction: xxxx\nAction Input:xxxx\nAction: xxxx\nAction Input: xxxx...
If you don't need to call the API, your output format should be:
Thought: xxxx
The following is a list of APIs and their parameters that you can use:
[apis information]
\end{lstlisting} 
\end{tcolorbox}
\newpage

\paragraph{Examples for computing metrics.\label{appx:example-metrics}} The examples are listed below: 

% \rule{\linewidth}{0.4pt}

% \paragraph{\colorbox{orange!15}{Tool Selection Accuracy (abbr. TS)}} (for S-S, M-S)
\fcolorbox{black}{orange!15}{\parbox{\linewidth}{\paragraph{Tool Selection Accuracy (abbr. TS)} (for S-S, M-S)\strut}}

\colorbox{green!15}{\textbf{\cmark Positive Example:}}

- \textbf{User}: ``Search for the latest news on artificial intelligence.''

- \textbf{Assistant}: OK! Let me search the web. \texttt{search\_web(query="latest news on AI")}

- \textbf{Ground Truth Tool}: \texttt{search\_web}

- \textbf{Assistant's Tool}: \texttt{search\_web}

- \textbf{Score}: TS = 1

- \textbf{Explanation}: The assistant correctly selected the tool \texttt{search\_web}, matching the ground truth.

\colorbox{red!15}{\textbf{\xmark Negative Example:}}

- \textbf{User}: ``Calculate the square root of 16.''

- \textbf{Assistant}: OK! Let me use the calculator. \texttt{search\_web(query="square root of 16")}

- \textbf{Ground Truth Tool}: \texttt{calculate}

- \textbf{Assistant's Tool}: \texttt{search\_web}

- \textbf{Score}: TS = 0

- \textbf{Explanation}: The assistant incorrectly selected \texttt{search\_web} instead of the correct tool \texttt{calculate}.

\rule{\linewidth}{0.4pt}

\fcolorbox{black}{orange!15}{\parbox{\linewidth}{\paragraph{Parameter Selection Accuracy (abbr. PS)} (for S-S, M-S)\strut}}

\colorbox{green!15}{\textbf{\cmark Positive Example:}}

- \textbf{User}: ``Translate `good evening' to German.''

- \textbf{Assistant}: Sure, translating now. \texttt{translate(text="good evening", target\_language="German")}

- \textbf{Ground Truth Parameters}: \{text=``good evening'', target\_language=``German''\}

- \textbf{Assistant's Parameters}: \{text=``good evening'', target\_language=``German''\}

- \textbf{Score}: PS = 1

- \textbf{Explanation}: The assistant provided the correct parameters matching the ground truth.

\colorbox{red!15}{\textbf{\xmark Negative Example:}}

- \textbf{User}: ``Set an alarm for 6 AM tomorrow.''

- \textbf{Assistant}: Setting your alarm. \texttt{set\_alarm(time="6 PM", name="tomorrow")}

- \textbf{Ground Truth Parameters}: \{time=``6 AM'', date=``tomorrow''\}

- \textbf{Assistant's Parameters}: \{time=``6 PM'', name=``tomorrow''\}

- \textbf{Score}: PS = 0

- \textbf{Explanation}: The assistant incorrectly set the value for the time parameter. The assistant incorrectly set the key for the date parameter.

\rule{\linewidth}{0.4pt}

\fcolorbox{black}{orange!15}{\parbox{\linewidth}{\paragraph{Success Rate (abbr. SR)} (for M-S, M-M)\strut}}

\colorbox{green!15}{\textbf{\cmark Positive Example:}}

- \textbf{User}: ``I need to send an email to Alex.''

- \textbf{Assistant}: Sure, composing an email. \texttt{compose\_email(recipient="Alex", subject="Hello", body="Hi Alex, how are you?")}

- \textbf{User}: ``Also, schedule a meeting with him next Monday at 2 PM.''

- \textbf{Assistant}: Scheduling the meeting. \texttt{schedule\_meeting(participant="Alex", date="next Monday", time="2 PM")}

- \textbf{Ground Truth}: No errors throughout all turns.

- \textbf{Score}: SR = 1

- \textbf{Explanation}: The assistant correctly used the tools in all dialogue turns without any errors.

\colorbox{red!15}{\textbf{\xmark Negative Example:}}

- \textbf{User}: ``Find the fastest route to the airport.''

- \textbf{Assistant}: Finding the fastest route. \texttt{get\_directions(destination="airport")}

- \textbf{User}: ``Book a taxi to pick me up in 30 minutes.''

- \textbf{Assistant}: Booking a taxi. \texttt{book\_taxi(pickup\_time="in 3 minutes")}

- \textbf{Ground Truth}: Assistant should have set the pickup time to ``in 30 minutes''.

- \textbf{Score}: SR = 0

- \textbf{Explanation}: The assistant made an error in the second turn (incorrect pickup time), so the entire dialogue is considered unsuccessful.

\rule{\linewidth}{0.4pt}

\fcolorbox{black}{orange!15}{\parbox{\linewidth}{\paragraph{Averaged Turn Success Rate (abbr. ATS)} (for M-S, M-M)\strut}}

\colorbox{green!15}{\textbf{\cmark Positive Example:}}

- \textbf{Turn 1}:

  - \textbf{User}: ``What's on my schedule today?''

  - \textbf{Assistant}: Checking your schedule. \texttt{get\_schedule(date="today")}

  - \textbf{Turn Success}: 1

- \textbf{Turn 2}:

  - \textbf{User}: ``Add a dentist appointment at 4 PM.''

  - \textbf{Assistant}: Adding the appointment. \texttt{add\_event(event="dentist appointment", time="4 PM", date="today")}

  - \textbf{Turn Success}: 1

- \textbf{Turn 3}:

  - \textbf{User}: ``Remind me an hour before.''

  - \textbf{Assistant}: Setting a reminder. \texttt{set\_reminder(event="dentist appointment", time="3 PM", date="today")}

  - \textbf{Turn Success}: 1

- \textbf{Score}: ATS = (1 + 1 + 1)/3 = 1.0

- \textbf{Explanation}: All turns were executed correctly.

\colorbox{red!15}{\textbf{\xmark Negative Example:}}

- \textbf{Turn 1}:

  - \textbf{User}: ``Play my workout playlist.''

  - \textbf{Assistant}: Playing your playlist. \texttt{play\_music(playlist="workout")}

  - \textbf{Turn Success}: 1

- \textbf{Turn 2}:

  - \textbf{User}: ``Skip to the next song.''

  - \textbf{Assistant}: Restarting the song. \texttt{play\_music(action="restart")}

  - \textbf{Turn Success}: 0 (Incorrect action parameter)

- \textbf{Turn 3}:

  - \textbf{User}: ``Increase the volume.''

  - \textbf{Assistant}: Increasing volume. \texttt{adjust\_volume(level="up")}

  - \textbf{Turn Success}: 1

- \textbf{Score}: ATS = (1 + 0 + 1)/3 = 0.6667

- \textbf{Explanation}: One out of three turns was unsuccessful, leading to an ATS of approximately 0.6667.

\rule{\linewidth}{0.4pt}

\fcolorbox{black}{orange!15}{\parbox{\linewidth}{\paragraph{Soft Averaged Turn Success Rate (abbr. SATS)} (for M-S, M-M)\strut}}

\colorbox{green!15}{\textbf{\cmark Positive Example:}}

- \textbf{Turn 1}:

  - \textbf{User}: ``What's the current time in Tokyo?''

  - \textbf{Assistant}: The current time is... \texttt{get\_time(location="Tokyo") $\to$ 3:00 PM}

  - \textbf{Turn Success}: 1

- \textbf{Turn 2}:

  - \textbf{User}: ``Schedule a call with the Japan office at that time tomorrow.''

  - \textbf{Assistant}: Scheduling the call. \texttt{schedule\_call(participant="Japan office", time="3:00 PM", date="tomorrow")}

  - \textbf{Turn Success}: 1

- \textbf{Score}: SATS = (1 + 1)/2 = 1.0

- \textbf{Explanation}: Both turns were successful, so the SATS is 1.

\colorbox{red!15}{\textbf{\xmark Negative Example:}}

- \textbf{Turn 1}:

  - \textbf{User}: ``Check the weather in New York.''

  - \textbf{Assistant}: Checking the weather. \texttt{get\_weather(location="New York")}

  - \textbf{Turn Success}: 1

- \textbf{Turn 2}:

  - \textbf{User}: ``Schedule a meeting tomorrow at 10 AM.''

  - \textbf{Assistant}: Scheduling the meeting. \texttt{schedule\_meeting(date="tomorrow", time="10 AM")}

  - \textbf{Turn Success}: 1

- \textbf{Turn 3}:

  - \textbf{User}: ``Set an alarm for 9 AM.''

  - \textbf{Assistant}: Setting an alarm. \texttt{set\_alarm(time="9 PM")}

  - \textbf{Turn Success}: 0 (Incorrect time parameter)

- \textbf{Turn 4}:

  - \textbf{User}: ``Remind me to call John at 11:00 AM.''

  - \textbf{Assistant}: Setting a reminder. \texttt{set\_reminder(event="call John", time="11:00 AM")}

  - \textbf{Turn Success}: 1

- \textbf{Turn 5}:

  - \textbf{User}: ``Send an email to Sarah.''

  - \textbf{Assistant}: Composing an email. \texttt{compose\_email(recipient="Sarah", subject="Hello", body="Hi Sarah, ...")}

  - \textbf{Turn Success}: 1

\textbf{Calculations}:

- Last incorrect turn index \( i = 3 \)

- For \( j = 1 \) and \( j = 2 \):

  - Since \( j < i \), Score = 1

- For \( j = 3 \):

  - Turn is incorrect, Score = 0

- For \( j = 4 \):

  - \( j > i \), Score = \( 1 - e^{-(4 - 3)} = 1 - e^{-1} \approx 0.6321 \)

- For \( j = 5 \):

  - \( j > i \), Score = \( 1 - e^{-(5 - 3)} = 1 - e^{-2} \approx 0.8647 \)

- \textbf{Score}: SATS = \( \frac{1 + 1 + 0 + 0.6321 + 0.8647}{5} \approx \frac{3.4968}{5} \approx 0.6994 \)

- \textbf{Explanation}: The error in the third turn reduces the scores of subsequent turns due to the exponential decay, resulting in a SATS of approximately 0.6994.

\rule{\linewidth}{0.4pt}

\fcolorbox{black}{orange!15}{\parbox{\linewidth}{\paragraph{Task Process Rate (abbr. TPR)} (for M-S, M-M)\strut}}

\colorbox{green!15}{\textbf{\cmark Positive Example:}}

- \textbf{Total Turns}: \( n = 4 \)

- \textbf{First Incorrect Turn Index}: No incorrect turns.

- \textbf{Score}: TPR = \( \frac{n}{n} = 1 \)

- \textbf{Explanation}: Since there are no errors, the task was processed completely.

\colorbox{red!15}{\textbf{\xmark Negative Example:}}

- \textbf{Total Turns}: \( n = 5 \)

- \textbf{First Incorrect Turn Index}: \( i = 3 \) (error occurs at turn 3)

- \textbf{Score}: TPR = \( \frac{i - 1}{n} = \frac{2}{5} = 0.4 \)

- \textbf{Explanation}: The task process rate indicates the proportion of the task completed before the first error.

\rule{\linewidth}{0.4pt}

\fcolorbox{black}{orange!15}{\parbox{\linewidth}{\paragraph{Tool Number Accuracy (abbr. TN)} (for S-M, M-M)\strut}}

\colorbox{green!15}{\textbf{\cmark Positive Example:}}

- \textbf{Ground Truth Tool List (GT)}: \{\texttt{search\_web}, \texttt{summarize\_text}, \texttt{translate\_text}\}

- \textbf{Assistant's Predicted Tool List (Pred)}: \{\texttt{search\_web}, \texttt{summarize\_text}, \texttt{translate\_text}\}

- \textbf{Score}: TN = \( \frac{|\text{Pred} \cap \text{GT}|}{|\text{Pred} \cup \text{GT}|} = \frac{3}{3} = 1.0 \)

- \textbf{Explanation}: All tools predicted by the assistant match the ground truth.

\colorbox{red!15}{\textbf{\xmark Negative Example:}}

- \textbf{Ground Truth Tool List (GT)}: \{\texttt{search\_web}, \texttt{translate\_text}\}

- \textbf{Assistant's Predicted Tool List (Pred)}: \{\texttt{search\_web}, \texttt{play\_music}, \texttt{set\_alarm}\}

- \textbf{Score}: TN = \( \frac{|\text{Pred} \cap \text{GT}|}{|\text{Pred} \cup \text{GT}|} = \frac{1}{4} = 0.25 \)

- \textbf{Explanation}: Only \texttt{search\_web} is common between the predicted and ground truth lists, out of four unique tools.

\rule{\linewidth}{0.4pt}

\fcolorbox{black}{orange!15}{\parbox{\linewidth}{\paragraph{Tool Order Accuracy (abbr. TO)} (for S-M, M-M)\strut}}

\colorbox{green!15}{\textbf{\cmark Positive Example:}}

- \textbf{Ground Truth Tool Sequence (GT)}: [\texttt{search\_web}, \texttt{extract\_data}, \texttt{generate\_report}]

- \textbf{Assistant's Predicted Tool Sequence (Pred)}: [\texttt{search\_web}, \texttt{extract\_data}, \texttt{generate\_report}]

- \textbf{Longest Common Subsequence (LCS)}: [\texttt{search\_web}, \texttt{extract\_data}, \texttt{generate\_report}]

- \textbf{Starting Position (i)}: 1

- \textbf{Total Tools in Pred}: \( |\text{Pred}| = 3 \)

- \textbf{Decay Coefficient (t)}:

  \( t = \cos\left(\frac{\pi}{2} \times \frac{i}{|\text{Pred}|}\right) = \cos\left(\frac{\pi}{2} \times \frac{1}{3}\right) \approx 0.8660 \)

- \textbf{Score}:

  \( \text{TO} = t \times \frac{|\text{LCS}|}{|\text{GT}|} = 0.8660 \times \frac{3}{3} = 0.8660 \times 1 = 0.8660 \)

- \textbf{Explanation}: The assistant's sequence perfectly matches the ground truth, resulting in a high TO score.

\colorbox{red!15}{\textbf{\xmark Negative Example:}}

- \textbf{Ground Truth Tool Sequence (GT)}: [\texttt{get\_weather}, \texttt{plan\_route}, \texttt{book\_hotel}]

- \textbf{Assistant's Predicted Tool Sequence (Pred)}: [\texttt{book\_hotel}, \texttt{plan\_route}, \texttt{get\_weather}]

- \textbf{Longest Common Subsequence (LCS)}: [\texttt{plan\_route}]

- \textbf{Starting Position (i)}: 2

- \textbf{Total Tools in Pred}: \( |\text{Pred}| = 3 \)

- \textbf{Decay Coefficient (t)}:

  \( t = \cos\left(\frac{\pi}{2} \times \frac{i}{|\text{Pred}|}\right) = \cos\left(\frac{\pi}{2} \times \frac{2}{3}\right) \approx 0.5000 \)

- \textbf{Score}:

  \( \text{TO} = t \times \frac{|\text{LCS}|}{|\text{GT}|} = 0.5000 \times \frac{1}{3} \approx 0.1667 \)

- \textbf{Explanation}: Only one tool matches in sequence, and it starts at the second position, leading to a low TO score.

\rule{\linewidth}{0.4pt}
\newpage

\section{More Experimental Analysis\label{appx:more-analysis}} 

\paragraph{Detailed Evaluation Results on Hard Set.\label{appendex:hard_result}} 
In Table~\ref{tab:singletool-hard} and Table~\ref{tab:multitool-hard}, we present the performance of various LLMs evaluated on the hard subset of MTU-Eval. 
Our analysis indicates that among closed-source models, GLM-4-Plus excels in handling single-tool scenarios, surpassing even GPT-4. Conversely, Qwen-Max demonstrates superior performance in multi-turn and multi-tool scenarios, also outperforming GPT-4. 
In single-turn multi-tool scenarios, GPT-4 and DeepSeek V2.5 exhibit relatively better performance. 
Furthermore, there remains a significant gap between open-source models and their closed-source counterparts. 
Among the open-source models, our MTU-LLaMA, Qwen2-72B, and LLaMA3-70B lead the performance metrics. 
Specifically, Qwen2-72B shows a slight advantage in multi-tool scenarios, while LLaMA3-70B excels in single-tool scenarios. MTU-LLaMA, however, demonstrates a balanced ability across almost all evaluated settings.

\begin{table*}[]
\caption{Results of different models on the \textbf{hard} set of MTU-Eval \textbf{(S-S \& M-S)}. ``S-S'' and ``M-S'' denote ``Single-Turn Single-Tool'' and ``Multi-Turn Single-Tool'' settings, respectively.  All the baselines are instruction-tuned models.}
\vspace{-2mm}
\resizebox{1.0\textwidth}{!}{
% \begin{tabular}{c|ccc|ccc|ccc|ccc|c}
% \begin{tabular}{c|cccccccccccc|>{\columncolor[gray]{0.93}}c@{}}
\begin{tabular}{c|ccc|ccccccc}
\toprule
\multirow{2}{*}{\textbf{Models}} & \multicolumn{3}{c|}{\textbf{S-S}} & \multicolumn{7}{c}{\textbf{M-S}}\\
\cmidrule(lr){2-11} 

& \textbf{TS} & \textbf{PS}   &\textbf{Avg.}  & \textbf{TS}   & \textbf{PS}    & \textbf{ATS}     & \textbf{SATS}  & \textbf{SR}     & \textbf{TPR}        & \textbf{Avg.}   \\ 

\midrule
\multicolumn{11}{c}{\textit{Closed-Source Large Language Models}}\\ 

\midrule 
GPT-4 &88.46 	&67.31 	&77.88  	&\textbf{85.47} 	&\underline{62.01} 	&55.77 	&46.57 	&0.00 	&17.80 	&44.61   \\
GPT-3.5 &50.00 	&33.93 	&41.96 	&71.51 	&43.02 	&39.45 	&29.33 	&0.00 	&1.83 	&30.86   \\
Qwen-Max &88.46 	&67.31 	&77.88 	&80.45 	&56.98 	&50.54 	&41.77 	&\underline{3.23} 	&19.67 	&42.11 \\
GLM-4-Plus &\textbf{92.31} 	&\textbf{73.08} 	&\textbf{82.69}	&\underline{84.92} &\textbf{63.69} &\textbf{58.40} &\textbf{48.79} &\textbf{6.45} &\textbf{23.44} &\textbf{47.61} \\
DeepSeek V2.5 &\underline{90.38} 	&\underline{71.15} 	&\underline{80.77} 	&83.24 	&58.10 	&\underline{56.10} 	&\underline{47.28} 	&\underline{3.23} 	&\underline{21.66} 	&\underline{44.94} \\
\midrule
\multicolumn{11}{c}{\textit{Open-Source Large Language Models}}\\ 
\midrule
LLaMA2-7B &32.14 	&25.00 	&28.57 	&34.64 	&26.82 	&22.32 	&16.32 	&0.00 	&2.70 	&17.13 \\
LLaMA2-70B &32.14 	&25.00 	&28.57 	&52.51 	&30.17 	&28.45 &21.62 	&0.00 	&8.03 	&23.46   \\ \midrule
LLaMA3-8B &33.93 	&17.86 	&25.89 	&37.99 	&13.97 	&11.64 	&8.15 	&0.00 	&5.38 	&12.85  \\ 
LLaMA3-70B &\resultone{82.14} 	&\resultone{60.71} 	&\resultone{71.43} 	&\resultthird{75.42} 	&\resulttwo{52.51} 	&\resulttwo{51.68} 	&\resulttwo{42.03} 	&\resultone{3.23} 	&\resultone{17.53} 	&\resulttwo{40.40}   \\  \midrule
Qwen1.5-14B &44.64 	&44.64 	&44.64 	&45.81 	&45.81 	&40.28 	&32.05 	&0.00 	&12.40 	&29.39 \\
Qwen1.5-72B &57.69 	& \resultthird{55.77} 	&56.73 	&48.11 	&48.11  	&40.10 	&31.62 	&0.00 	&11.57 	&29.92  \\  \midrule
Qwen2-7B &\resultthird{67.86} 	& 50.00 	&\resultthird{58.93} &64.80 	&36.31 	&32.90 	&24.91 	&0.00 	&13.48 	&28.73  \\
Qwen2-72B &\resulttwo{78.93} 	&\resulttwo{57.86} 	&\resulttwo{68.40} 	&\resultone{78.77} 	&\resultthird{50.84} 	&\resultthird{47.19} 	&\resultthird{37.20} 	&0.00 	&\resultthird{16.50} 	&\resultthird{38.42}  \\  \midrule
Mistral-7B &36.54 	&17.31 	&26.92 	&63.69 	&34.64  	&31.29 	&23.90 	&0.00 	&2.74 	&26.04  \\  \midrule
ChatGLM3-6B &13.64 	&4.55 	&9.09 	&21.44 	&5.21 	&4.13 	&2.61 	&0.00 	&0.00 	&5.57   \\
GLM-4-9B &59.62 	&34.62 	&47.12 	&64.80 	&37.99  &36.56 	&29.42 	&0.00 	&12.52 	&30.22 \\ 
\midrule
\multicolumn{11}{c}{\textit{Tool-Use-Specific Large Language Models}}\\  \midrule
ToolLLaMA2-7B &32.69 	&3.85 	&18.27 	&25.70 	&3.35 	&3.50 	&2.60 	&0.00 	&0.65 	&10.19  \\
MTU-LLaMA(ours) &51.79	&28.57	&40.18 &\resulttwo{78.21} 	&\resultone{59.22} 	&\resultone{57.35} 	&\resultone{46.90} 	&0.00 	&\resulttwo{16.93} 	&\resultone{43.10}   \\
\bottomrule
\end{tabular}
}
\label{tab:singletool-hard}
\end{table*}

\begin{table*}[]
\caption{Results of different models on the \textbf{hard} set of MTU-Eval \textbf{(S-M \& M-M)}. ``S-M'' and ``M-M'' denote ``Single-Turn Multi-Tool'' and ``Multi-Turn Multi-Tool'' settings, respectively.  All the baselines are instruction-tuned models.}
\vspace{-2mm}
\resizebox{1.0\textwidth}{!}{
% \begin{tabular}{c|ccc|ccc|ccc|ccc|c}
% \begin{tabular}{c|cccccccccccc|>{\columncolor[gray]{0.93}}c@{}}
\begin{tabular}{c|ccc|ccccccc}
\toprule
\multirow{2}{*}{\textbf{Models}} & \multicolumn{3}{c|}{\textbf{S-M}} & \multicolumn{7}{c}{\textbf{M-M}}\\
\cmidrule(lr){2-11} 

& \textbf{TN} & \textbf{TO}   &\textbf{Avg.}  & \textbf{TN}   & \textbf{TO}    & \textbf{ATS}     & \textbf{SATS}  & \textbf{SR}     & \textbf{TPR}        & \textbf{Avg.}   \\ 

\midrule
\multicolumn{11}{c}{\textit{Closed-Source Large Language Models}}\\ 
\midrule 
GPT-4 &\textbf{57.77} 	&\textbf{58.37} 	&\textbf{58.07} 	&\textbf{57.08} 	&\underline{58.31} 	&\underline{49.28} 	&\underline{45.44} 	&0.00 	&\underline{38.06} 	&\underline{41.36}  \\
GPT-3.5 &17.75 	&19.04 &18.39	&21.31 	&22.39 	&15.09 	&10.41 	&0.00 	&2.03  &11.87 \\
Qwen-Max &23.75 	&24.28 	&24.01 	&\underline{57.04} 	&\textbf{58.56} 	&\textbf{49.96} 	&\textbf{49.13} 	&\textbf{8.11} 	&\textbf{47.70} 	&\textbf{45.08} \\
GLM-4-Plus &32.43 	&29.36 	&30.90 	&52.74 	&53.74 	&47.03 	&43.11 	&\underline{5.41} 	&35.18 	&39.53 \\
DeepSeek V2.5 &\underline{40.06} 	&\underline{39.96} 	&\underline{40.01} 	&43.51 	&45.21 	&35.45 	&32.49 	&0.00 	&27.07 	&30.62  \\ \midrule
\multicolumn{11}{c}{\textit{Open-Source Large Language Models}}\\ 
\midrule
LLaMA2-7B &2.14 &2.56 &2.35 &15.14 	&15.27 	&15.00 	&13.69 	&0.00 	&11.44 	&11.76 \\
LLaMA2-70B  &1.39 	&2.08 	&1.74 	&20.23 	&20.74 	&20.54 	&20.04 	&0.00 	&\resulttwo{19.19} 	&16.79  \\  \midrule
LLaMA3-8B &9.25 	&10.57 	&9.91 	&11.45 	&12.60 	&6.26 	&4.33 	&0.00 	&0.68 	&5.89     \\ 
LLaMA3-70B & 19.70 	&\resultthird{21.64} 	&\resultthird{20.67} 	&\resultthird{31.74} 	&\resultthird{33.13} 	&\resultthird{24.86} 	&\resultthird{20.63} 	&0.00 	& 13.02 	&\resultthird{20.56}    \\  \midrule
Qwen1.5-14B &12.23 	&13.39 	&12.81 	&15.14 	&16.92 	&11.98 	&8.82 	&0.00 	&3.38 	&9.37   \\
Qwen1.5-72B &19.27 	&18.42 	&18.85 	&26.92 	&27.07 	&21.71 	&18.61 	&0.00 	&13.29 	&17.93    \\  \midrule
Qwen2-7B  &15.24 	&16.37 	&17.50 	&16.03 	&16.41 	&13.92 	&10.37 	&0.00 	&4.28 	&10.17  \\
Qwen2-72B &\resulttwo{35.12} 	&\resulttwo{36.13} 	&\resulttwo{37.14} 	&\resulttwo{35.05} 	&\resultone{36.83} 	&\resulttwo{30.72} 	&\resultone{27.52} 	&0.00 	&\resultone{20.68} 	&\resultone{25.13}   \\  \midrule
Mistral-7B &10.18 	&12.79 	&11.48 	&17.05 	&17.81 	&13.38 	&10.69 	&0.00 	&6.08 	&10.84    \\  \midrule
ChatGLM3-6B &4.58 	&6.57 	&5.58 	&16.54 	&16.67 	&13.92 	&12.32 	&0.00 	&9.23 	&11.45   \\
GLM-4-9B &\resultthird{20.00} 	&17.96 	&18.98 	&17.43 	&17.94 	&12.75 	&8.31 	&0.00 	&0.68 	&9.52   \\ \midrule
\multicolumn{11}{c}{\textit{Tool-Use-Specific Large Language Models}}\\  \midrule
ToolLLaMA2-7B &0.51 	&0.51 	&0.51 	&3.82 	&3.82 	&3.15 	&2.32 	&0.00 	&0.90 	&2.34   \\
MTU-LLaMA(ours) &\resultone{39.77}	&\resultone{38.85}	&\resultone{39.31}	&\resultone{35.90}  &\resulttwo{36.67} 	&\resultone{34.14} 	&\resulttwo{27.37} 	&0.00 	& \resultthird{14.14}	&\resulttwo{24.70}     \\
\bottomrule
\end{tabular}
}
% \vspace{-3mm}
\label{tab:multitool-hard}
\end{table*}

\paragraph{Error Analysis\label{appx:error-analysis}.} 
The error cases of GPT-4 results on S-S, M-S, and S-M are enumerated in Tables \ref{errorcase:ss}, \ref{errorcase:ms}, and \ref{errorcase:sm}, respectively. 
Additionally, a comprehensive breakdown of error frequencies across different models and error categories is illustrated in Table \ref{tab:error_counts}.

\begin{table}[t]
\caption{Error cases of GPT-4 results on S-S.} 
\vspace{-0.2cm}
\resizebox{1\textwidth}{!}{%
\begin{tabular}{l|l|l}
% \hline
\toprule
\textbf{Type} & \textbf{Description} & \textbf{Case} \\ 
% \hline
\midrule
\begin{tabular}[c]{@{}l@{}}\textbf{Tool}\\\textbf{Selection}\end{tabular}
    &  \begin{tabular}[c]{@{}l@{}}The model incorrectly\\selects an unintended\\tool.\end{tabular}
    &  \begin{tabular}[c]{@{}l@{}}
        \textcolor{blue}{User}: Please book 5 economy bus tickets from Sacramento to San Diego forthe 10th\\of this month. \\
        \textcolor{red}{Predict}: Action: \colorbox{orange!15}{FindBus} Action Input: \{``departure\_date'' : ``2024-01-10'', ...  \\
        \textcolor{green}{Target}: Action: \colorbox{orange!15}{BuyBusTicket} Action Input: \{``departure\_date'' : ``2024-01-10'', ... \\
    \end{tabular} \\
    \midrule
    % \hline
\begin{tabular}[c]{@{}l@{}}\textbf{Paramete}\\\textbf{Omission}\end{tabular} 
    &  \begin{tabular}[c]{@{}l@{}}The model omits nece\\ssary parameters in \\the results.\end{tabular}
    & \begin{tabular}[c]{@{}l@{}}
    \textcolor{blue}{User}: I need to find an affordable round trip flight from Washington to Mexico for 4\\people, departing on the 10th and returning on the 13th of this month. \\
    \textcolor{red}{Predict}: Action: SearchRoundtripFlights Action Input: \{``origin\_city'' : ``washington'',\\ ``destination\_city'':``mexico'', ``departure\_date'': ``2024-01-10'', ``return\_date'':\\``2024-01-13'', ``passengers'' : ``4''\} \\
    \textcolor{green}{Target}: Action: SearchRoundtripFlights Action Input: \{``origin\_city'' : ``washington'',\\ ``destination\_city'':``mexico'', ``departure\_date'': ``2024-01-10'', ``return\_date'':\\``2024-01-13'', ``passengers'' : ``4'', \colorbox{orange!15}{``seating\_class'' : ``economy''} \} \\
    \end{tabular} \\
    % \hline
    \midrule

\begin{tabular}[c]{@{}l@{}}\textbf{Parameter}\\\textbf{Misalignment}\end{tabular}
    & \begin{tabular}[c]{@{}l@{}} The model did not \\ populate the parameters \\ according to the \\ specified criteria. \end{tabular}
    & \begin{tabular}[c]{@{}l@{}}
    \textcolor{blue}{User}: I need to find a round trip flight to NYC from London, leaving day after\\tomorrow and returning on the 9th, in Economy class. \\
    \textcolor{red}{Predict}: Action: SearchRoundtripFlights Action Input: \{``departure\_date'' : ``2024-01-06'',\\ ``return\_date'' : ``2024-01-09'', ``seating\_class'' : ``Economy'', ``origin\_airport'' : ``London'',\\ ``destination\_airport'' : \colorbox{orange!15}{``nyc''}\} \\ 
    \textcolor{green}{Target}: Action: SearchRoundtripFlights Action Input: \{``departure\_date'' : ``2024-01-06'',\\``return\_date'' : ``2024-01-09'', ``seating\_class'' : ``Economy'', ``origin\_airport'' : ``London'',\\ ``destination\_airport'' : \colorbox{orange!15}{``New York''}\}
    \end{tabular} \\
    \bottomrule
    % \hline
\end{tabular}%
}
\label{errorcase:ss}
\vspace{-0.5cm}
\end{table}

\begin{table*}[ht]
\centering
\caption{Error cases of GPT-4 results on M-S.} 
\resizebox{\textwidth}{!}{%
\begin{tabular}{l|l|l}
\toprule
\textbf{Type} & \textbf{Description} & \textbf{Case} \\ 
\midrule 
\begin{tabular}[c]{@{}l@{}}\textbf{Repeated}\\\textbf{Call}\end{tabular} & 
\begin{tabular}[c]{@{}l@{}}When the information\\already exists in the\\previous conversation\\history, the tool shou-\\ld not be called\\repeatedly.\end{tabular} & 
\begin{tabular}[c]{@{}l@{}}
    \textcolor{blue}{User}: Milton Country Park \\ 
    \textcolor{purple}{Assistant}: ...Action: SearchAttraction...\\ 
    \textcolor{yellow!50!black}{Observation}: \{``name'': ``Milton Country Park'', ``address'': ``milton county park, milton'', \\ ``type'': ``park'', ``entrance\_fee'': ``free'', \textbf{``phone'': ``765-987554''}\} \\ 
    \textcolor{purple}{Assistant}: It's a free park on the north side of town.\\ 
    \textcolor{blue}{User}: Could I get the phone number for that? \\ 
    \textcolor{red}{Predict}: Thought: The user wants the phone about Milton Country Park, \\so I need to use the searchAttraction API to get the details. \\ 
    \colorbox{orange!15}{Action: SearchAttraction} Action Input: ... \\ 
    \textcolor{green}{Target}: Thought: The phone number has be received in the history,\\so there is no need to call any tool.
    \colorbox{orange!15}{The phone number is 765987554.} \\ 
    \end{tabular} \\ 
\midrule 
\begin{tabular}[c]{@{}l@{}}\textbf{Parameter}\\\textbf{Hallucination}\end{tabular} & 
\begin{tabular}[c]{@{}l@{}}The model incorrectly \\ assumes a parameter \\ without sufficient context.\end{tabular} & 
\begin{tabular}[c]{@{}l@{}}
    ...(Dialogue history)... \\ 
    \textcolor{blue}{User}: ... I need a bus for two people that would be leaving Portland on the 5th. \\
    ...(Dialogue history)... \\ 
    \textcolor{purple}{Assistant}: ...Action: FindBus...\\ 
    \textcolor{yellow!50!black}{Observation}: \{``fare'': ``39'', ``from\_location'': ``Portland'', ``from\_station'': ``Portland Bus\\Station'', ``leaving\_date'': ``2024-01-05'', ``leaving\_time'': ``10:20'', ``to\_location'': ``Vancouver'',\\``to\_station'':``Pacific Central Station'', ``transfers'': ``0'', ``travelers'': ``2''\}\\ 
    \textcolor{blue}{User}: I want to reserve the hotel room. \\ 
    \textcolor{purple}{Assistant}: ...You would like to check in on 5th? \\ 
    \textcolor{blue}{User}: I want to check in on the 8th.\\ 
    \textcolor{red}{Predict}: Thought: The user has provided a check-in date, which, combined with information\\from past queries (hotel name and destination), enables the ReserveHotel API to reserve a\\hotel for the user. 
    Action: ReserveHotel Action Input: \{``hotel\_name'': ``Atrium Inn Vancouver'', \\``check\_in\_date'':``2024-01-08'', \colorbox{orange!15}{``number\_of\_days'': ``1''}, ``destination'': ``Vancouver''\} \\ 
    \textcolor{green}{Target}: Thought: A specific check-in date has been provided, but the duration of the stay is still \\unknown.\colorbox{orange!15}{For how many days?}\\ 
\end{tabular} \\ 
\midrule 
\begin{tabular}[c]{@{}l@{}}\textbf{Parameter}\\\textbf{Inheritance}\end{tabular} & 
\begin{tabular}[c]{@{}l@{}}The model should inherit\\the information paramet-\\ers from the conversation\\history and update the\\current state in a timely\\manner.\end{tabular} & 
    \begin{tabular}[c]{@{}l@{}}
    \textcolor{blue}{User}: Is Dumbo playing today in Santa Rosa? \\ 
    \textcolor{purple}{Assistant}: ...Action: GetTimesForMovie... \\ 
    \textcolor{blue}{User}: Could you get me a taxi there? How long is that? \\
    \textcolor{yellow!50!black}{Observation}: \{``genre'': ``Fantasy'', ``location'': ``Santa Rosa'', ``movie\_name'': ``Dumbo'',  ``price'':\\``9'',``show\_date'': ``2024-01-04'', ``show\_time'': ``22:00'', ``show\_type'': ``3d'', \textbf{``street\_address'':}\\\textbf{``409 Aviation Boulevard''}, ``theater\_name'': ``Airport Stadium 12''\} \\
    \textcolor{red}{Predict}: Thought: The user want to get a taxi to go to the movie theater. \\ 
    Action: GetRide Action Input: \{``destination'': ``\colorbox{orange!15}{theater}'', ``number\_of\_riders'': ``1''\} \\ 
    \textcolor{green}{Target}: Thought: The user want to get a taxi to go to the movie  which is at 409 Aviation\\Boulevard. I can call GetRide. \\ 
    Action: GetRide Action Input: \{``destination'': ``\colorbox{orange!15}{409 Aviation Boulevard}'', ``number\_of\_riders'':\\ ``1''\} \\ 
\end{tabular} \\ 
\bottomrule
\end{tabular}%
}
\label{errorcase:ms} % Add a label for referencing if needed
\end{table*}

\begin{table}[ht]
\caption{Error cases of GPT-4 results on S-M.} 
\vspace{-0.2cm}
\resizebox{1\textwidth}{!}{%
\begin{tabular}{l|l|l}
% \hline
\toprule
\textbf{Type} & \textbf{Description} & \textbf{Case} \\ 
% \hline
\midrule
    \begin{tabular}[c]{@{}l@{}}\textbf{Call}\\\textbf{Fewer} \textbf{Tools}\end{tabular}
        &  \begin{tabular}[c]{@{}l@{}}The model fails to call all\\the necessary tools.\end{tabular}
        &  \begin{tabular}[c]{@{}l@{}}
        \textcolor{blue}{User}: 
        I need to set up an appointment with a dentist in Los Angeles who \\
        offers cosmetic services on June 15th at 2:00 pm. Afterwards, set me\\
        a reminder two hours before the appointment named ``Dentist Appointment''. \\ 
        \textcolor{red}{Predict}: \\
        Action: FindProvider... \\
        Action: SetAppointmentReminder... \\
        \textcolor{green}{Target}: \\
        Action: FindProvider... \\
        \colorbox{orange!15}{Action: BookAppointment}... \\
        Action: SetAppointmentReminder...
        \end{tabular} \\ 
        % \hline
        \midrule
    \begin{tabular}[c]{@{}l@{}}\textbf{Call}\\\textbf{More} \textbf{Tools}\end{tabular} 
        &  \begin{tabular}[c]{@{}l@{}}The model calls \\ additional unnecessary tools.\end{tabular}
        & \begin{tabular}[c]{@{}l@{}}
        \textcolor{blue}{User}: Hi, can you help me find a song of 2019. And modify it as my 3pm\\alarm's sound?  \\ 
        \textcolor{red}{Predict}: \\
        Action: LookupSong... \\
        \colorbox{orange!15}{Action: GetAlarm}... \\
        Action: ChangeAlarmSound... \\
        \textcolor{green}{Target}: \\
        Action: LookupSong... \\
        Action: ChangeAlarmSound...
        \end{tabular} \\ 
        % \hline
    \midrule
\begin{tabular}[c]{@{}l@{}}\textbf{Call}\\\textbf{Wrong} \textbf{Tool}\end{tabular}
    & \begin{tabular}[c]{@{}l@{}}The model selects an inapp-\\ropriate tool for the task, po-\\tentially leading to incorrect\\ results.\end{tabular}
    & \begin{tabular}[c]{@{}l@{}}
    \textcolor{blue}{User}: 
    I'm looking for a pop music to play and set as my alarm sound for my\\
    3pm alarm today. Can you help me with that? \\
    \textcolor{red}{Predict}: \\
    Action: LookupSong... \\
    \colorbox{orange!15}{Action: SetAlarm}... \\
    \textcolor{green}{Target}: \\
    Action: LookupSong... \\
    \colorbox{orange!15}{Action: ChangeAlarmSound}
\end{tabular} \\ 
\bottomrule
% \hline
\end{tabular}%
}
\label{errorcase:sm}
\vspace{-0.5cm}
\end{table}

\begin{table}[t]
\caption{Error Frequencies across Various Models and Error Types. \textbf{Bold} number signifies the most critical error type for each model. \underline{Underlined} number represents the most critical model for each error type.}
% \resizebox{1.0\textwidth}{!}{
\resizebox{\textwidth}{!}{
\begin{tabular}{c|cccc|ccc}
\toprule
\multirow{2}{*}{\textbf{Models}} & \multicolumn{4}{c|}{\textbf{Action Error}} & \multirow{2}{*}{\textbf{Parameter Error}} &\multirow{2}{*}{\textbf{Format Error}}  &\multirow{2}{*}{\textbf{SUM}}\\
\cmidrule(lr){2-5} 
& \textbf{Missed} & \textbf{Excessive}   &\textbf{Incorrect}  & \textbf{SUM}&&&\\ 
\midrule

\rowcolor{red!20}\multicolumn{8}{c}{\textit{S-S}}\\ 
\midrule
GPT4   & \multicolumn{1}{c|}{0} &\multicolumn{1}{c|}{10} & \multicolumn{1}{c|}{3} & 13 & \multicolumn{1}{c|}{\textbf{26}} & 4 & 43\\ 
LLaMA3-70B-Instruct & \multicolumn{1}{c|}{2} & \multicolumn{1}{c|}{9} & \multicolumn{1}{c|}{5} & 16 & \multicolumn{1}{c|}{\textbf{27}} & 1 & 44 \\ 
ChatGLM4-9B  & \multicolumn{1}{c|}{16}  & \multicolumn{1}{c|}{4} & \multicolumn{1}{c|}{8} & 28 & \multicolumn{1}{c|}{\textbf{35}} & 1 & 64\\ 
LLaMA3-8B-Instruct & \multicolumn{1}{c|}{9} & \multicolumn{1}{c|}{3} & \multicolumn{1}{c|}{10}  & 22 & \multicolumn{1}{c|}{\textbf{38}} & 12 & 72\\ 
MTU-LLaMA(Ours) & \multicolumn{1}{c|}{8}      & \multicolumn{1}{c|}{6} & \multicolumn{1}{c|}{10} & 24 & \multicolumn{1}{c|}{\textbf{26}} & 0 & 50\\  \midrule

\rowcolor{red!5}\multicolumn{8}{c}{\textit{S-M}}\\ 
\midrule
GPT4 &  \multicolumn{1}{c|}{8} & \multicolumn{1}{c|}{6}         & \multicolumn{1}{c|}{4}  & 18  & \multicolumn{1}{c|}{\textbf{36}} & 0 &  54 \\  
LLaMA3-70B-Instruct     & \multicolumn{1}{c|}{9}      & \multicolumn{1}{c|}{9} & \multicolumn{1}{c|}{13} & 31 & \multicolumn{1}{c|}{\textbf{56}} & 4 & 91 \\ 
ChatGLM4-9B & \multicolumn{1}{c|}{28}     & \multicolumn{1}{c|}{2}         & \multicolumn{1}{c|}{10}      & 40             & \multicolumn{1}{c|}{\textbf{52}} & 1     & 93 \\ 
LLaMA3-8B-Instruct & \multicolumn{1}{c|}{8} & \multicolumn{1}{c}{25}       & \multicolumn{1}{c|}{11} & \textbf{44}  & \multicolumn{1}{c|}{22} & 28    & 94 \\ 
MTU-LLaMA(Ours) & \multicolumn{1}{c|}{8}      & \multicolumn{1}{c|}{4}         & \multicolumn{1}{c|}{5} & 17 & \multicolumn{1}{c|}{\textbf{49}}    & 0 & 66 \\
\midrule

\rowcolor{orange!20}\multicolumn{8}{c}{\textit{M-S}}\\ 
\midrule
% \hline
GPT4  & \multicolumn{1}{c|}{25}     & \multicolumn{1}{c|}{18}        & \multicolumn{1}{c|}{4} & 47 & \multicolumn{1}{c|}{\textbf{96}} & 14  & 157 \\ 
LLaMA3-70B-Instruct  & \multicolumn{1}{c|}{69}     & \multicolumn{1}{c|}{22}        & \multicolumn{1}{c|}{3}  & \textbf{94} & \multicolumn{1}{c|}{92}                                 & 11   & 197 \\ 
ChatGLM4-9B & \multicolumn{1}{c}{\underline{111}}    & \multicolumn{1}{c|}{9}         & \multicolumn{1}{c|}{9}  & \textbf{129} & \multicolumn{1}{c|}{112} & 40  & 281 \\ 
LLaMA3-8B-Instruct & \multicolumn{1}{c|}{91}     & \multicolumn{1}{c|}{24}        & \multicolumn{1}{c|}{\underline{17}} & \textbf{132} & \multicolumn{1}{c|}{\underline{113}} & \underline{101}    & 346              \\ 
MTU-LLaMA(Ours)  & \multicolumn{1}{c|}{56}  & \multicolumn{1}{c|}{24} & \multicolumn{1}{c|}{3}   & \textbf{83}  & \multicolumn{1}{c|}{73}  & 30   & 186\\
\midrule

\rowcolor{orange!10}\multicolumn{8}{c}{\textit{M-M}}\\ 
\midrule
GPT4   & \multicolumn{1}{c|}{11}     & \multicolumn{1}{c|}{38}        & \multicolumn{1}{c|}{4} & \textbf{53} & \multicolumn{1}{c|}{44} & 0 & 97       \\ 
LLaMA3-70B-Instruct     & \multicolumn{1}{c|}{8}      & \multicolumn{1}{c|}{114}       & \multicolumn{1}{c|}{7} & \textbf{129}                   & \multicolumn{1}{c|}{62} & 5 & 196                                      \\ 
ChatGLM4-9B  & \multicolumn{1}{c|}{28}     & \multicolumn{1}{c|}{145}       & \multicolumn{1}{c|}{10}   & \textbf{183}                & \multicolumn{1}{c|}{38}                                 & 15 & 236                                     \\ 
LLaMA3-8B-Instruct    & \multicolumn{1}{c|}{2}      & \multicolumn{1}{c|}{\underline{207}}       & \multicolumn{1}{c|}{4}    & \textbf{\underline{213}}                & \multicolumn{1}{c|}{6}  & 46 & 265                                     \\ 
MTU-LLaMA(Ours)      & \multicolumn{1}{c|}{14}     & \multicolumn{1}{c|}{103}       & \multicolumn{1}{c|}{3}   & \textbf{120}   & \multicolumn{1}{c|}{28} & 1 & 149 \\
\midrule
\textbf{SUM} & \multicolumn{1}{c|}{\textbf{511}}&\multicolumn{1}{c|}{\textbf{782}}&\multicolumn{1}{c|}{\textbf{143}}&\multicolumn{1}{c|}{\textbf{1436}}&\multicolumn{1}{c|}{\textbf{1031}}&\multicolumn{1}{c|}{\textbf{314}}&{\textbf{2781}}
\\ 
\bottomrule

% \bottomrule
\end{tabular}
}
% \vspace{-2mm}
\label{tab:error_counts}
% \caption{Results of different models on the \textbf{normal} set of MTU-Eval \textbf{(S-S \& M-S)}. ``S-S'' and ``M-S'' denote ``Single-Turn Single-Tool'' and ``Multi-Turn Single-Tool'' settings, respectively.}
% % Note that ``ZS'', ``ZS-COT'', ``FS'' represents ``zero-shot'', ``zero-shot w/ chain-of-thought'' and ``few-shot'', repsectively. Models are grouped roughly according to their model sizes.}
% % \vspace{-3mm}
% \label{tab:singletool-normal}
\end{table}

\paragraph{Details on the Human Evaluation.}\label{app:human} 
Five annotators with extensive expertise in large language models (LLM) are engaged to assess the prediction outcomes. The annotators assign win/lose scores where a score of 1 represents a victory for GPT-3.5, a score of 0 indicates a tie, and a score of -1 signifies a win for LLaMA3-8B.
% The annotation guidelines are as follows
% % For the three metrics, we require human annotators to assess them from three perspectives: 
% the success rate of multi-turn tool-use, the number of tools called in multi-tool scenarios, and the order of tool called. 
% We then compare the human scores with those derived from the new metrics and calculate their Pearson correlation coefficient. 

\end{document}